\pdfoutput=1

\documentclass[11pt]{article}

\usepackage[preprint]{acl}

\usepackage{times}
\usepackage{latexsym}

\usepackage[T1]{fontenc}

\usepackage[utf8]{inputenc}

\usepackage{microtype}

\usepackage{inconsolata}

\usepackage{graphicx}

\usepackage{microtype}
\usepackage{amsmath}
\usepackage{amssymb}
\usepackage{mathtools}
\usepackage{amsthm}
\usepackage{bbm}
\usepackage{subcaption}
\usepackage{multirow}
\usepackage{algpseudocode}
\usepackage{enumitem}
\usepackage{floatrow}
\usepackage{booktabs}

\title{Matryoshka-Adaptor: Unsupervised and Supervised Tuning \\for Smaller Embedding Dimensions}

\author{Jinsung Yoon, Raj Sinha, Sercan \"{O}. Ar{\i}k, Tomas Pfister \\
  Google Cloud AI \\
  \texttt{\{jinsungyoon, sinharaj, soarik, tpfister\}@google.com}}

\begin{document}
\maketitle
\begin{abstract}
Embeddings from Large Language Models (LLMs) have emerged as critical components in various applications, particularly for information retrieval. 
While high-dimensional embeddings generally demonstrate superior performance as they contain more salient information, their practical application is frequently hindered by elevated computational latency and the associated higher cost.
To address these challenges, we propose Matryoshka-Adaptor, a novel tuning framework designed for the customization of LLM embeddings.
Matryoshka-Adaptor facilitates substantial dimensionality reduction while maintaining comparable performance levels, thereby achieving a significant enhancement in computational efficiency and cost-effectiveness.
Our framework directly modifies the embeddings from pre-trained LLMs which is designed to be seamlessly integrated with any LLM architecture, encompassing those accessible exclusively through black-box APIs. 
Also, it exhibits efficacy in both unsupervised and supervised learning settings.
A rigorous evaluation conducted across a diverse corpus of English, multilingual, and multimodal datasets consistently reveals substantial gains with Matryoshka-Adaptor. 
Notably, with Google and OpenAI Embedding APIs, Matryoshka-Adaptor achieves a reduction in dimensionality ranging from two- to twelve-fold without compromising performance across multiple BEIR datasets.
\end{abstract}

\section{Introduction}
\begin{figure}[t!]
    \centering
    \includegraphics[width=\textwidth]{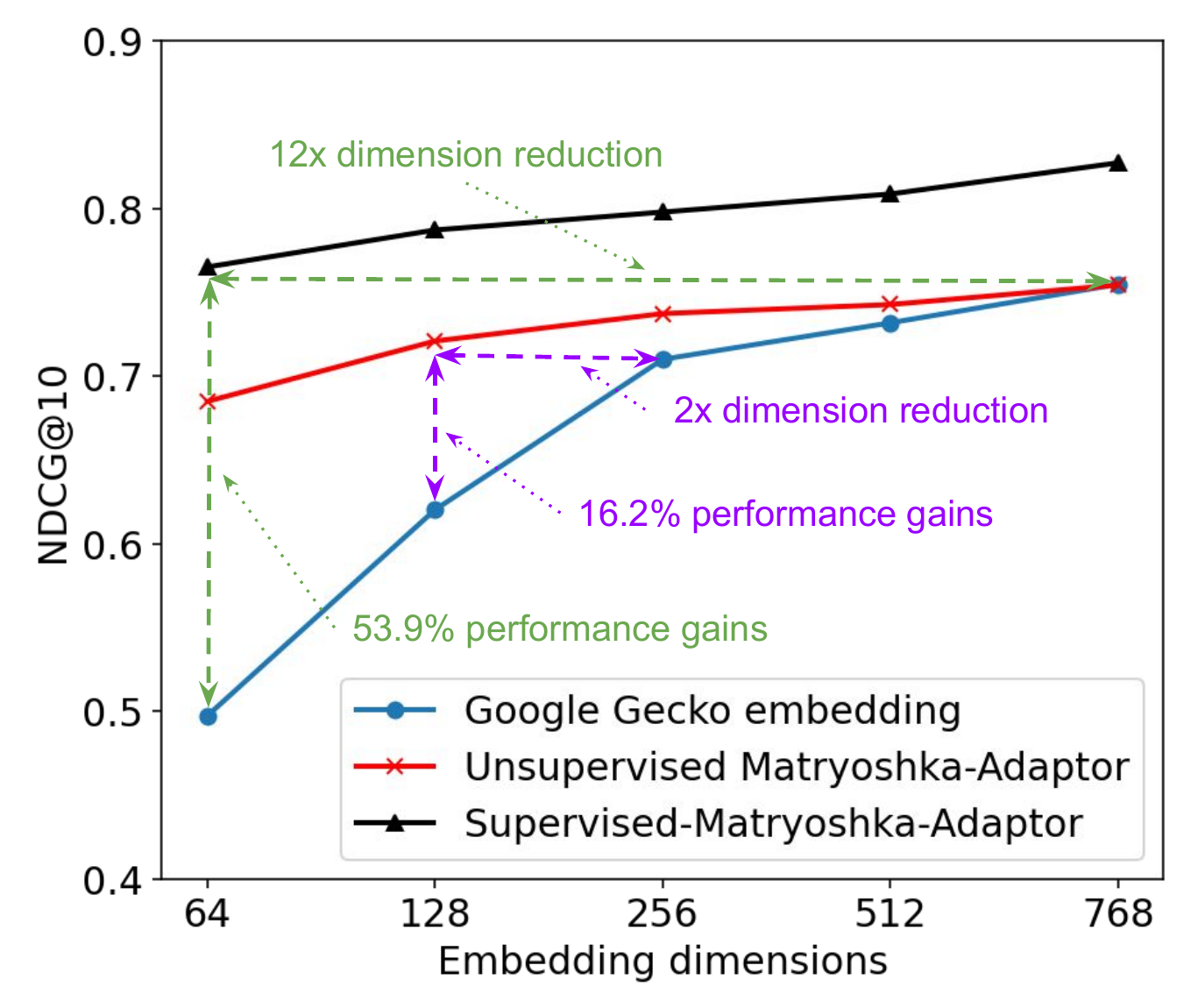}
    \caption{The effectiveness of the Matryoshka Adaptor in dimensionality reduction.
    In both unsupervised (red line) and supervised (black line) settings, the Matryoshka Adaptor showcases the capability to considerably decrease embedding dimensions while maintaining a negligible impact on nDCG@10 retrieval performance with BEIR SciFact dataset. 
    Notably, at the same embedding dimensionality, the utilization of our approach results in significantly improved performance.}
    \label{fig:teaser_figure}
\end{figure}

Large language models (LLMs) have showcased remarkable proficiency in handling various text processing tasks, encompassing question answering, summarization, and mathematical reasoning \citep{brown2020language, chowdhery2022palm, zhang2022opt}.
This success can be partially attributed to their ability to transform raw text into semantically enriched representations, with the quality of these text-to-embedding mappings being of paramount importance \citep{ouyang2022training}.

Embeddings find extensive utilization in a wide array of downstream tasks, with information retrieval (IR) serving as a prominent one \citep{wang2023improving, muennighoff2024generative, li2023towards}. 
IR involves the process of searching for relevant information within a corpus database using queries, and language modeling plays a pivotal role as both queries and corpus data are frequently textual in nature.
In IR systems, text embeddings are commonly employed to rank relevant corpora based on their similarity to queries. 

Numerous LLMs have been developed specifically for extracting embeddings from raw text, including Sentence T5 \citep{ni2021sentence}, OpenAI Embedding APIs \citep{openai-text-embedding}, and Google Embedding APIs  \citep{gcp-text-embedding}. 
However, generation of high-dimensional embeddings often entail high latency and computational costs, thereby limiting their practical application in latency-sensitive scenarios such as large-scale recommendation systems.

Matryoshka representation learning (MRL) \citep{kusupati2022matryoshka,cai2024matryoshka} addresses this limitation at pretraining stage, by generating embeddings that retain similar characteristics even when utilizing only a subset of their dimensions (i.e., Matryoshka properties). 
This enables efficient similarity comparisons with lower-dimensional embeddings and is leveraged by many state-of-the-art models, including those from Google and OpenAI \cite{lee2024gecko, openai-text-embedding}.

In this study, we present Matryoshka-Adaptor, a novel framework designed to transform arbitrary embeddings into embeddings with Matryoshka properties in both unsupervised and supervised learning setups. 
In an unsupervised learning setting, the adaptor learns to transform input embeddings into Matryoshka embeddings using only corpus embeddings.
We introduced pairwise and top-k similarity loss functions to facilitate this process.
In supervised learning setup, the adaptor can be further refined by incorporating relevant (query, corpus) pairs as labeled tuning data. 
Note that we customize embeddings to better suit specific datasets, using both unsupervised and supervised approaches. 
This refinement results in improved Matryoshka properties, surpassing even those of embeddings derived from MRL-trained models.

Extensive experiments conducted across 13 BEIR datasets \citep{thakur2021beir}, 17 MIRACL datasets \citep{zhang2022making}, and 5 multimodal Fashion-200K datasets \citep{han2017automatic} validate the effectiveness of Matryoshka-Adaptor. 
We demonstrate significant performance improvements over the latest Google (for English only, multilingual, and multimodal data) \citep{lee2024gecko} and OpenAI text embedding models \citep{openai-text-embedding}. 
Matryoshka-Adaptor exhibits broad applicability to diverse embedding types, encompassing text, multimodal, multilingual, and use case-specific embeddings. The contributions of this paper are threefold:
\begin{itemize}[leftmargin=*]
    \item We propose a novel tuning framework, Matryoshka-Adaptor that achieves substantial dimensionality reduction in embeddings without sacrificing performance achieved by tailoring embeddings to better fit specific datasets.
    \item Matryoshka-Adaptor is applicable in both unsupervised and supervised settings, consistently improving retrieval performance across various datasets. Also, Matryoshka-Adaptor demonstrates a roughly two-fold (unsupervised) and six-fold (supervised) reduction in dimensionality, with no loss in performance.
    \item The benefits of Matryoshka-Adaptor extend to multimodal learning and multilingual scenarios.
\end{itemize}

\section{Related Work}
\textbf{Matryoshka embeddings. }
\citep{kusupati2022matryoshka} pioneered the development of embedding models whose representations could be substantially reduced in dimensionality without incurring a significant degradation in performance. 
These Matryoshka embedding models are specifically trained to ensure the utility of such truncated embeddings. 
The Matryoshka properties inherent in these models enable fine-grained control over the trade-off between latency and accuracy in downstream tasks utilizing the embeddings. 
Given this advantage, recent embedding models, including those from OpenAI and Google \cite{lee2024gecko}, have integrated Matryoshka properties by employing Matryoshka Representation Learning (MRL) during pre-training.
Our proposed work further enhances the Matryoshka properties through tuning.
We introduce modifications to embeddings that are tailored to target datasets in both an unsupervised and supervised tuning setups.
The proposed tuning yields enhanced Matryoshka properties compared to the original embeddings, even those derived from MRL-trained models.

\textbf{Dimensionality reduction. }
Dimensionality reduction is a well-established framework for reducing the dimensionality of vectors while preserving their inherent properties. 
Principal Component Analysis (PCA) \citep{jolliffe2016principal}, Independent Component Analysis (ICA) \citep{hyvarinen2000independent}, and Non-negative Matrix Factorization (NMF) \citep{lee2000algorithms} are widely used unsupervised dimensionality reduction techniques. 
Linear Discriminant Analysis (LDA) \citep{mclachlan2005discriminant} is a supervised method that leverages labeled data for dimensionality reduction.
While ICA is possible, its limitations in this context arise due to the equal importance of all components, requiring separate model training for each reduced dimension.
NMF, on the other hand, is not suitable for this application as it is only applicable to non-negative embeddings.
Also, PCA's orthogonality properties may limit its reliability for dimensionality reduction for vectors with highly non-linear relationships, especially at high dimensionality.

\textbf{Embedding customization. }
In lieu of employing a single unified model for zero-shot retrieval, embedding customization tailored to individual datasets presents a viable alternative. 
TART \citep{asai2022task} constructs a retrieval system that adapts retrieval based on instructions, incorporating different tasks (e.g. for code, question, or answer) to enhance dense embedding retrieval. 
InstructOR \cite{su2022one} integrates task and domain descriptions for retrieval while tuning embedding models.
Promptagator \cite{dai2022promptagator} leverages in-context learning to generate synthetic query-corpus pairs using a limited number of original pairs, subsequently fine-tuning pre-trained LLMs with these synthetic pairs.
Search-Adaptor \citep{yoon2023search} customizes embeddings for information retrieval using a small number of query-corpus pairs in supervised learning setup.
Unlike these approaches, Matryoshka-Adaptor is applicable to both supervised and unsupervised settings, eliminating the need for labeled query-corpus pairs. 
Additionally, Matryoshka-Adaptor aims not only to enhance full-dimensional embedding performance but also to improve performance across all reduced-dimensionality embeddings.

\section{Unsupervised Matryoshka-Adaptor}
\subsection{Problem formulation}
To facilitate a clear understanding, the proposed framework is formulated within the context of information retrieval. 
However, it is crucial to emphasize that it can be readily generalized to any application involving embeddings. 
In unsupervised settings, we assume the availability of a corpus set, denoted as  $\mathcal{C}=\{c_1, c_2, ..., c_N\}$ and a pre-trained embedding model ($E$). 
The embeddings extracted from the corpus are represented as $\mathcal{C}_E=\{ce_1, ce_2, ..., ce_N\}$, where each embedding vector $ce_i = E(c_i) \in \mathbb{R}^d$. 
Notably, our framework permits the embedding model, $E$, to be treated as a black-box model, and Matryoshka-Adaptor can be directly applied to the extracted embeddings, $\mathcal{C}_E$.

A Matryoshka embedding, characterized by $m$ dimensions, is defined as the initial $m$ dimensions of the original $d$-dimensional embedding, where $m<d$. 
This can be expressed as $\mathcal{C}_E[:m]=\{ce_1[:m], ce_2[:m], ..., ce_N[:m]\}$ where each reduced embedding vector $ce_i[:m] \in \mathbb{R}^m$. 
A fundamental characteristic of Matryoshka embeddings is their capacity to preserve the essential properties of the original embeddings, even within a reduced-dimensional space.

\begin{figure}[t!]
    \centering
    \includegraphics[width=\textwidth]{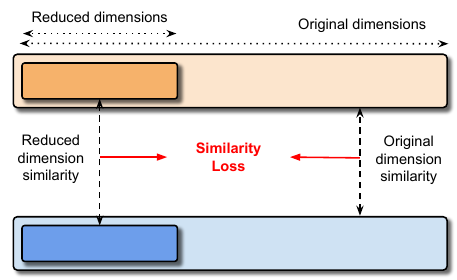}
    \caption{
    Similarity loss is a measure of the discrepancy between the similarity of two embeddings in their original dimensional space and their similarity in a reduced dimensional space.
    If the orange and blue embeddings are chosen randomly, this loss is referred to as pairwise similarity loss.
    If the orange and blue embeddings are selected based on similarity in their original dimensional space (top-k nearest embeddings), this loss is referred to as top-k similarity loss.
    Note that top-k similarity loss focuses on preserving local similarity relationships among neighboring embeddings.}
    \label{fig:unsupervised_matryoshka_adaptor_fig}
\end{figure}

\begin{figure*}[t!]
    \centering
    \includegraphics[width=0.99\textwidth]{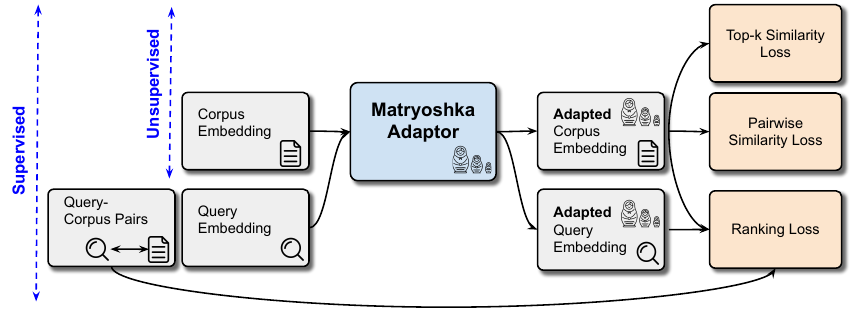}
    \caption{Block diagrams illustrating both the unsupervised and supervised Matryoshka-Adaptor frameworks.
    \textit{Unsupervised Matryoshka-Adaptor}: This variant exclusively utilizes corpus embeddings as input. 
    The training of the adaptor is achieved through a combination of top-k similarity loss and pairwise loss, which are calculated across multiple Matryoshka embeddings with various reduced dimensions.
    \textit{Supervised Matryoshka-Adaptor}: In this variant, query embeddings and query-corpus pairs are provided as supplementary inputs. 
    A ranking loss is incorporated alongside the top-k and pairwise losses to facilitate the training of the adaptor. 
    Similar to the unsupervised setting, all losses are computed across Matryoshka embeddings with various reduced dimensions.}
    \label{fig:matryoshka_adaptor_block_diagram}
\end{figure*}

\subsection{Tuning objective functions}
The proposed Matryoshka-Adaptor, represented by the function $f: \mathbb{R}^d \rightarrow \mathbb{R}^d$, is designed to modify the original embeddings in order to enhance their inherent Matryoshka properties. 
We define the set of customized corpus embeddings as $\hat{\mathcal{C}}_E=\{\hat{ce}_1, \hat{ce}_2, ..., \hat{ce}_N\}$ and their corresponding Matryoshka embeddings as $\hat{\mathcal{C}}_E[:m]=\{\hat{ce}_1[:m], \hat{ce}_2[:m], ..., \hat{ce}_N[:m]\}$ where $\hat{ce}_i = f(ce_i)$. 
The primary objective of the function $f$ is to maximize the Matryoshka properties through this customization process.
This means ensuring the similarity between any two embeddings remains as consistent as possible, whether they are represented in the original high-dimensional space or the reduced low-dimensional space.

To achieve this objective, we introduce two loss functions. The first loss function, denoted as $\mathcal{L}_{pair}$, is designed to preserve the pairwise similarity between the original embeddings in their reduced-dimension Matryoshka form, expressed as:
\begin{align}
    \mathcal{L}_{pair} = \sum_i&\sum_j\sum_m |Sim(ce_i, ce_j) - \\
    &Sim(f(ce_i)[:m], f(ce_j)[:m])|,\nonumber
\end{align}
where $Sim$ represents an arbitrary similarity function, which is chosen to be the cosine similarity.

The second loss function, denoted as $\mathcal{L}_{topk}$, focuses on preserving local similarity relationships among neighboring embeddings:
\begin{align}
    \mathcal{L}_{topk} = \sum_i&\sum_{j\in NN_k(i)}\sum_m |Sim(ce_i, ce_j) - \\
    &Sim(f(ce_i)[:m], f(ce_j)[:m])|,\nonumber
\end{align}
where $NN_k(i)$ denotes the set of the top k most similar embeddings to $ce_i$. 
A visual representation of these two loss functions, illustrating their application across multiple Matryoshka embeddings of varying dimensions, is provided in Fig.~\ref{fig:unsupervised_matryoshka_adaptor_fig}.

In order to mitigate any substantial deviation from the original embeddings, we have integrated regularizations into our methodology. 
Primarily, a skip connection is implemented within the architecture of the learnable function, $f$, ensuring that this function learns solely the difference from the original embedding, represented as $\hat{ce}_i = ce_i + f(ce_i)$.
Furthermore, a reconstruction loss, denoted as $\mathcal{L}_{rec}$, is introduced as an additional regularizer:
\begin{align}
    \mathcal{L}_{rec} = \sum_i |ce_i - f(ce_i)|.
\end{align}
The overall objective function, designed to minimize the aggregate loss, is given as:
\begin{equation}\label{eq:unsup_main}
  \min_f \mathcal{L}_{topk}(f) + \alpha\mathcal{L}_{pair}(f) + \beta\mathcal{L}_{rec}(f),
\end{equation}
with $\alpha, \beta > 0$ being hyperparameters. 
Within the context of the unsupervised tuning setting, we fix their values as $\alpha=1.0, \beta=1.0$.

\section{Supervised Matryoshka-Adaptor}
\subsection{Problem formulation}
For many information retrieval applications, the availability of pairwise data, which indicates the relevance between specific queries and corpora, can considerably improve retrieval performance \citep{yoon2023search}. 
In this section, we introduce a method to further refine the Matryoshka-Adaptor by utilizing (a limited number of) such paired samples.

Let $\mathcal{Q}=\{q_1, q_2, ..., q_M\}$ denote the set of queries, and let $R=\{(q_i, c_j, y_{ij})\}_{i=1:M,j=1:N}$ represent the set of query-corpus relevance triplets, where $y_{ij} > 0$ signifies the relevance score between query $q_i$ and corpus $c_j$. 
The query embeddings extracted from the model ($E$) are denoted as $\mathcal{Q}_E=\{qe_1, qe_2, ..., qe_N\}$ with each query embedding vector $qe_i=E(q_i) \in \mathbb{R}^d$. 
Furthermore, we define Matryoshka query embeddings as $\mathcal{Q}_E[:m]=\{qe_1[:m], qe_2[:m], ..., qe_N[:m]\}$, where $qe_i[:m] \in \mathbb{R}^m$. 
Note that $m$ can be any integer less than the original embedding dimension.

\begin{figure*}[t!]
\centering
\subfloat[OpenAI text-embedding-3-large]{
\includegraphics[width=0.32\textwidth]{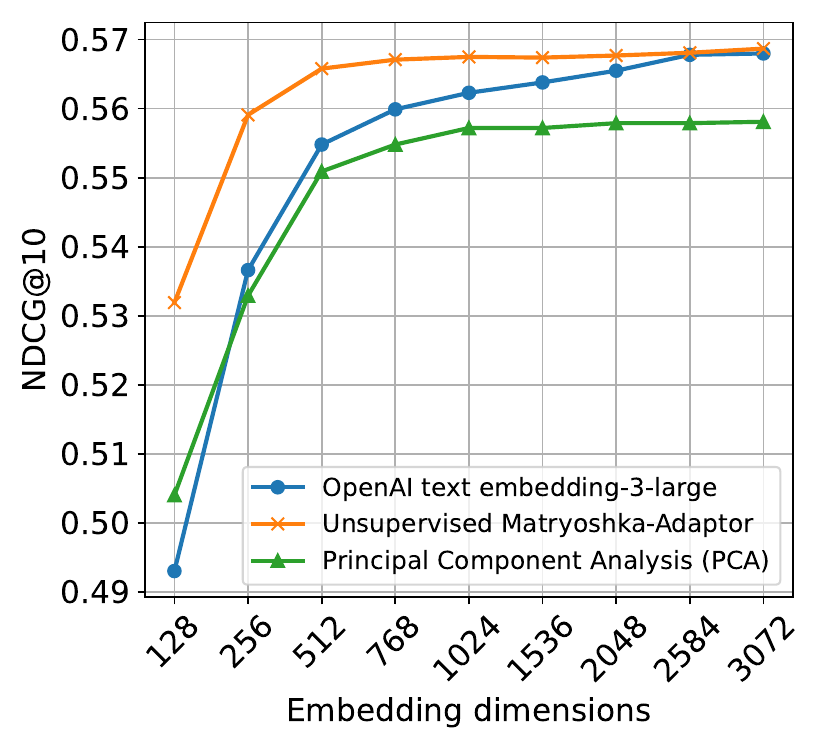}
\label{fig:beir_openai_large}}
\subfloat[OpenAI text-embedding-3-small]{
\includegraphics[width=0.32\textwidth]{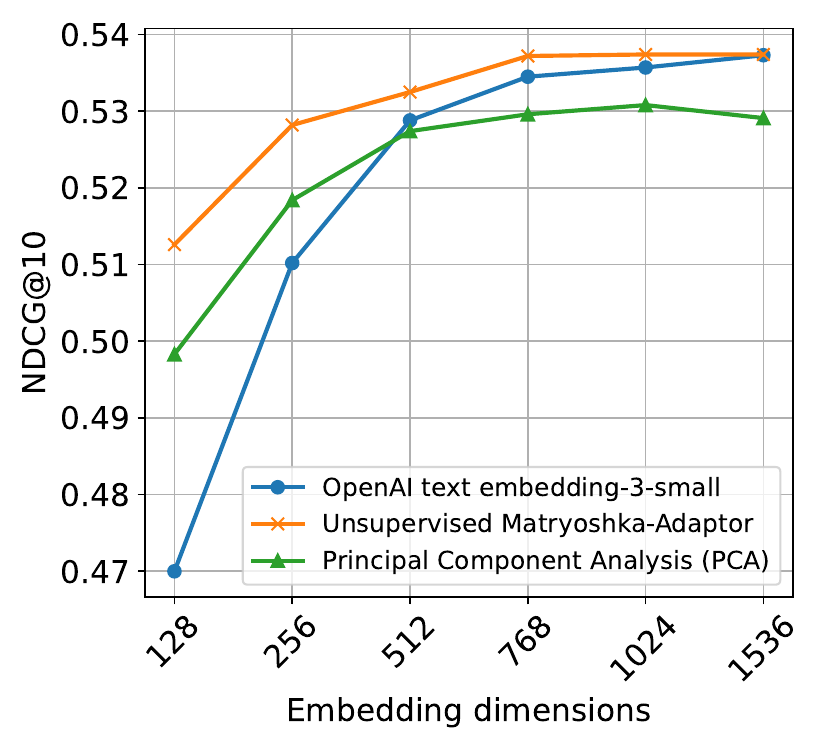}
\label{fig:beir_openai_small}}
\subfloat[Google multimodal embeddings]{
\includegraphics[width=0.32\textwidth]{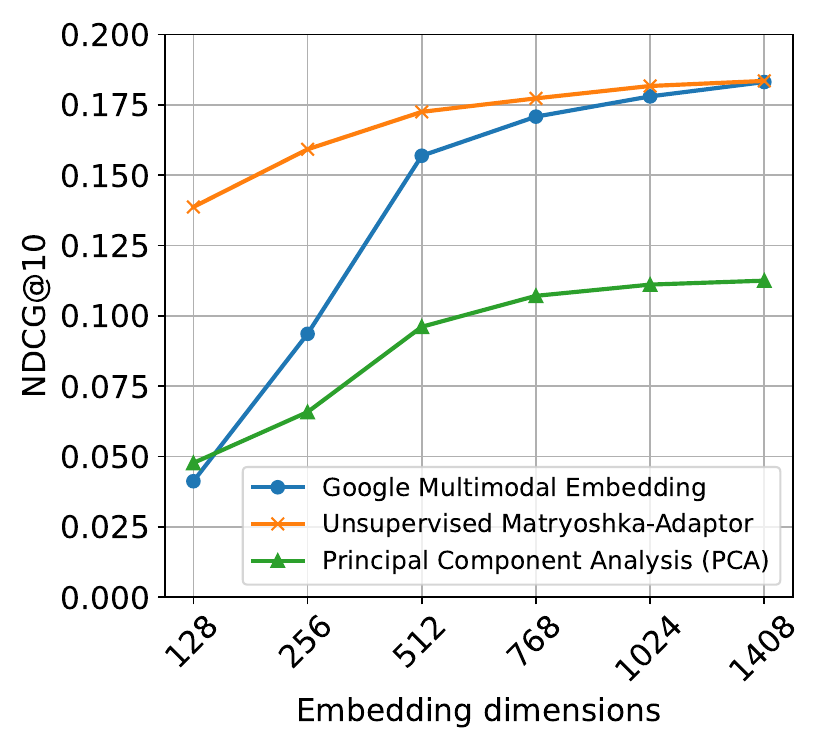}
\label{fig:fashion_gecko_multimodal}}
\caption{Experimental results of the unsupervised Matryoshka-Adaptor applied to three different embedding models: OpenAI text-embedding-3-large (with 3072 dimensions), OpenAI text-embedding-3-small (with 1536 dimensions), and Google multimodal (with 1408 dimensions). 
Text embedding results were obtained using 8 BEIR datasets, while multimodal embedding results were obtained using 5  Fashion-200K datasets.}
\label{fig:exp_unsupervised}
\end{figure*}

\subsection{Tuning objective functions}
In the supervised setting, the Matryoshka-Adaptor ($f$) undergoes optimization to enhance both its Matryoshka properties and the retrieval performance of the Matryoshka embeddings. 
This is accomplished through the utilization of paired query-corpus samples, in conjunction with the original query and corpus embeddings.
Matryoshka ranking loss, denoted as $\mathcal{L}_{rank}$, is introduced to align the ranking between query and corpus considering different Matryoshka embedding dimensions:
\begin{align}
    \mathcal{L}_{rank} &= \sum_i\sum_j\sum_k\sum_m I(y_{ij} > y_{ik})  \\
    &(y_{ij} - y_{ik})\log(1+\exp(s_{ik}[:m] - s_{ij}[:m])),\nonumber
\end{align}
where $s_{ij}[:m]$ represents the cosine similarity between the adapted query embedding $\hat{qe}_i[:m]$ (where $\hat{qe}_i = qe_i + f(qe_i)$) and the adapted corpus embedding $\hat{ce}_j[:m]$.
We use the same adaptor ($f$) for both query and corpus embeddings.
This ranking loss is crucial for effective learning of lower dimensional representations with their information content for the ranking objective being considered.

The supervised Matryoshka-Adaptor is trained using a joint objective function that encompasses the ranking loss as well as the unsupervised Matryoshka losses ($\mathcal{L}_{topk}$ , $\mathcal{L}_{pair}$, and $\mathcal{L}_{rec}$).
This joint training approach aims to improve the quality of the embeddings while preserving their Matryoshka representations.
Query-corpus pairs are employed for the ranking loss, while query and corpus embeddings are utilized for the Matryoshka representation learning. The overall objective function is:
\begin{align}\label{eq:sup_main}
  \min_f & \mathcal{L}_{topk}(f) + \alpha\mathcal{L}_{pair}(f) +\\ &\beta\mathcal{L}_{rec}(f)  + \gamma\mathcal{L}_{rank}(f),  \nonumber
\end{align}
with $\alpha, \beta, \gamma \geq 0$ being hyper-parameters with fixed values as $\alpha=1.0, \beta=1.0$ and $\gamma=1.0$.

To improve convergence, two-stage training strategy is employed. Initially, the Matryoshka-Adaptor is trained in an unsupervised way using Eq. \ref{eq:unsup_main}. 
Subsequently, further tuning is conducted in a supervised way, utilizing Eq. \ref{eq:sup_main}.

\begin{figure*}[!t]
\centering
\subfloat[OpenAI text-embedding-3-large]{
\includegraphics[width=0.32\textwidth]{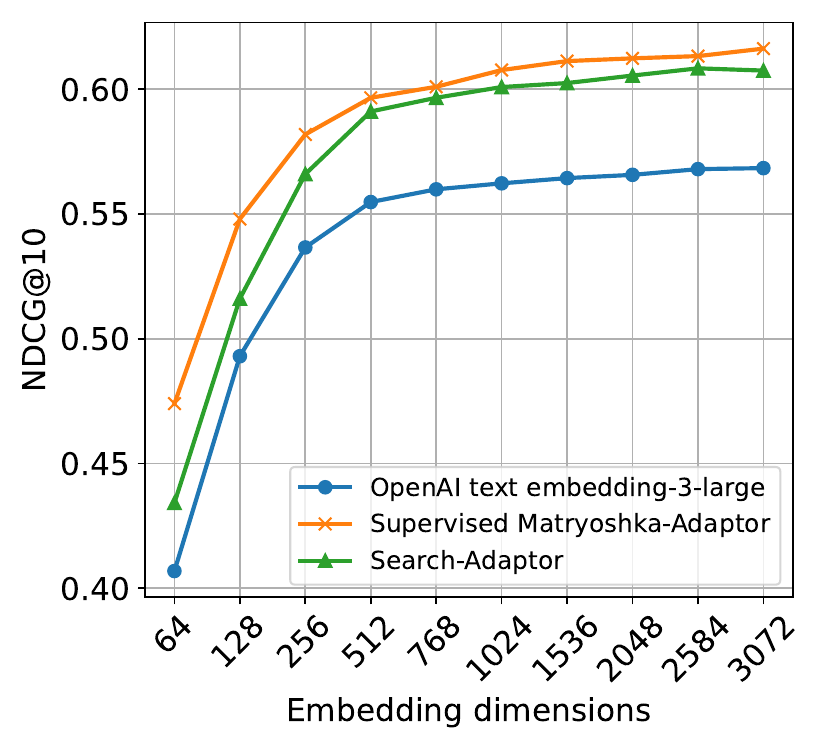}
\label{fig:beir_gecko_supervised}}
\subfloat[Google gecko multilingual embeddings]{
\includegraphics[width=0.32\textwidth]{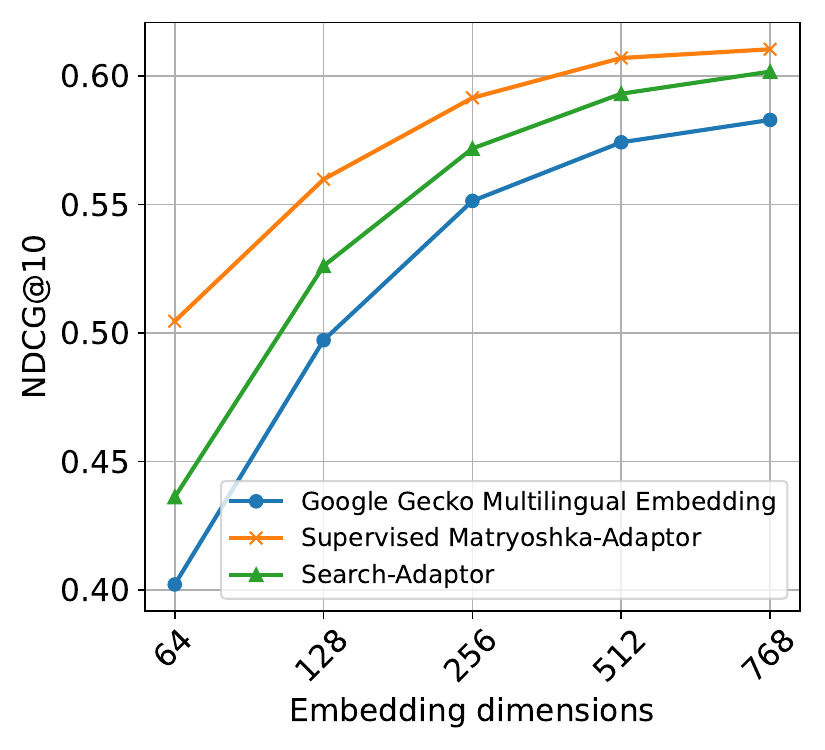}
\label{fig:miracl_gecko_supervised_multilingual}}
\subfloat[Google multimodal embeddings]{
\includegraphics[width=0.32\textwidth]{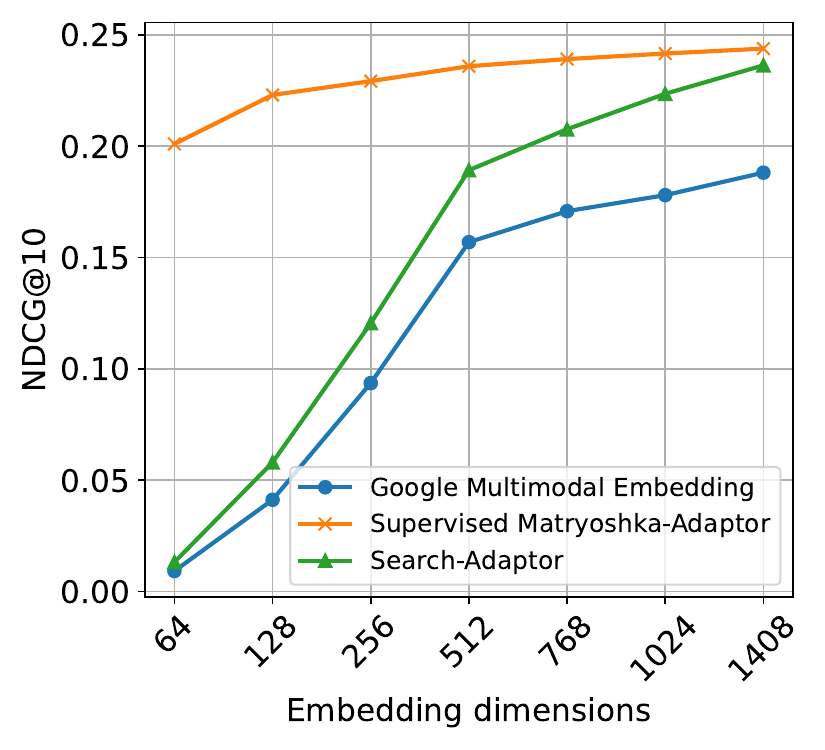}
\label{fig:fashion_gecko_supervised_multimodal}}
\caption{Experimental results of the supervised Matryoshka-Adaptor on retrieval tasks, utilizing three different embedding models: OpenAI text-embedding-3-large (on 8 BEIR datasets), Google Gecko multilingual (on 17 MIRACL datasets), and Google multimodal (on 5 Fashion-200K datasets). Additional results are in Appendix.~\ref{appx:additional_supervised results}.}
\label{fig:exp_supervised}
\end{figure*}

\section{Experiments}
\subsection{Experimental settings}
The Matryoshka-Adaptor is extensively evaluated across a diverse set of 13 BEIR datasets \cite{beir_dataset}, 17 MIRACL datasets \cite{miracl_dataset}, and 5 Fashion-200K datasets \cite{han2017automatic}.
Such a comprehensive evaluation highlights the data-agnostic nature of the Matryoshka-Adaptor. 
Query and corpus embeddings are generated using state-of-the-art models, including Google Gecko text embeddings (English and multilingual) \citep{gcp-text-embedding}, Google multimodal embeddings, and OpenAI text embeddings \citep{openai-text-embedding}. 
This further highlights the model-agnostic nature of the proposed Matryoshka-Adaptor.

During the evaluation phase, both query and corpus embeddings are transformed using the trained Matryoshka-Adaptor. 
Cosine similarity is then computed between the transformed query and corpus embeddings across a spectrum of reduced dimensions (e.g., $d=\{8,16,32,64,128,256,512,768\}$ for Gecko text embeddings).
Retrieval performance is evaluated using the normalized discounted cumulative gain at rank 10 (nDCG@10) metric \cite{jarvelin2002cumulated}, facilitating a comprehensive assessment of performance across various Matryoshka embedding dimensions. 
All reported results represent the average value across the datasets. 
Data statistics and hyper-parameters used in the experiments can be found in Appendix.~\ref{appx:data} and \ref{appx:hyperparams}.
Detailed results can be found in Appendix.~\ref{appendix:unsupervised_detail_result} and \ref{appendix:supervised_detail_result}.

\subsection{Unsupervised tuning}
The Matryoshka-Adaptor can be applied exclusively with corpus embeddings, referred to as the unsupervised setting, enabling customization of embeddings solely on the corpus side. 
In this subsection, we present the impact of the unsupervised Matryoshka-Adaptor on two OpenAI text embedding models and compare its performance with the commonly-used unsupervised dimensionality reduction method, Principal Component Analysis (PCA)  \cite{jolliffe2016principal}.

Fig. \ref{fig:beir_openai_large} and \ref{fig:beir_openai_small} illustrate that the Matryoshka-Adaptor yields significant performance improvements, particularly for lower dimensions, compared to the embeddings of the same dimensionality without it. 
Furthermore, lower dimensional embeddings processed with Matryoshka-Adaptor achieve comparable performance to original embeddings of high dimensionality. 
The adaptor achieves a faster saturation in performance with embedding dimensionality, towards the retrieval performance of original embeddings, underscoring its substantial impact in significantly reducing latency and memory requirements for retrieval applications.

Notably, the latest OpenAI embeddings are already trained with Matryoshka Representation Learning (MRL) \citep{openai-text-embedding}. 
The additional performance gains achieved by Matryoshka-Adaptor are attributed to the tuning process. 
With PCA, some improvements are observed for lower dimensions, however, at higher dimensions, noticeable performance degradation occurs, resulting in performance inferior even to the original embedding. 
This highlights the superiority of the Matryoshka-Adaptor, designed with the goal of salient information preservation for similarity within lower dimensions compared to PCA, a generic unsupervised dimensionality reduction approach.

\subsection{Supervised tuning}
The Matryoshka-Adaptor can be applied in supervised learning setup where a limited number of query-corpus pairs are available. 
In this context, the Supervised Matryoshka-Adaptor is trained utilizing paired query-corpus data. 
The effectiveness of the Supervised Matryoshka-Adaptor is evaluated on 13 BEIR, 17 MIRACL, and 5 Fashion-200K datasets, with query and corpus embeddings being generated using the latest Google Gecko models, including its multilingual and multimodal versions.

As illustrated in Fig. \ref{fig:exp_supervised}, the Supervised Matryoshka-Adaptor consistently outperforms the alternatives, such as Search-Adaptor \cite{yoon2023search}, particularly for lower embedding dimensions. 
Additionally, lower dimensional embeddings processed with the Supervised Matryoshka-Adaptor perform comparably to high dimensional embeddings, showcasing its potential to significantly reduce latency and memory requirements for applications like retrieval.

\begin{figure*}[t!]
\centering
\subfloat[Google gecko-latest embeddings]{
\includegraphics[width=0.32\textwidth]{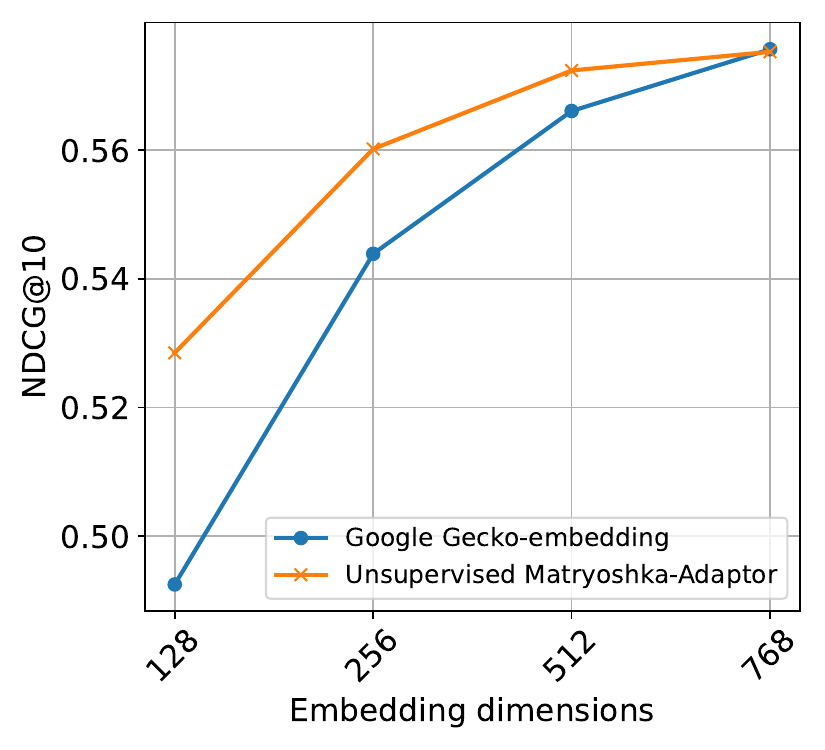}
\label{fig:beir_gecko}}
\subfloat[Google gecko-multilingual embeddings]{
\includegraphics[width=0.32\textwidth]{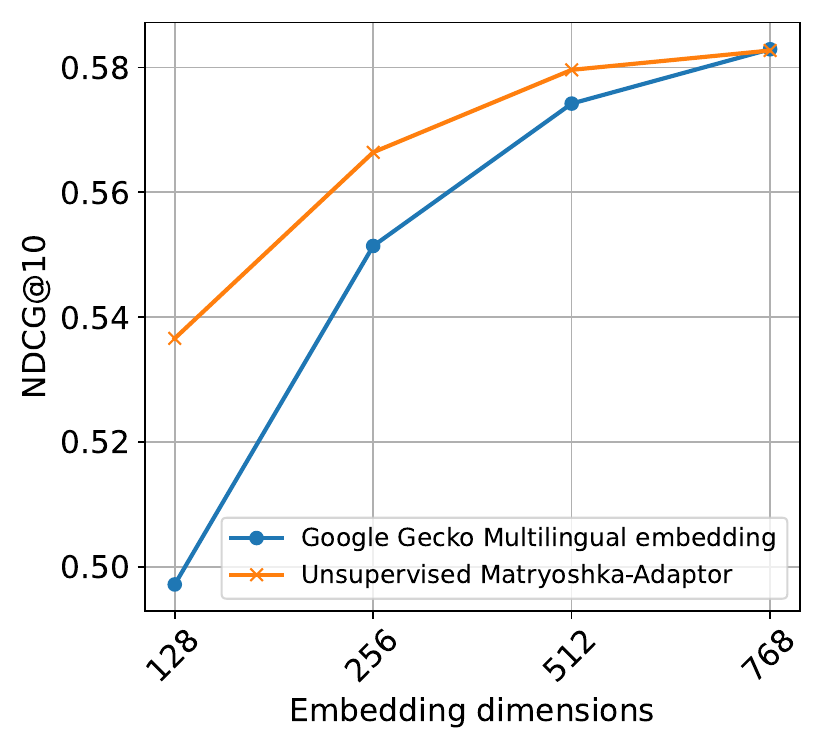}
\label{fig:miracl_gecko_multilingual}}
\subfloat[Google gecko-003 embeddings]{
\includegraphics[width=0.32\textwidth]{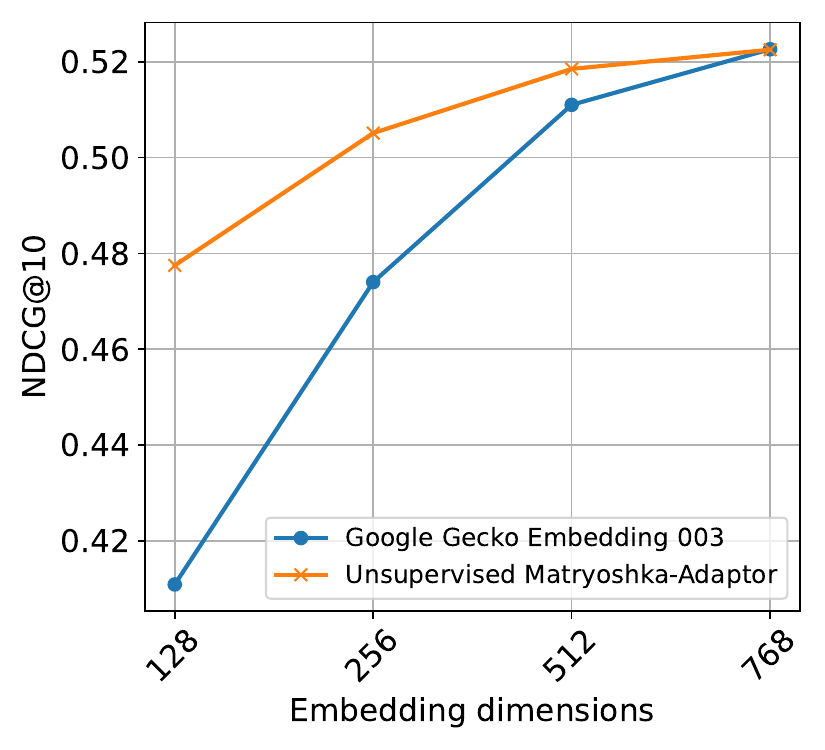}
\label{fig:beir_gecko_old}}
\caption{Experimental results of the unsupervised Matryoshka-Adaptor with three different text embedding models: Google Gecko, Google Gecko multilingual, and Google Gecko-003 (which is not trained with the Matryoshka Representation Learning technique).}
\label{fig:beir_unsupervised_analysis}
\end{figure*}

\subsection{Tuning for Multimodal Embeddings}
As previously established, the Matryoshka-Adaptor framework is not confined to text embeddings but can be generalized to multimodal embeddings as well. 
To illustrate this capability, we applied the Matryoshka-Adaptor to the latest gecko-multimodal embeddings, utilizing the Fashion-200K dataset, which comprises 5 sub-datasets designed for text-to-image retrieval tasks.

Fig. \ref{fig:fashion_gecko_multimodal} and \ref{fig:fashion_gecko_supervised_multimodal} demonstrate the effectiveness of the Matryoshka-Adaptor in consistently improving the performance of multimodal base embedding models for text-to-image retrieval tasks. 
These highlight that the Matryoshka-Adaptor significantly outperforms alternative methods such as PCA, in unsupervised learning setups and Search-Adaptor in supervised learning setups, particularly when lower embedding dimensions are considered.

\subsection{Tuning for Multilingual Embeddings}
Matryoshka-Adaptor is not only model-agnostic but also data-agnostic. Its applicability even extends beyond a single language.
To validate this, we evaluate the performance of Matryoshka-Adaptor on MIRACL datasets, which comprise 17 non-English languages.

Fig. \ref{fig:miracl_gecko_multilingual} and \ref{fig:miracl_gecko_supervised_multilingual} present the results of applying Matryoshka-Adaptor to multilingual retrieval tasks. 
The findings demonstrate that the performance gains achieved through the proposed tuning method are not limited to English but also extend to non-English language datasets. 
Furthermore, the improvements are observed to be model-agnostic, as evident from the successful application of Matryoshka-Adaptor to the latest Gecko multilingual embedding models.

\section{Discussions}
\subsection{Models that are not pretrained with MRL}
A significant advantage of the Matryoshka-Adaptor framework lies in its broad applicability to a wide array of embedding models.
We demonstrate the efficacy of Matryoshka-Adaptor when applied to embedding models that have not been trained using MRL. 
Specifically, we utilize earlier versions of the Gecko embedding models (Google Gecko-003), which do not utilize MRL in their pretraining, in conjunction with BEIR datasets to illustrate the impact of the Unsupervised Matryoshka-Adaptor.

Fig. \ref{fig:beir_gecko_old} presents evidence of the consistent performance improvements achieved by the Matryoshka-Adaptor when applied to non-MRL trained embedding models. 
This observation underscores that the performance gains of Matryoshka-Adaptor originate from customizing the embedding to the specific corpus, and this beneficial impact can be extended to any embedding model, even when they had been trained with MRL.

\begin{figure*}[t!]
\centering
\subfloat[Pairwise and top-k distance metrics]{
\includegraphics[width=0.32\textwidth]{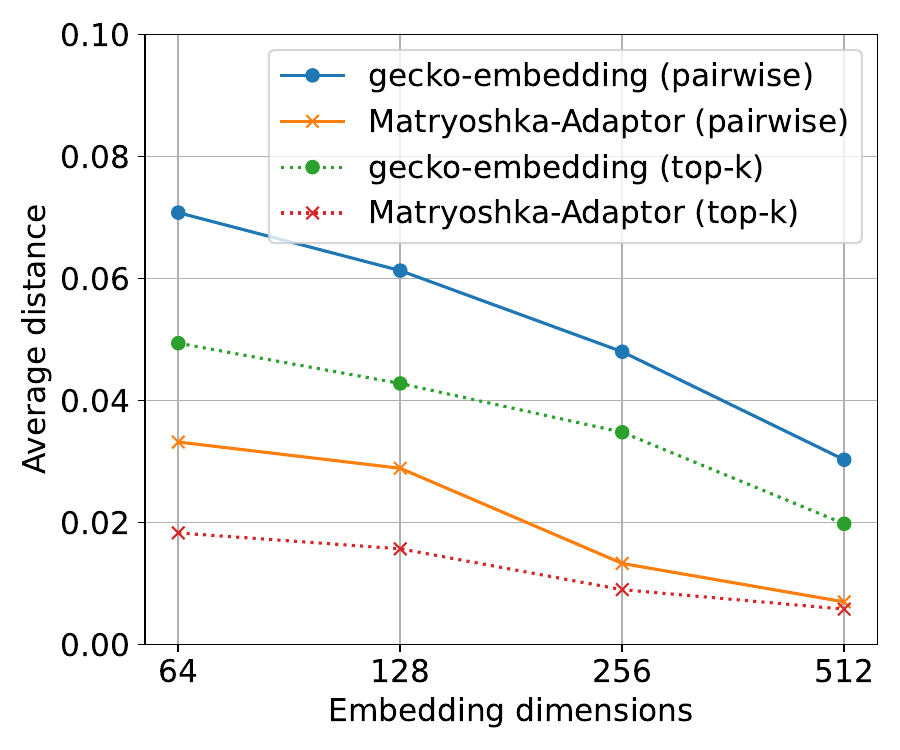}
\label{fig:topk_pairwise_dist}}
\subfloat[nDCG@10 vs pairwise distance]{
\includegraphics[width=0.32\textwidth]{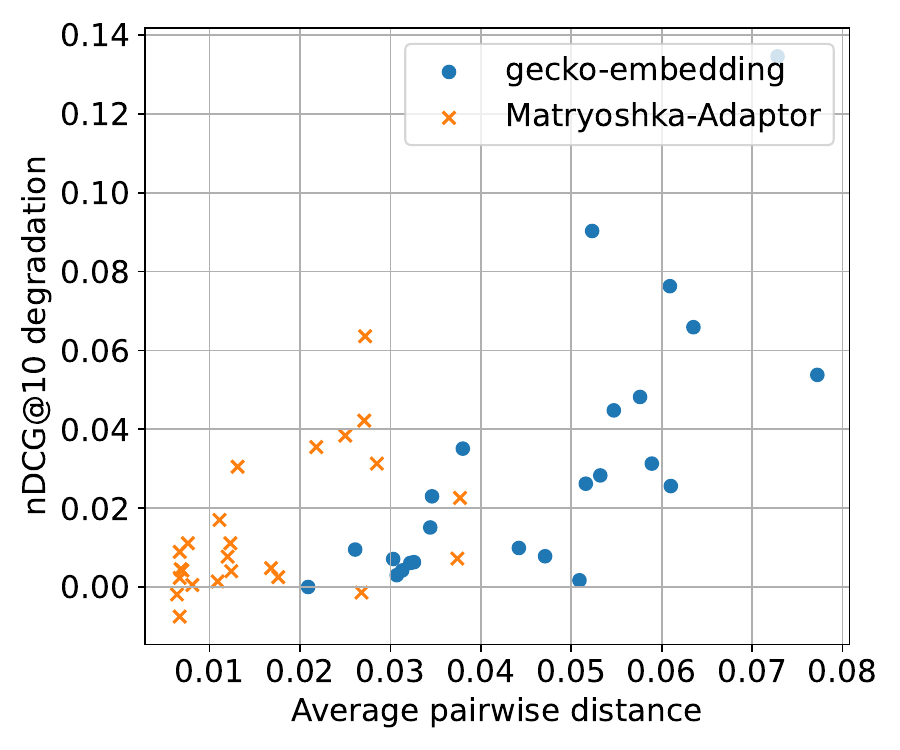}
\label{fig:ndcg_pairwise_dist}}
\subfloat[nDCG@10 vs Top-k distance]{
\includegraphics[width=0.32\textwidth]{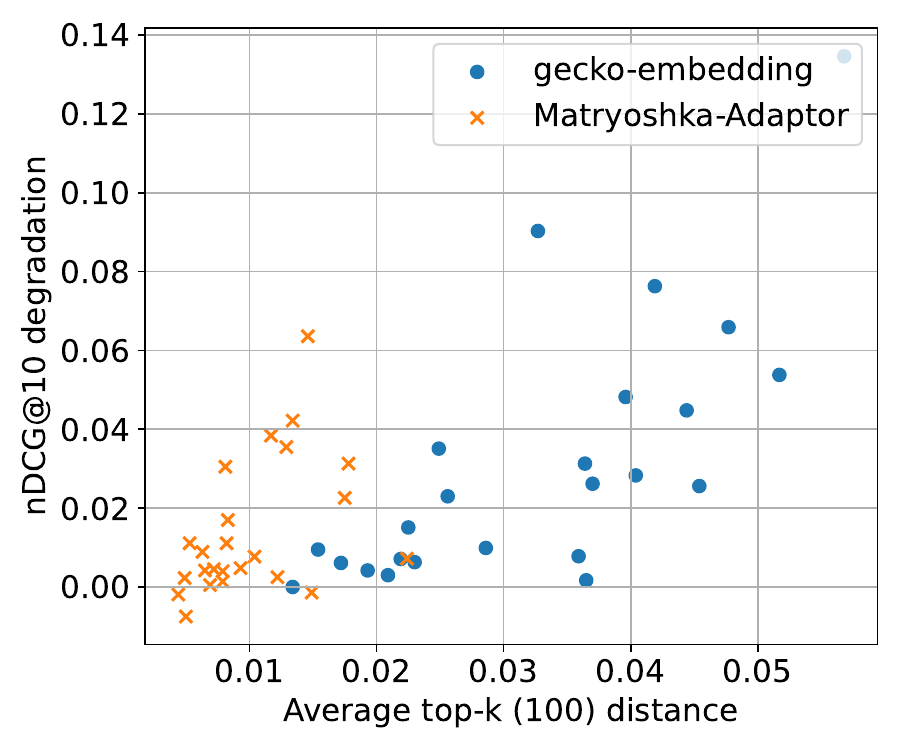}
\label{fig:ndcg_topk_dist}}
\caption{Analysis of distance metrics in unsupervised settings. 
(a) Average pairwise and top-k distances across varying embedding dimensions, compared to base embeddings. 
(b) Correlation between nDCG@10 and average pairwise distance. 
(c) Correlation between nDCG@10 and average top-k distance.}
\label{fig:distance}
\end{figure*}

\subsection{Ablation studies}
To examine the individual contribution of each loss component to the overall performance of Matryoshka-Adaptor, we conduct ablation studies. 
In the unsupervised setting, each of the top-k loss ($\mathcal{L}_{topk}$), pairwise loss ($\mathcal{L}_{pair}$), and reconstruction loss ($\mathcal{L}_{rec}$) is individually excluded from the original loss summation in Eq. \ref{eq:unsup_main}, and the corresponding performance degradation is monitored. 
Specifically, the pairwise loss was excluded by setting $\alpha=0$, and the reconstruction loss is excluded by setting $\beta=0$.
In the supervised setting, we utilize only the ranking loss ($\mathcal{L}_{rank}$) to assess the impact of three unsupervised losses in Eq. \ref{eq:sup_main}.

\begin{table}[h!]
    \centering
    \small
    \begin{tabular}{c|c|c|c|c}
    \toprule
        \multicolumn{5}{c}{\textbf{Unsuperivsed Matryoshka-Adaptor} (nDCG@10)}\\
        \midrule
        Reduced dims & 64 & 128 & 256 & 512 \\
        \midrule
        Baseline & 0.4332 &0.5044 &0.5461 &0.5590 \\
        All three losses & \textbf{0.4845} & \textbf{0.5380} & \textbf{0.5580} & \textbf{0.5652} \\
        \midrule
        w/o $\mathcal{L}_{topk}$ & 0.4798 & 0.5236 & 0.5463 & 0.5630 \\
        w/o $\mathcal{L}_{pair}$  & 0.4745 & 0.5230 & 0.5423 & 0.5598 \\
        w/o $\mathcal{L}_{rec}$ & 0.4824 & 0.5342 & 0.5477 & 0.5621 \\
        \midrule
        \multicolumn{5}{c}{\textbf{Supervised Matryoshka-Adaptor} (nDCG@10)}\\
        \midrule
        Reduced dims & 64 & 128 & 256 & 512 \\
        \midrule
        Baseline & 0.4332 &0.5044 &0.5461 &0.5590 \\
        All four losses & \textbf{0.5047} & \textbf{0.5473} & \textbf{0.5714} & \textbf{0.5902} \\
        \midrule
        Only with $\mathcal{L}_{rank}$ & 0.4767 & 0.5209 & 0.5607 & 0.5787 \\
         \bottomrule
    \end{tabular}
    \caption{Ablation studies on both supervised and unsupervised Matryoshka-Adaptor across 8 BEIR datasets. 
    The best performance in each scenario is in bold.}
    \label{tab:ablation}
\end{table}

As evident in Table \ref{tab:ablation}, each loss function contributes uniquely to the overall performance of Matryoshka-Adaptor in the unsupervised setting. 
Notably, $\mathcal{L}_{topk}$ is particularly beneficial for higher dimensions. 
In the supervised setting, the incorporation of the additional unsupervised losses, including $\mathcal{L}_{topk}$ and $\mathcal{L}_{pair}$, is critical for achieving performance improvements. 
Without these unsupervised losses, the performance gain diminishes by approximately 50\%.

\subsection{Relationship with distance metrics}
In unsupervised learning setup, the absence of labeled data presents a challenge in evaluating the impact of Matryoshka-Adaptor. 
Towards circumventing this issue and shedding light on the impact of dimensionality reduction, we analyze the average pairwise distances and top-k distances within the corpus embeddings on the disjoint validation corpus. 
Subsequently, we examine the relationship between these unsupervised metrics and the supervised retrieval metric nDCG@10, utilizing 8 BEIR datasets with Google Gecko embedding models.

Fig. \ref{fig:topk_pairwise_dist} reveals that the unsupervised Matryoshka-Adaptor results in significantly lower distance metrics compared to the original embeddings. 
This suggests that the proposed method effectively preserves both pairwise and top-k distances among corpus embeddings when employing a reduced subset of embedding dimensions. 
These distance metrics can serve as viable unsupervised proxies for assessing the efficacy of the unsupervised Matryoshka-Adaptor.
To further elucidate this relationship, we also show the correlation between these distance metrics and nDCG@10 (Fig. \ref{fig:ndcg_pairwise_dist} and \ref{fig:ndcg_topk_dist}). 
A strong correlation is observed, indicating that lower distance metrics are associated with smaller performance drop in retrieval tasks, even when using reduced embedding dimensions. 
This suggests that the unsupervised distance metrics can be indicators of retrieval performance, even in the absence of labeled data.

\section{Conclusions}
While high-dimensional embeddings often provide a richer representation of the original data, their practical deployment in real-world applications is frequently hampered by computational costs and latency issues. 
As a result, many applications opt for lower-dimensional embeddings, accepting a compromise in performance. 
The presented Matryoshka-Adaptor framework offers a solution to this dilemma by enabling substantial dimensionality reduction while minimizing performance degradation, in both supervised and unsupervised scenarios. 
Furthermore, due to its versatile nature and applicability to any embedding model, regardless of access to model parameters, the Matryoshka-Adaptor constitutes a valuable tool for improving the efficiency and practicality of embedding-based applications.

\section{Limitations and Future Works}
In the unsupervised setting, picking the optimal hyperparameters for Matryoshka-Adaptor poses a challenge due to the absence of validation data. 
While alternative metrics, such as the proposed distance metrics illustrated in Fig. \ref{fig:distance}, can be employed, these may exhibit higher levels of noise compared to supervised validation metrics. 
One major risk associated with Matryoshka-Adaptor lies in the potential for overfitting to the tuning data, during the tuning process.
Future research endeavors might include extending the proposed approach to encompass multiple modalities in the tuning objective; 
exploring semi-supervised variations; and
investigating simultaneous utilization of multiple datasets during tuning.


\bibliography{mat_adaptor}

\onecolumn

\newpage
\appendix
\section{Data Statistics}\label{appx:data}

\subsection{BEIR datasets}
\begin{table}[h!]
    \small
    \centering
    \begin{tabular}{c|c|c|c}
    \toprule
        \multirow{2}{*}{Datasets} & Number of  & Number of& Number of \\
        &  train pairs &  test pairs & corpus \\
        \midrule
        NFCorpus & 110575 & 12334 & 3633 \\
        SciFact & 919 & 339 & 5183 \\
        Arguana & 703 & 703 & 8674 \\
        SciDocs & 14972 & 14956 & 25657 \\
        FiQA & 14166 & 1706 & 57638 \\
        Trec-Covid & 35460 & 30876 & 171332 \\
        Touche & 1077 & 1137 & 382545 \\
        Quora & 7626 & 15675 & 522931 \\
        NQ & 2097 & 2104 & 2681468 \\
        DBPedia & 5673 & 43515 & 4635922 \\
        HotPotQA & 170000 & 14810 & 5233329 \\
        Fever & 140085 & 7937 & 5416568 \\
        Climate-fever & 2299 & 2382 & 5416593\\
        \bottomrule
    \end{tabular}
    \caption{The statistics of 13 BEIR datasets (sorted by the number of corpus).}
    \label{tab:data_stats}
\end{table}

\subsection{MIRACL datasets}

\begin{table}[h!]
    \small
    \centering
    \begin{tabular}{c|c|c|c}
    \toprule
        \multirow{2}{*}{Datasets} & Number of  & Number of& Number of \\
        &  train pairs &  test pairs & corpus \\
        \midrule
        Yoruba (yo) & 959 & 229 & 49043 \\
        Swahilli (sw) & 9359 & 5092 & 131924 \\
        Bengali (bn) & 16754 & 4206 & 297265 \\
        Hindi (hi) & 11668 & 3494 & 506264 \\
        Telugu (te) & 18608 & 1606 & 518079 \\
        Thai (th) & 21293 & 7573 & 542166 \\
        Indonesian (id) & 41358 & 9668 & 1446315 \\
        Korean (ko) & 12767 & 3057 & 1486752 \\
        Finnish (fi) & 20350 & 12008 & 1883509 \\
        Arabic (ar) & 25382 & 29197 & 2061414 \\
        Persian (fa) & 21844 & 6571 & 2207172 \\
        Chinese (zh) & 13113 & 3928 & 4934368 \\
        Japanese (ja) & 34387 & 8354 & 6953614\\
        Russian (ru) & 33921 & 13100 & 9543918 \\
        Spanish (es) & 21531 & 6443 & 10373953 \\
        French (fr) & 11426 & 3429 & 14636953 \\
        Germany (de) & 2526 & 628 & 15866222 \\
        \bottomrule
    \end{tabular}
    \caption{The statistics of 17 MIRACL datasets (sorted by the number of corpus).}
    \label{tab:miracl_data_stats}
\end{table}

\subsection{Fashion-200K datasets}

\begin{table}[h!]
    \small
    \centering
    \begin{tabular}{c|c|c|c}
    \toprule
        \multirow{2}{*}{Datasets} & Number of  & Number of& Number of \\
        &  train pairs &  test pairs & corpus \\
        \midrule
        Dresses & 15127 & 1567 & 72376 \\
        Jackets & 8105 & 1511 & 71118 \\
        Pants & 9264 & 1758 & 74470 \\
        Skirts & 6822 & 1247 & 47931 \\
        Tops & 13809 & 2536 & 72444 \\
        \bottomrule
    \end{tabular}
    \caption{The statistics of 5 Fashion-200K datasets.}
    \label{tab:fashion_mnist_data_stats}
\end{table}

\section{Hyper-parameters}\label{appx:hyperparams}
We summarize the hyper-parameters used to train Matryoshka-Adaptor. 
In all experiments (in both unsupervised and supervised settings), we utilize the fixed hyper-parameters that enable applying Matryoshka-Adaptor without extensive hyper-parameter tuning.

\begin{table}[h!]
    \small
    \centering
    \begin{tabular}{p{50mm}|l}
    \toprule
        Hyper-parameters & Fixed values \\
        \midrule
        Pairwise loss coefficient ($\alpha$) & $1.0$\\
        Recovery loss coefficient ($\beta$) & $1.0$\\
        Ranking loss coefficient ($\gamma$) & $1.0$\\
        Batch size for training & 128 \\
        Batch size for corpus during training & 50000 \\
        Maximum number of training iterations & 5000 \\
        Patience for early stopping & 500 \\
        Learning rates & 0.001 \\
        Optimizer & Adam \\
    \bottomrule
    \end{tabular}
    \caption{Hyper-parameters used to train Matryoshka-Adaptor in all experiments.}
    \label{tab:hyperparams}
\end{table}

\section{Computational complexity}
All experimental procedures were conducted utilizing a single NVIDIA V100 GPU with 16GB of memory. 
In the unsupervised Matryoshka-Adaptor training regime, processing time did not exceed 10 minutes for datasets ranging from 3,000 to 10 million corpus samples. 
For supervised Matryoshka-Adaptor training, datasets with fewer than 1 million corpus samples were processed in under 30 minutes, while datasets with up to 10 million corpus samples were processed in less than one hour. 
It is important to note that the adaptor architecture consists of a shallow multi-layer perceptron, rendering the computational complexity of inference negligible.

\section{Additional results on supervised Matryoshka-Adaptor}\label{appx:additional_supervised results}
\begin{figure*}[h!]
\centering
\subfloat[Google gecko-latest embeddings]{
\includegraphics[width=0.4\textwidth]{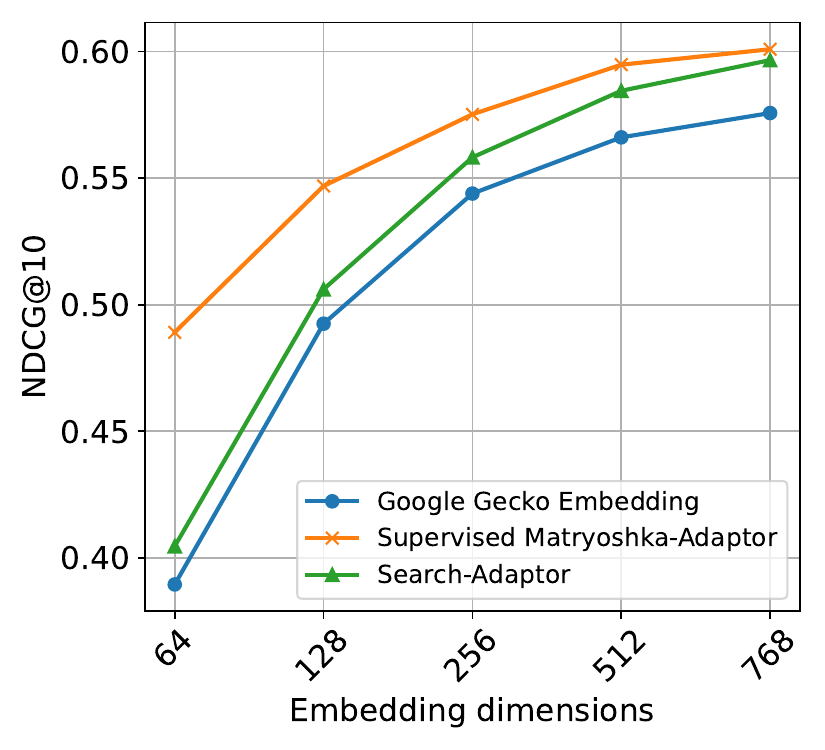}
\label{fig:beir_gecko_supervised}}
\subfloat[OpenAI text-embedding-3-small]{
\includegraphics[width=0.4\textwidth]{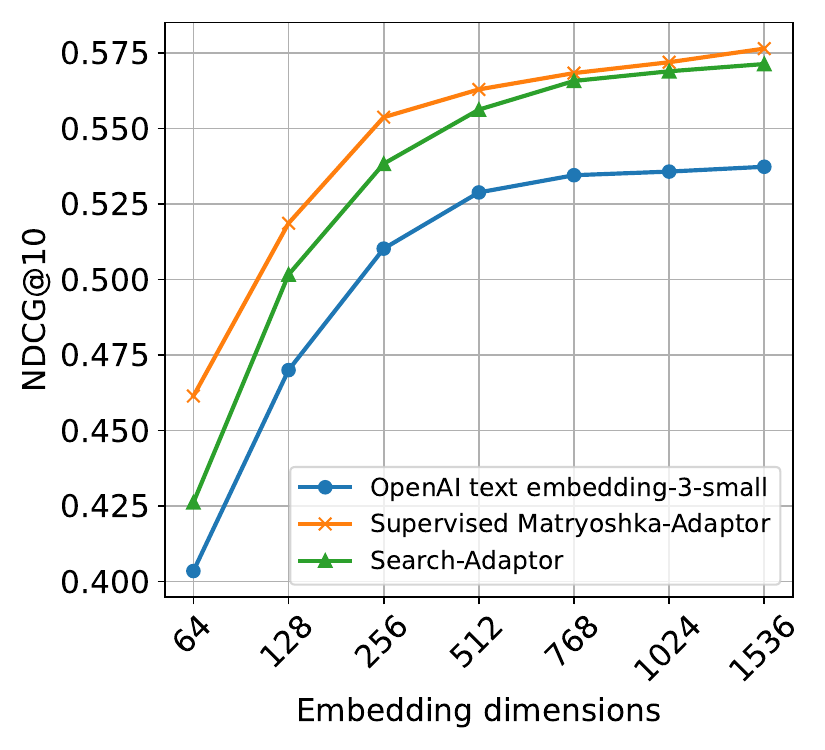}
\label{fig:beir_openai_small_supervised}}
\caption{Experimental results of the supervised Matryoshka-Adaptor on retrieval tasks, utilizing two different embedding models: Google Gecko (on 13 BEIR datasets), and OpenAI text-embedding-3-small (on 8 BEIR datasets).}
\label{fig:beir_supervised_additional}
\end{figure*}

\newpage
\section{Detailed Experimental Results on Unsupervised Settings}\label{appendix:unsupervised_detail_result}
In this section, we present the experimental results per each dataset in unsupervised settings.
In the main manuscript, we report the average values among the entire datasets.
\subsection{Unsupervised Matryoshka-Adaptor with OpenAI embedding models}
\begin{figure*}[h!]
\subfloat[NFCorpus]{
\includegraphics[width=0.24\textwidth]{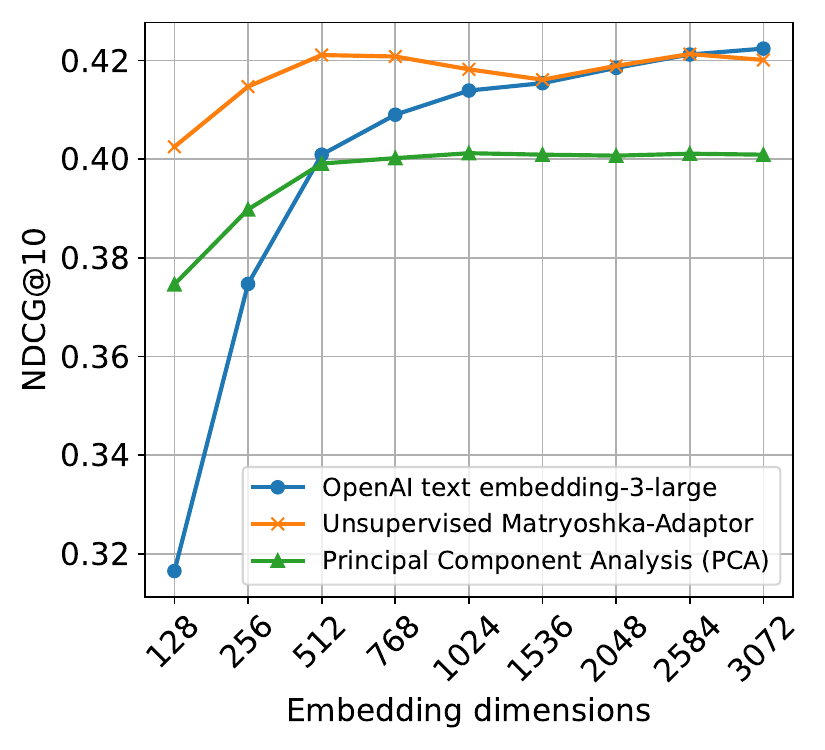}}
\subfloat[SciFact]{
\includegraphics[width=0.24\textwidth]{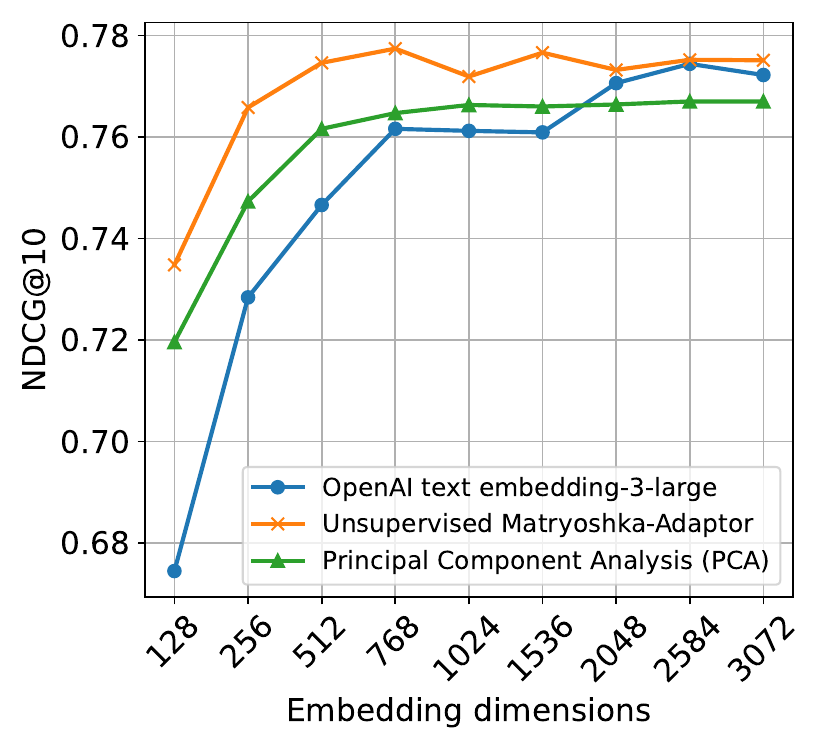}}
\subfloat[Arguana]{
\includegraphics[width=0.24\textwidth]{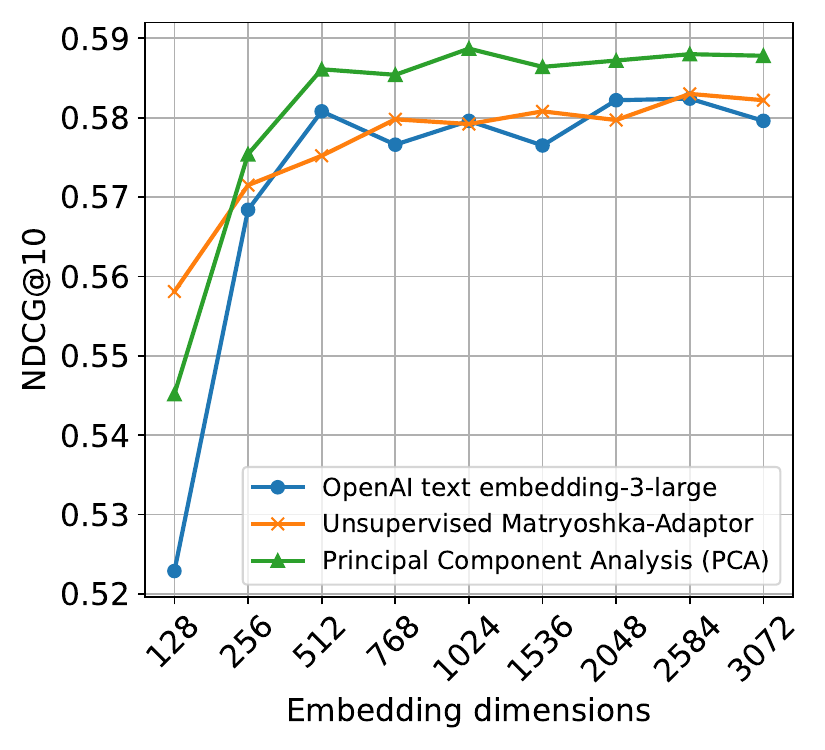}}
\subfloat[SciDocs]{
\includegraphics[width=0.24\textwidth]{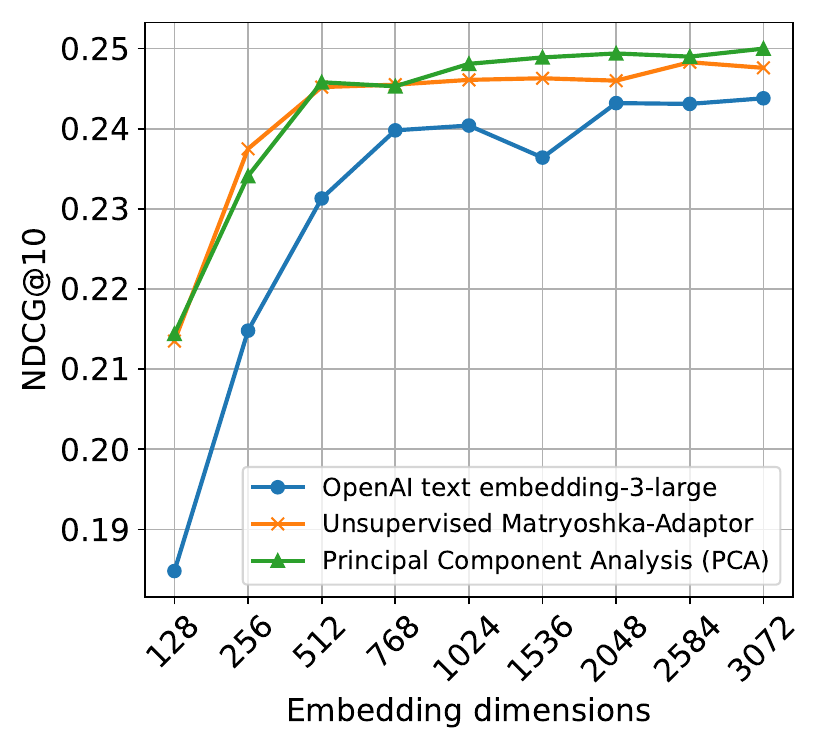}}\\
\subfloat[Fiqa]{
\includegraphics[width=0.24\textwidth]{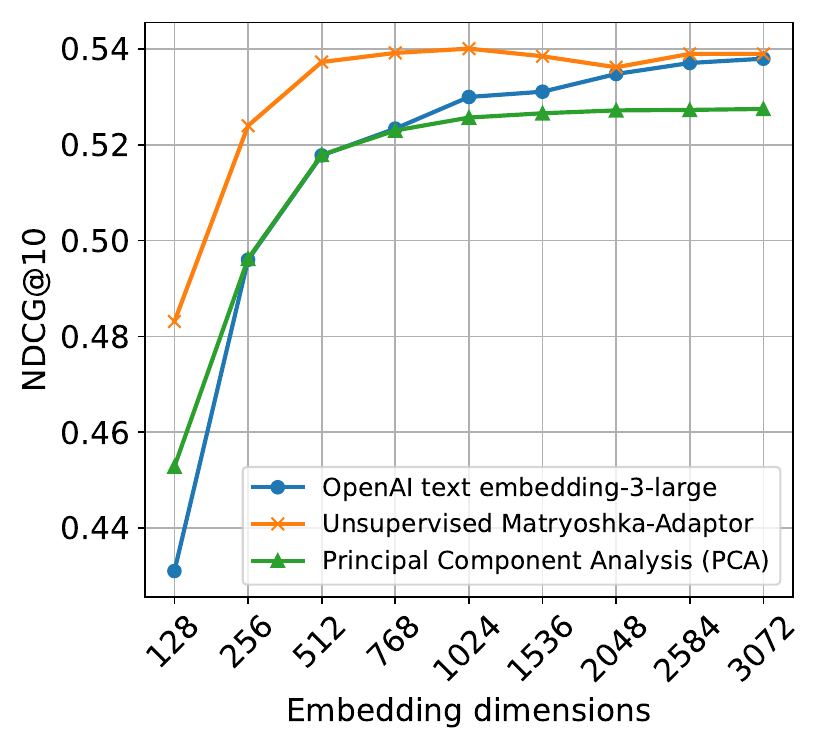}}
\subfloat[Trec-Covid]{
\includegraphics[width=0.24\textwidth]{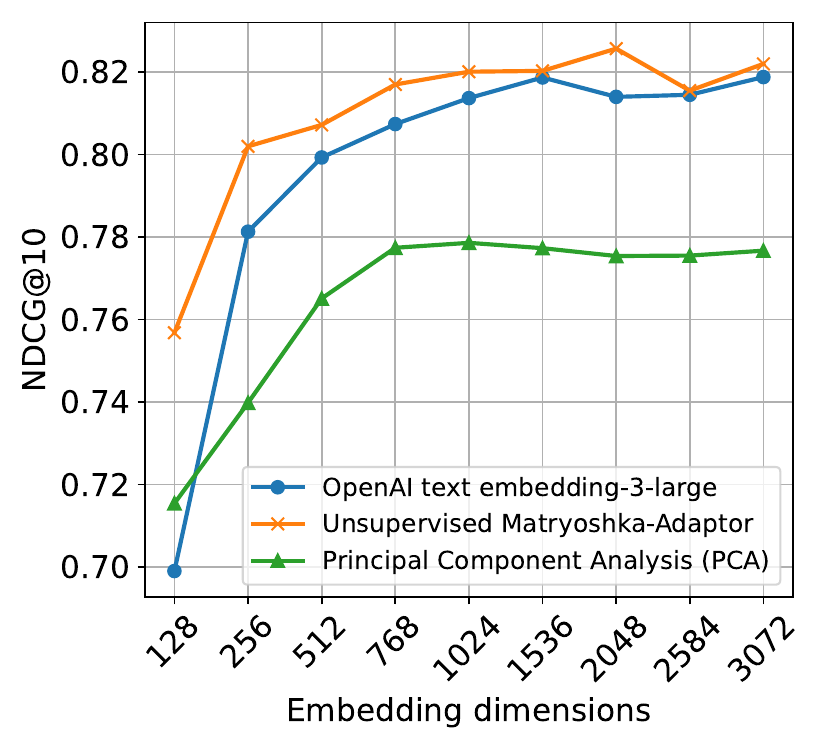}}
\subfloat[Touche]{
\includegraphics[width=0.24\textwidth]{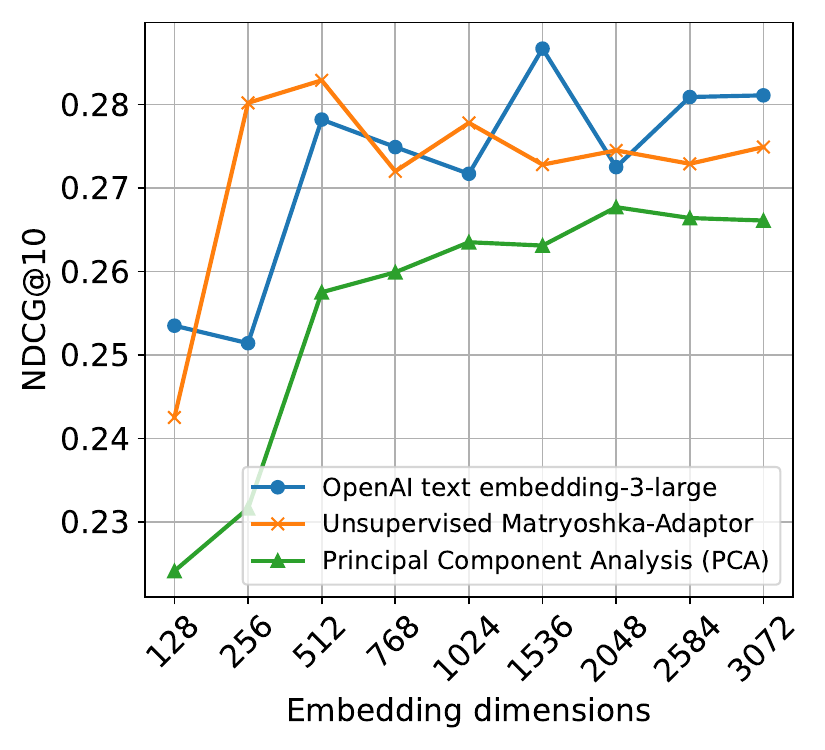}}
\subfloat[Quora]{
\includegraphics[width=0.24\textwidth]{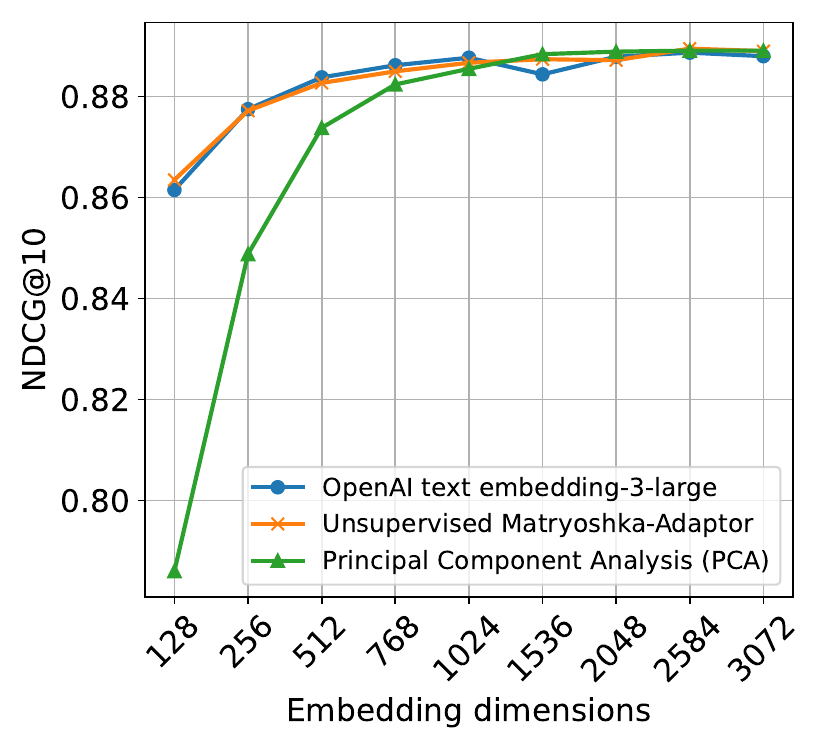}}\\
\caption{Experimental results of unsupervised Matryoshka-Adaptor with OpenAI-textembedding-3-large on 8 BEIR datasets.}
\label{fig:each_openai_large_beir}
\end{figure*}

\begin{figure*}[h!]
\subfloat[NFCorpus]{
\includegraphics[width=0.24\textwidth]{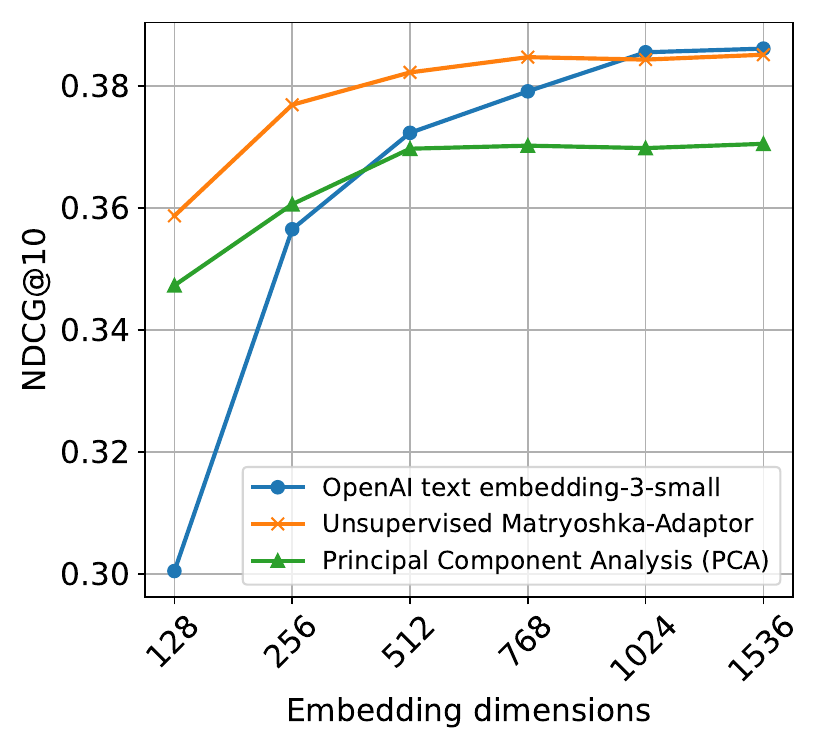}}
\subfloat[SciFact]{
\includegraphics[width=0.24\textwidth]{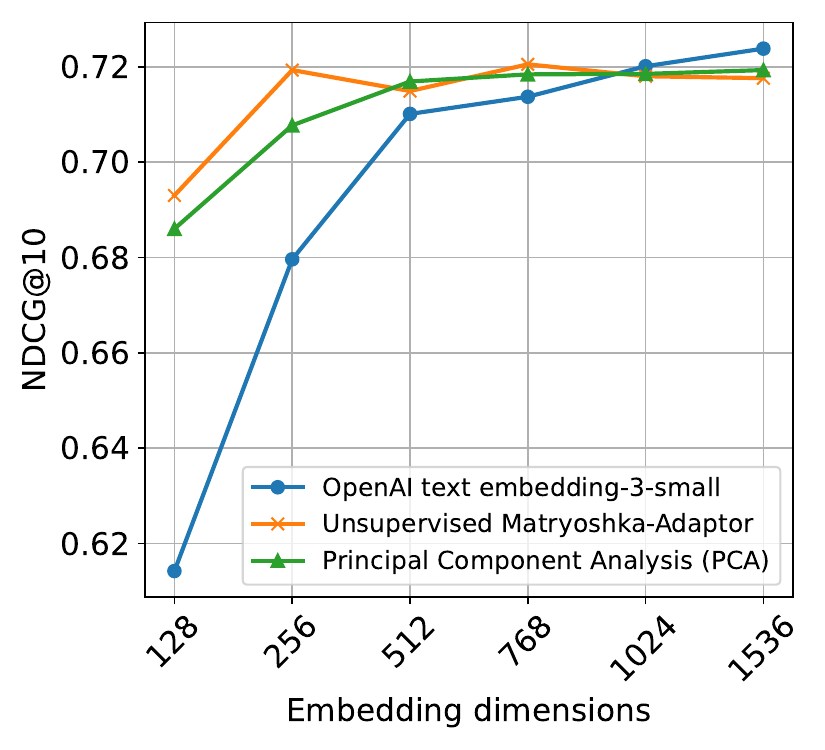}}
\subfloat[Arguana]{
\includegraphics[width=0.24\textwidth]{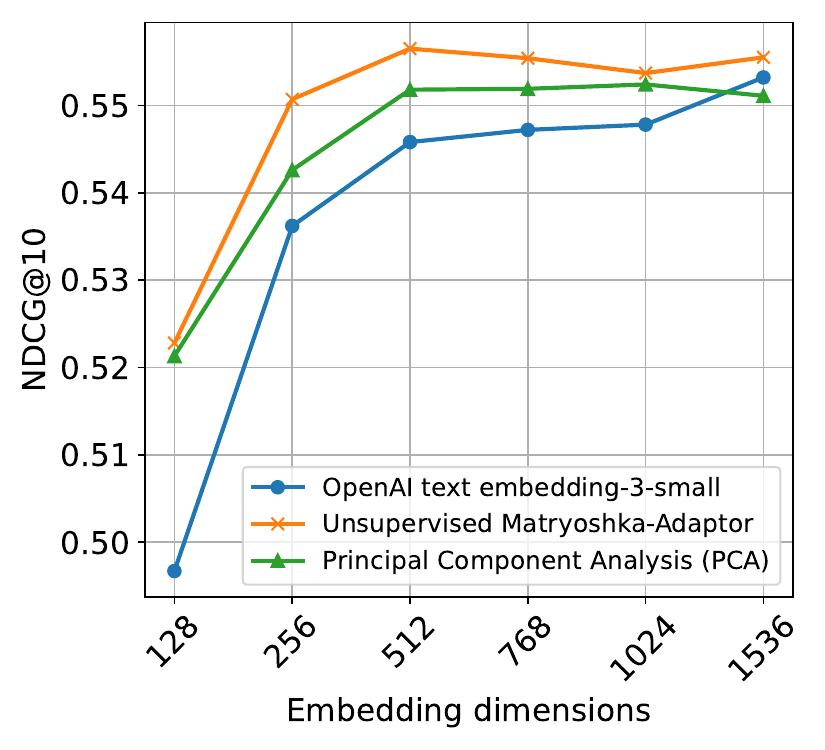}}
\subfloat[SciDocs]{
\includegraphics[width=0.24\textwidth]{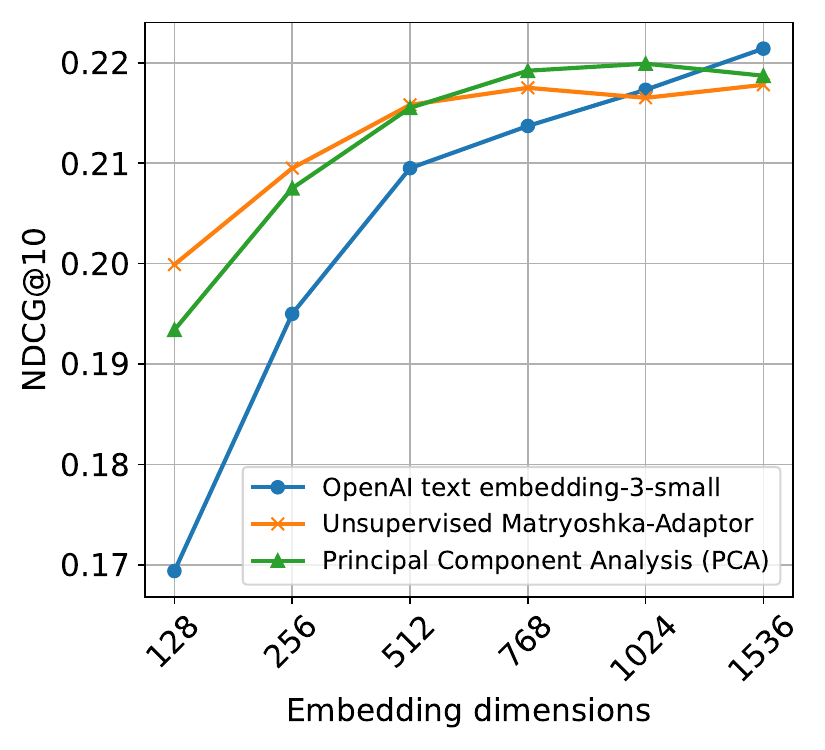}}\\
\subfloat[Fiqa]{
\includegraphics[width=0.24\textwidth]{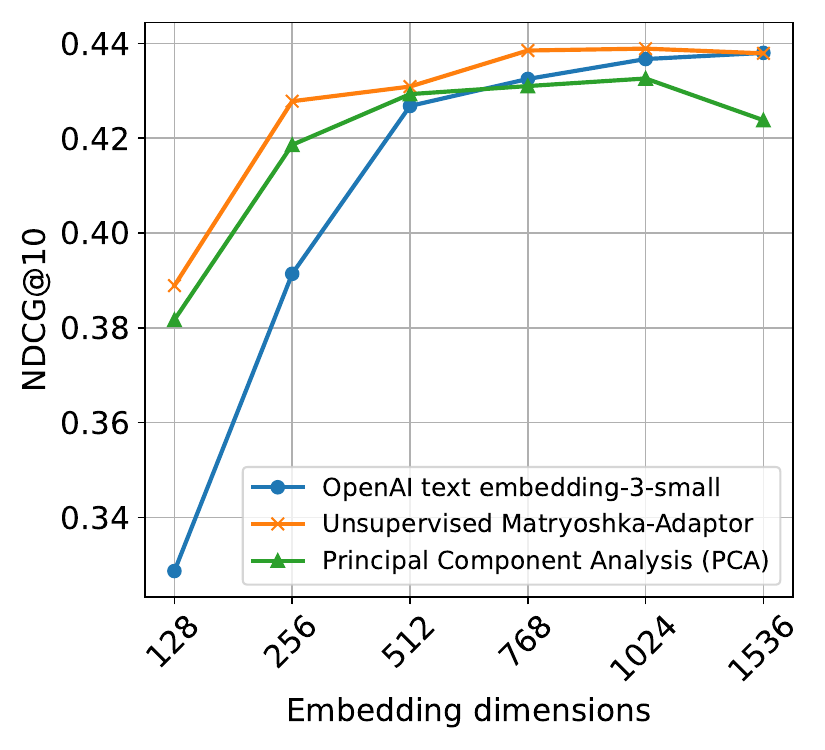}}
\subfloat[Trec-Covid]{
\includegraphics[width=0.24\textwidth]{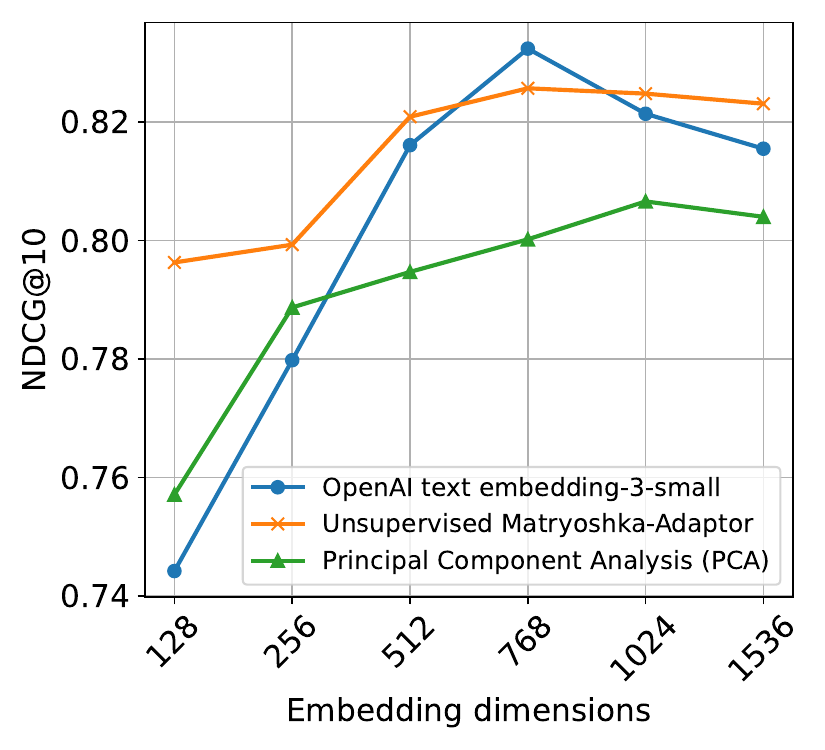}}
\subfloat[Touche]{
\includegraphics[width=0.24\textwidth]{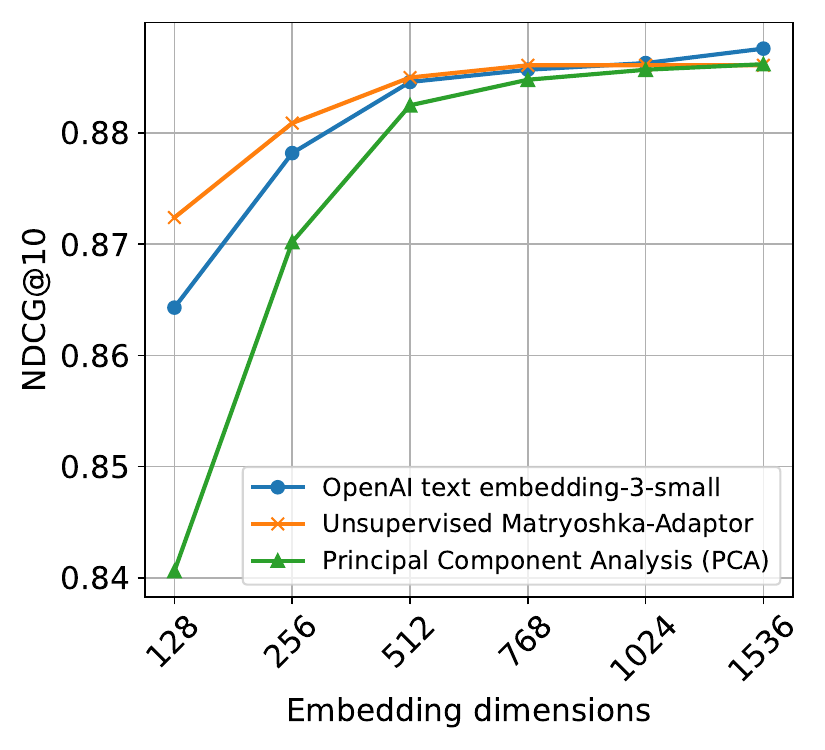}}
\subfloat[Quora]{
\includegraphics[width=0.24\textwidth]{result/each/openai_large_beir/openai_large_beir_quora.pdf}}\\
\caption{Experimental results of unsupervised Matryoshka-Adaptor with OpenAI-textembedding-3-small on 8 BEIR datasets.}
\label{fig:each_openai_small_beir}
\end{figure*}

\newpage
\subsection{Unsupervised Matryoshka-Adaptor with Google multimodal embedding models}

\begin{figure*}[h!]
\subfloat[Dresses]{
\includegraphics[width=0.24\textwidth]{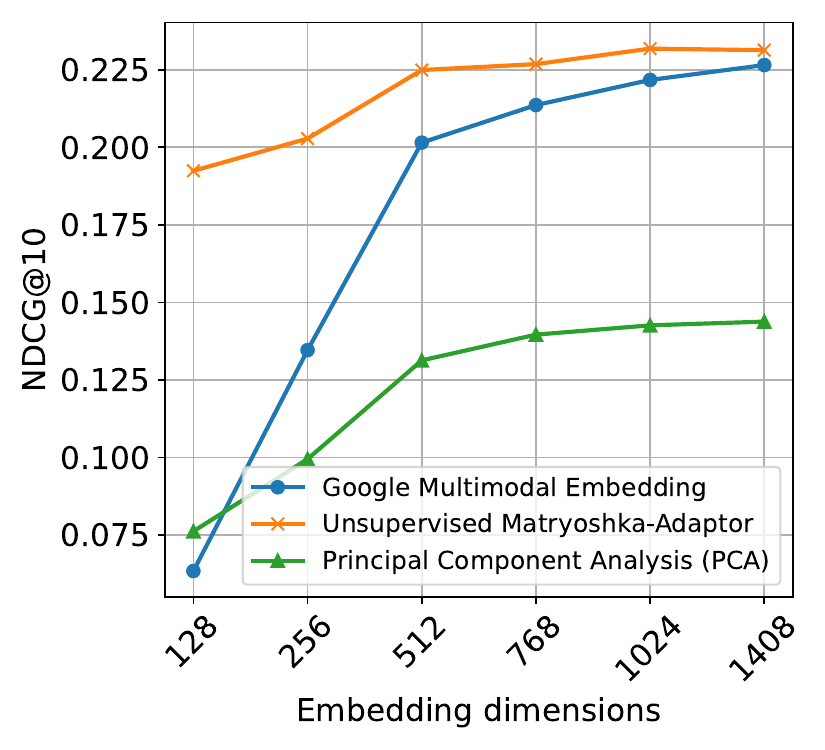}}
\subfloat[Jackets]{
\includegraphics[width=0.24\textwidth]{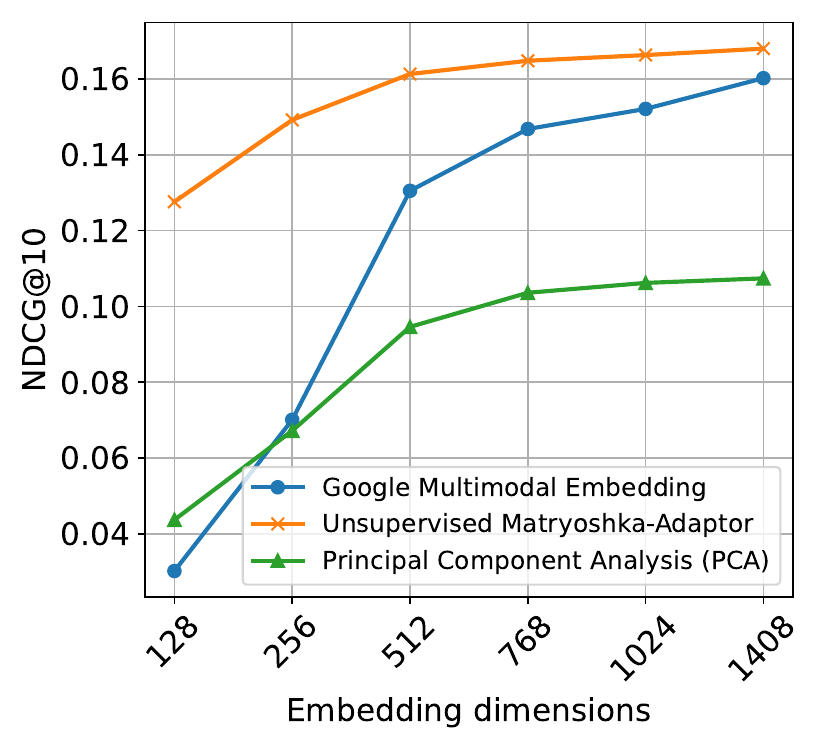}}
\subfloat[Pants]{
\includegraphics[width=0.24\textwidth]{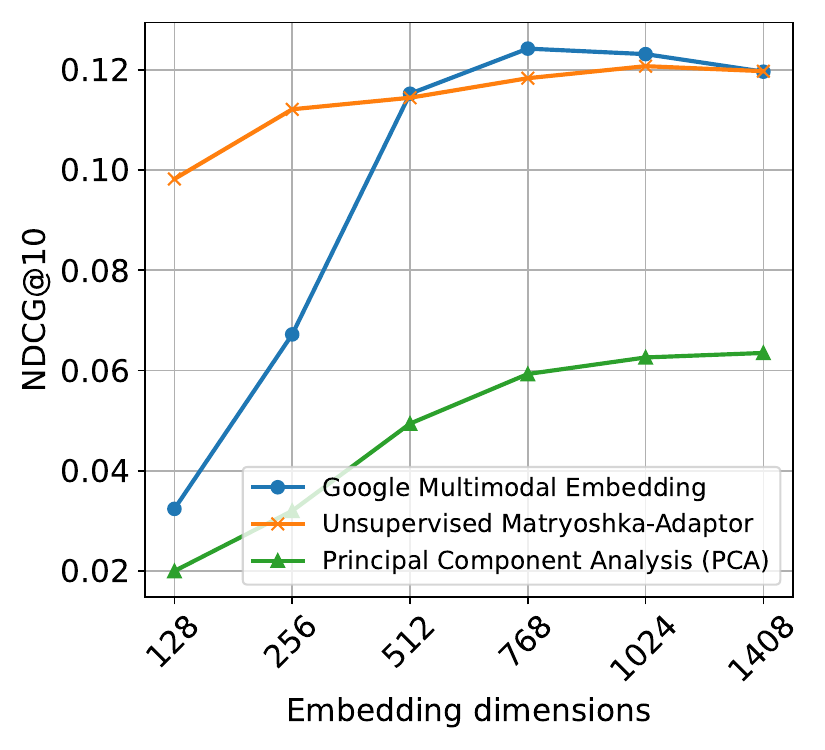}}
\subfloat[Skirts]{
\includegraphics[width=0.24\textwidth]{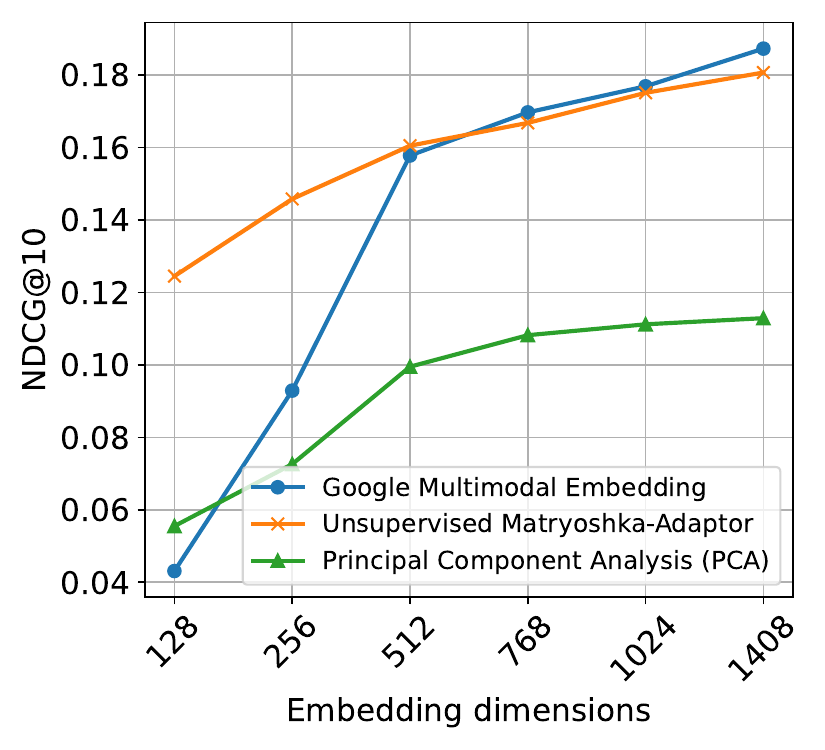}}\\
\subfloat[Tops]{
\includegraphics[width=0.24\textwidth]{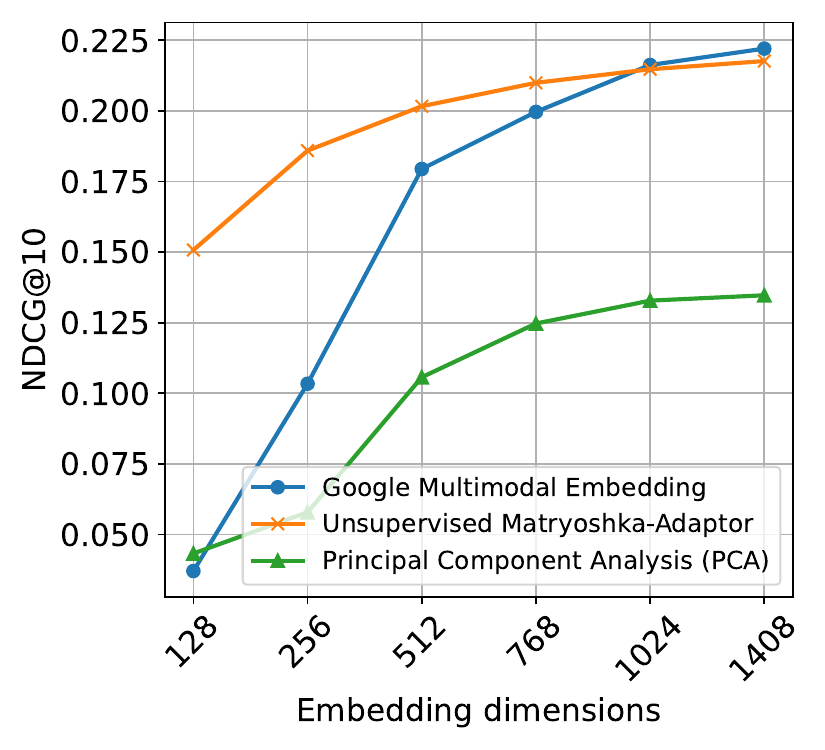}}
\caption{Experimental results of unsupervised Matryoshka-Adaptor with Google multimodal embedding models on 5 Fashion-200K datasets.}
\label{fig:each_google_multimodal_fashion}
\end{figure*}

\newpage
\subsection{Unsupervised Matryoshka-Adaptor with Google Gecko multilingual embedding models}

\begin{figure*}[h!]
\subfloat[Yoruba (yo)]{
\includegraphics[width=0.24\textwidth]{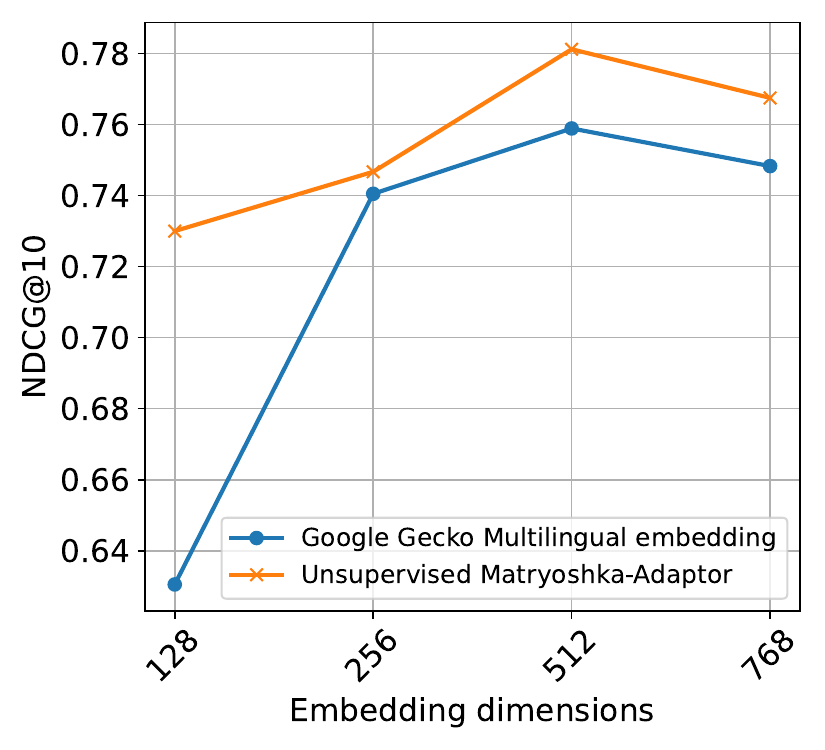}}
\subfloat[Swahilli (sw)]{
\includegraphics[width=0.24\textwidth]{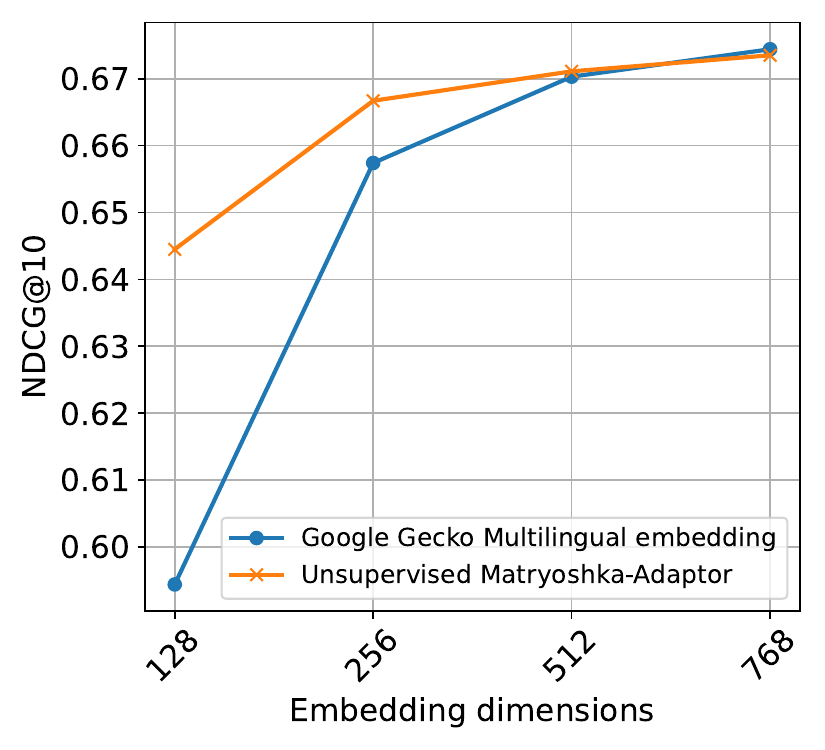}}
\subfloat[Bengali (bn)]{
\includegraphics[width=0.24\textwidth]{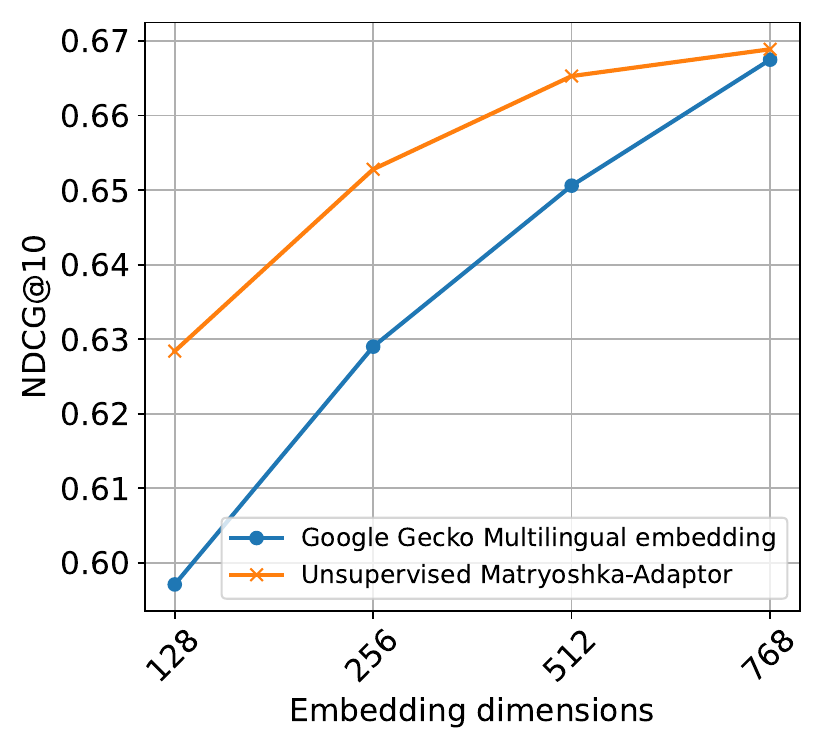}}
\subfloat[Hindi (hi)]{
\includegraphics[width=0.24\textwidth]{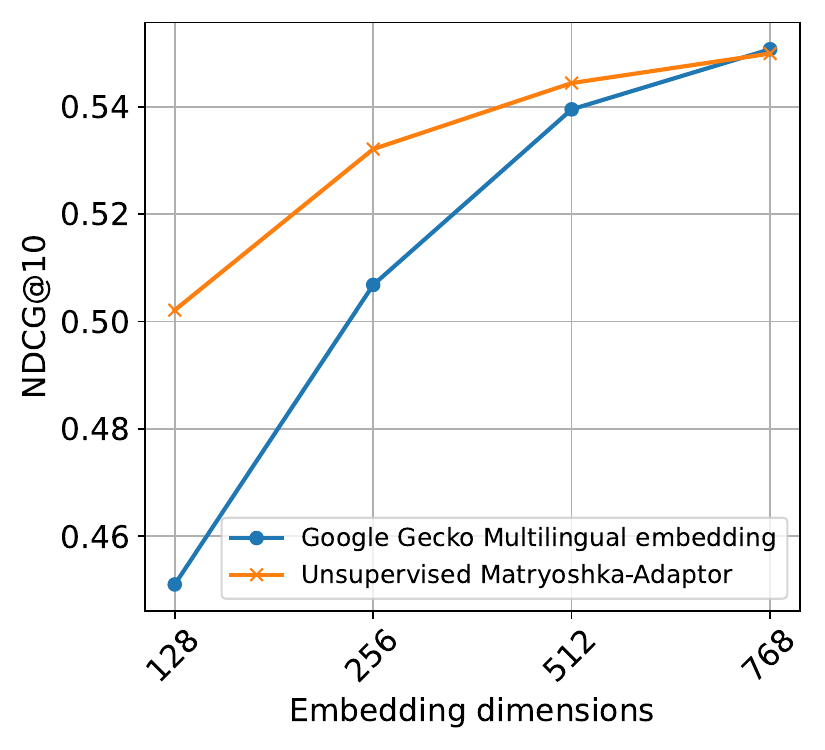}}\\
\subfloat[Telugu (te)]{
\includegraphics[width=0.24\textwidth]{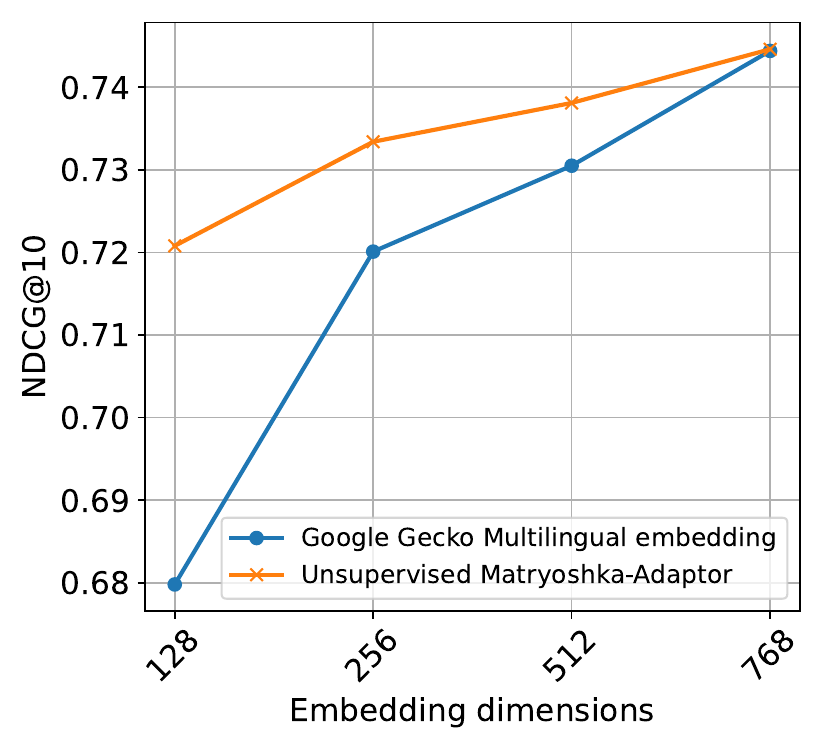}}
\subfloat[Thai (th)]{
\includegraphics[width=0.24\textwidth]{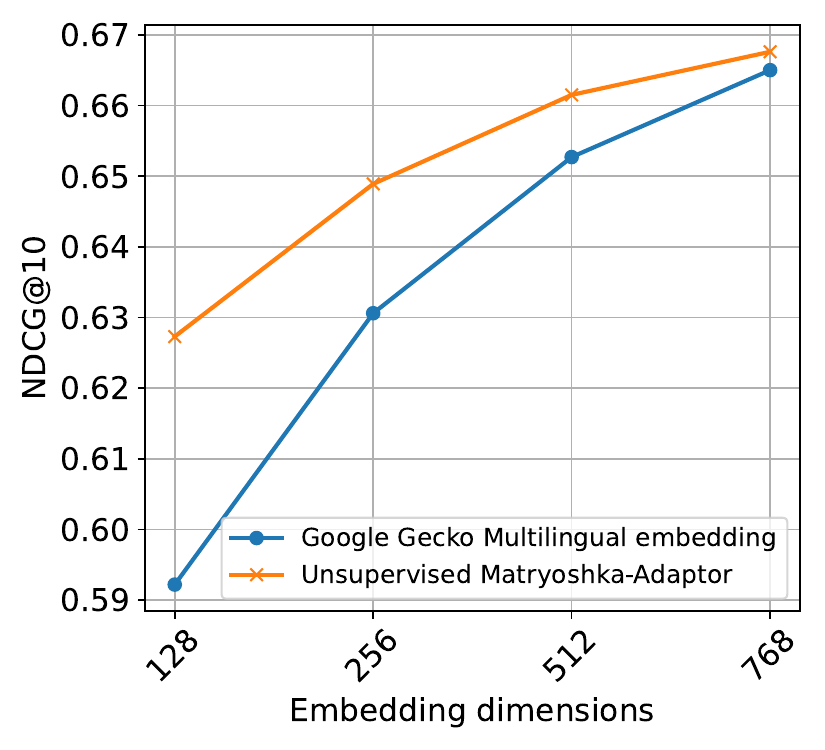}}
\subfloat[Indonesian (id)]{
\includegraphics[width=0.24\textwidth]{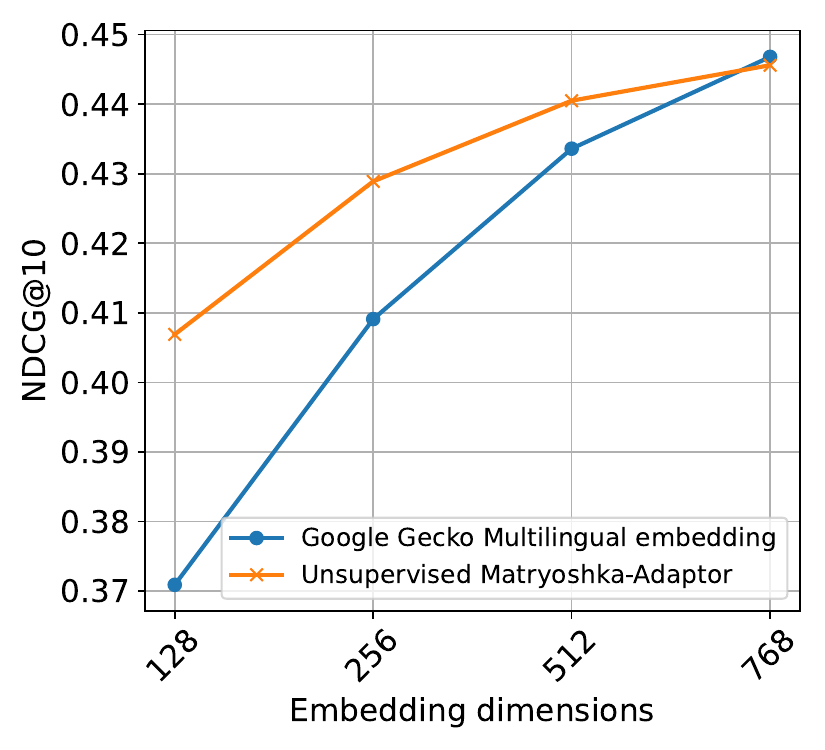}}
\subfloat[Korean (ko)]{
\includegraphics[width=0.24\textwidth]{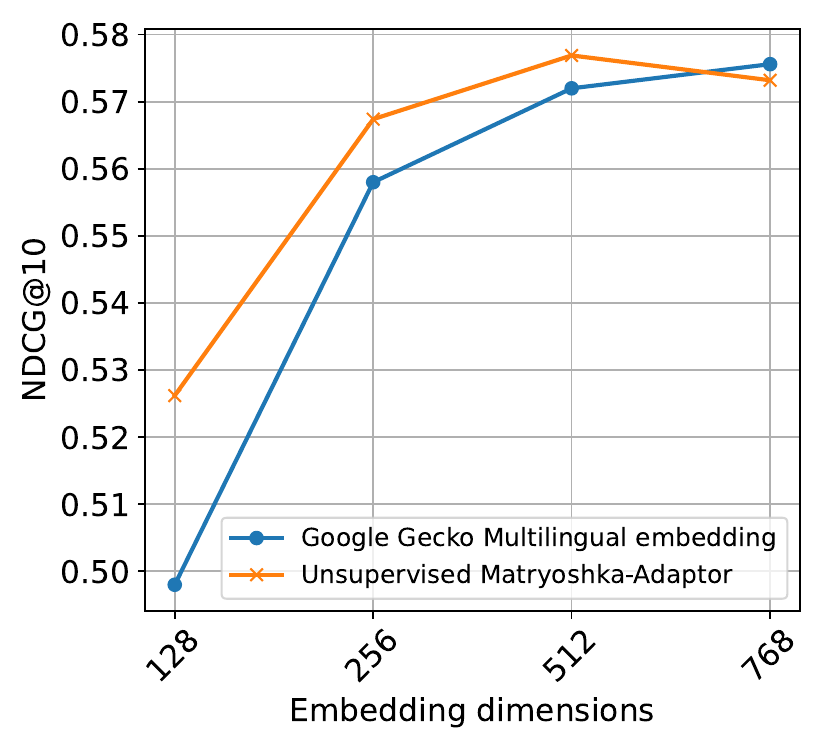}}\\
\subfloat[Finnish (fi)]{
\includegraphics[width=0.24\textwidth]{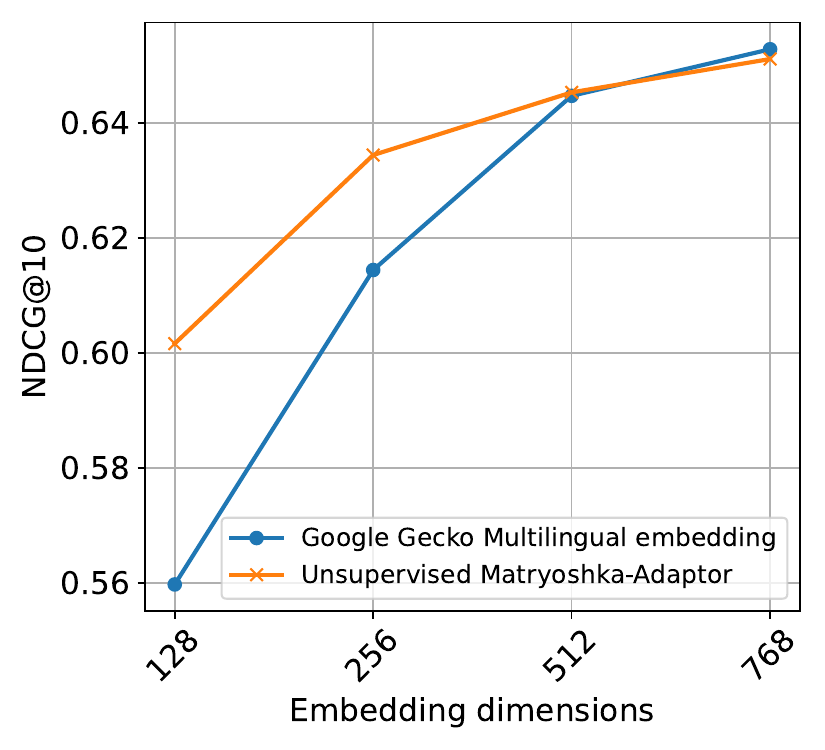}}
\subfloat[Arabic (ar)]{
\includegraphics[width=0.24\textwidth]{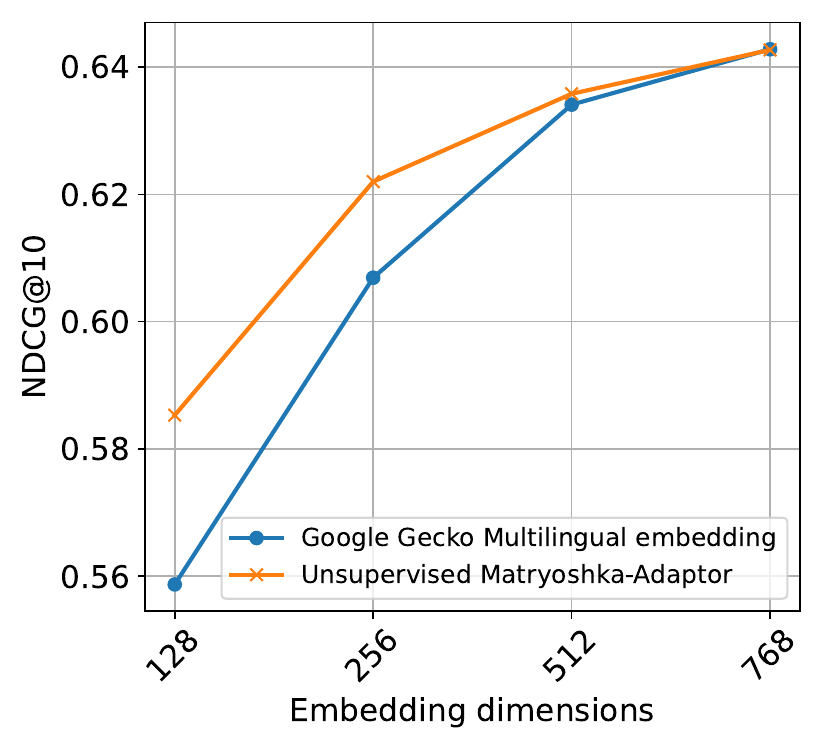}}
\subfloat[Persian (fa)]{
\includegraphics[width=0.24\textwidth]{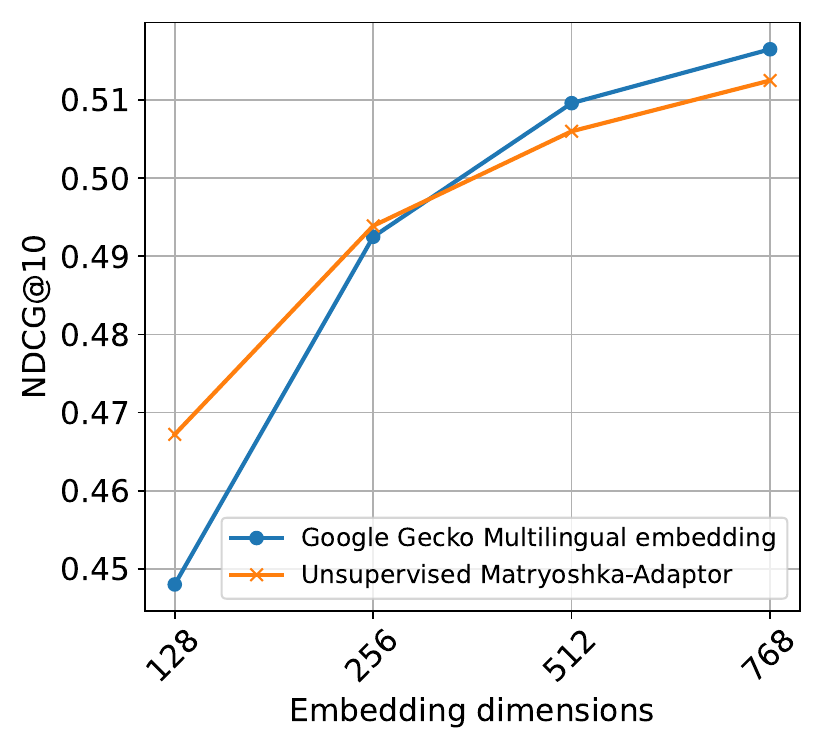}}
\subfloat[Chinese (zh)]{
\includegraphics[width=0.24\textwidth]{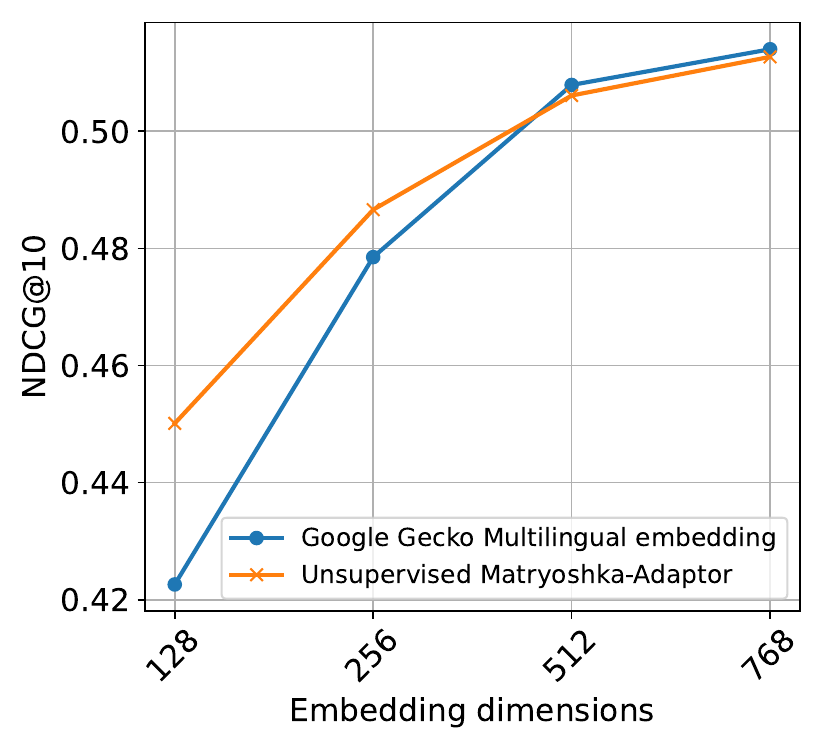}}\\
\subfloat[Japanese (ja)]{
\includegraphics[width=0.24\textwidth]{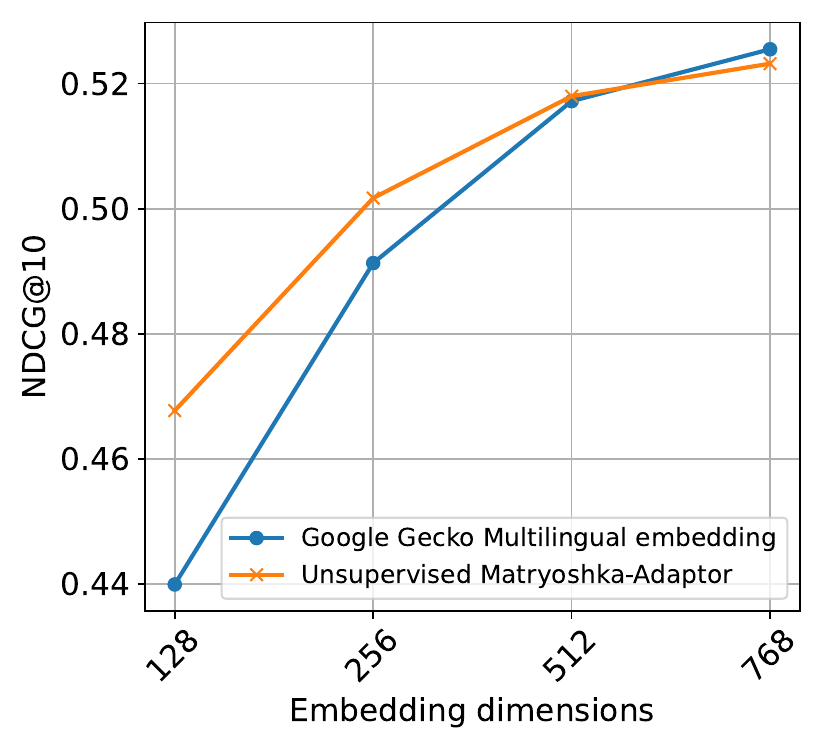}}
\subfloat[Russian (ru)]{
\includegraphics[width=0.24\textwidth]{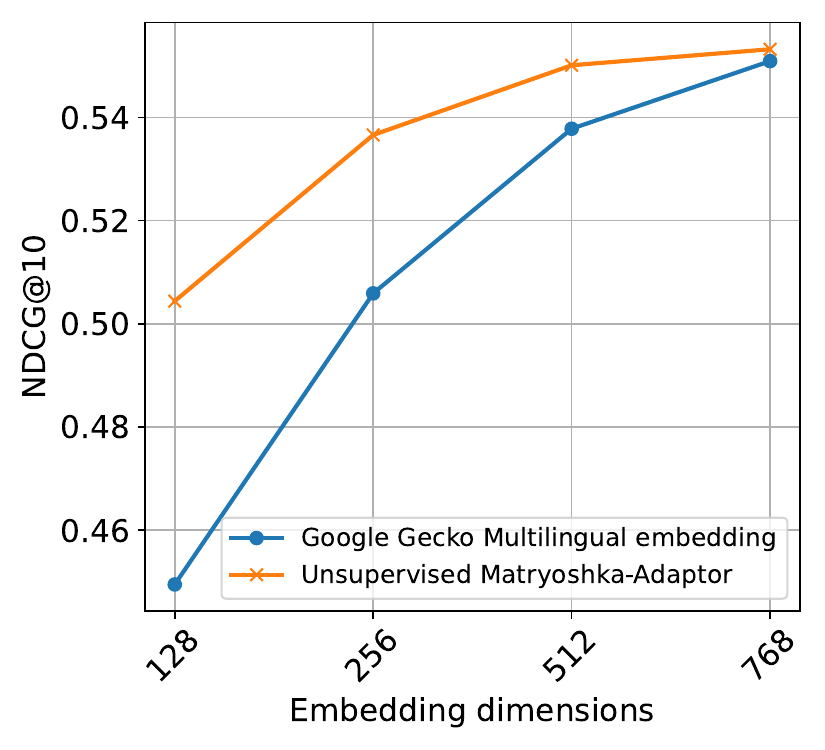}}
\subfloat[Spannish (es)]{
\includegraphics[width=0.24\textwidth]{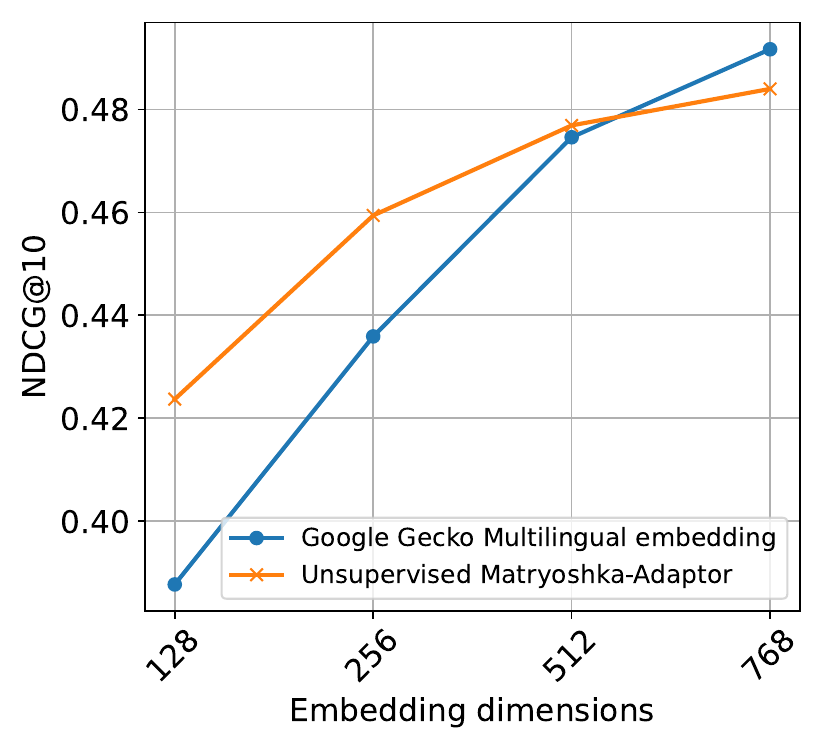}}
\subfloat[French (fr)]{
\includegraphics[width=0.24\textwidth]{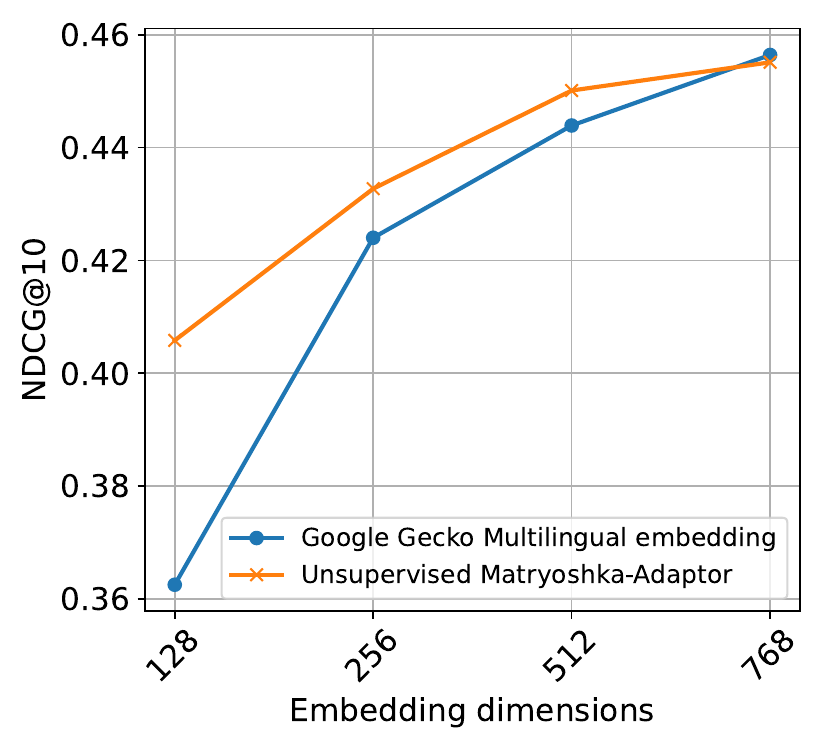}}\\
\subfloat[Germany (de)]{
\includegraphics[width=0.24\textwidth]{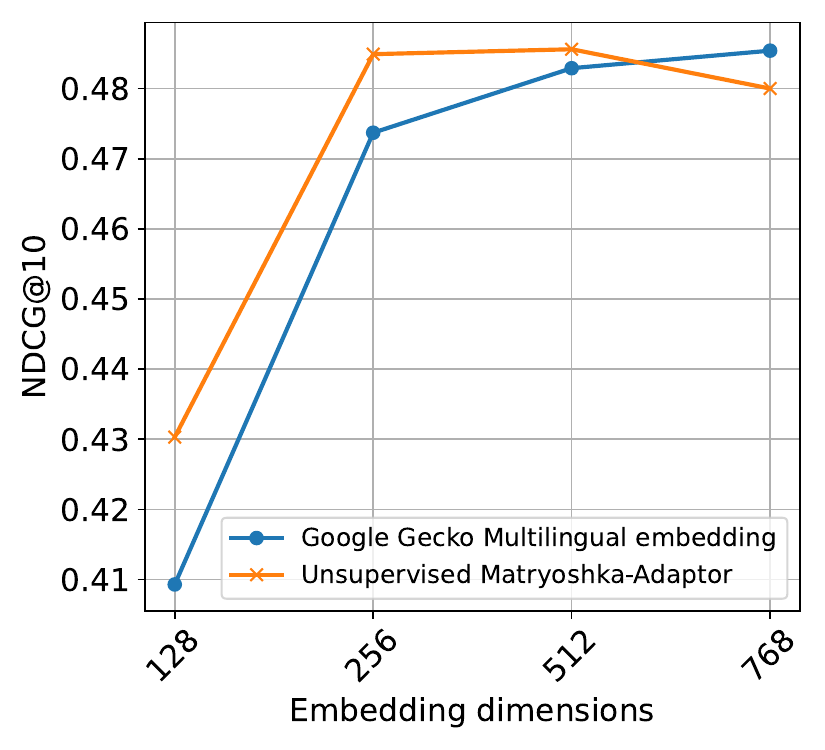}}\\
\caption{Experimental results of unsupervised Matryoshka-Adaptor with Google Gecko multilingual embedding models on 17 MIRACL datasets.}
\label{fig:each_google_multilingual_miracl}
\end{figure*}

\newpage
\section{Detailed Experimental Results on Supervised Settings}\label{appendix:supervised_detail_result}
In this section, we present the experimental results per each dataset in supervised settings.
In the main manuscript, we report the average values among the entire datasets.

\subsection{Supervised Matryoshka-Adaptor with Google Gecko embedding models}

\begin{figure*}[h!]
\subfloat[NFCorpus]{
\includegraphics[width=0.24\textwidth]{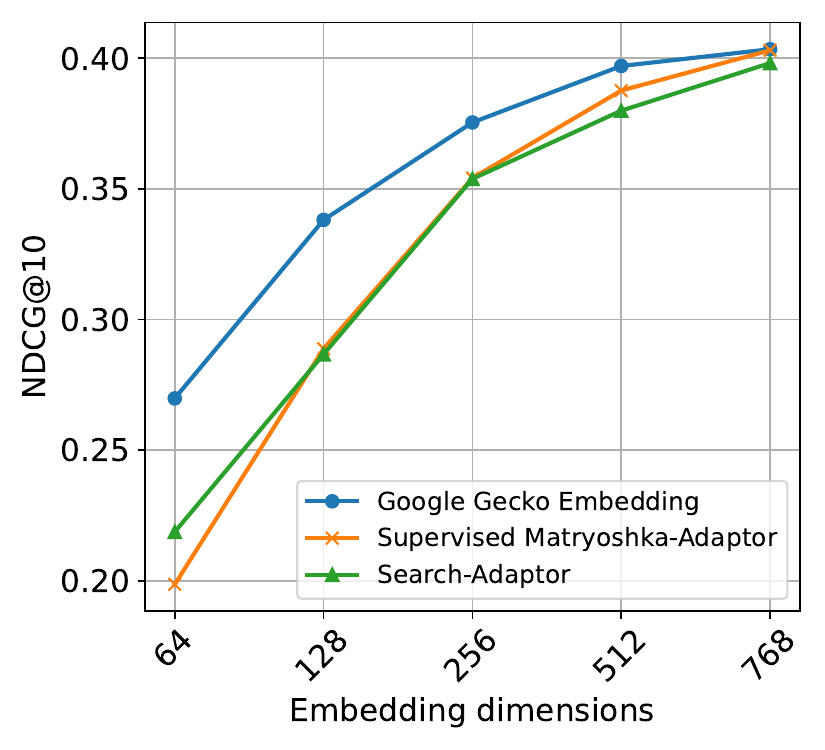}}
\subfloat[Scifact]{
\includegraphics[width=0.24\textwidth]{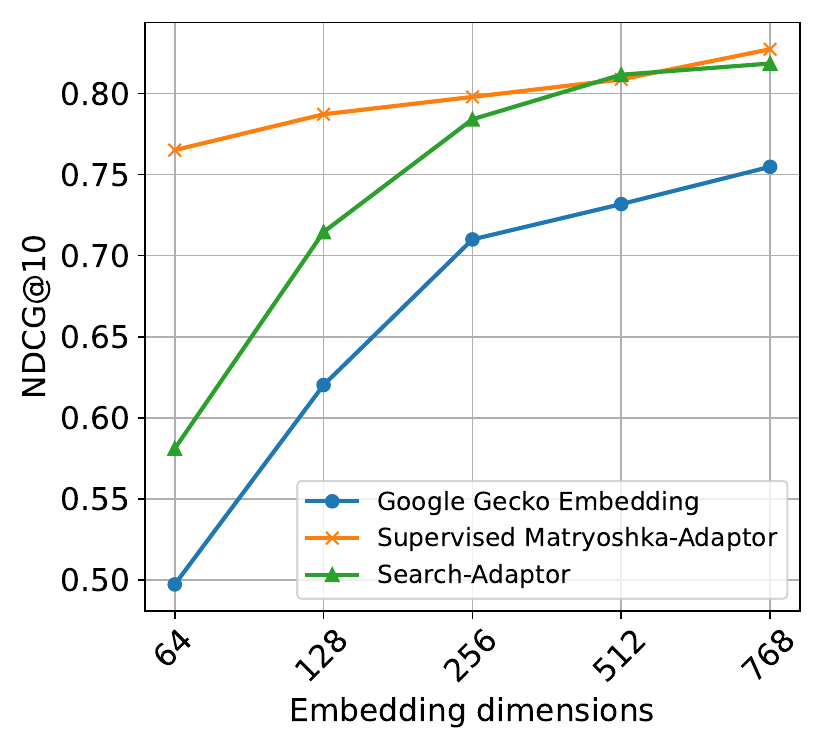}}
\subfloat[Arguana]{
\includegraphics[width=0.24\textwidth]{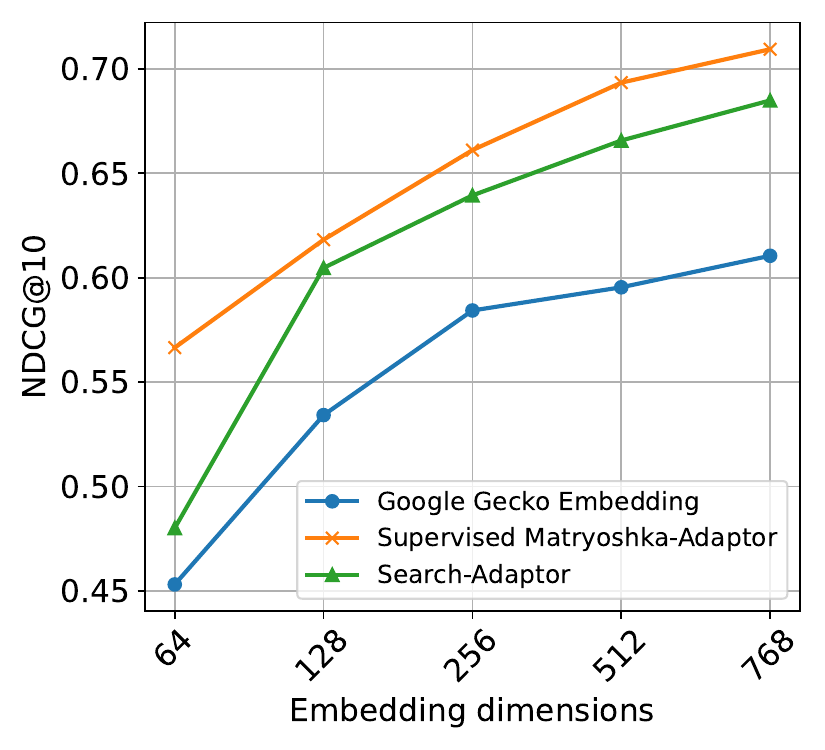}}
\subfloat[SciDocs]{
\includegraphics[width=0.24\textwidth]{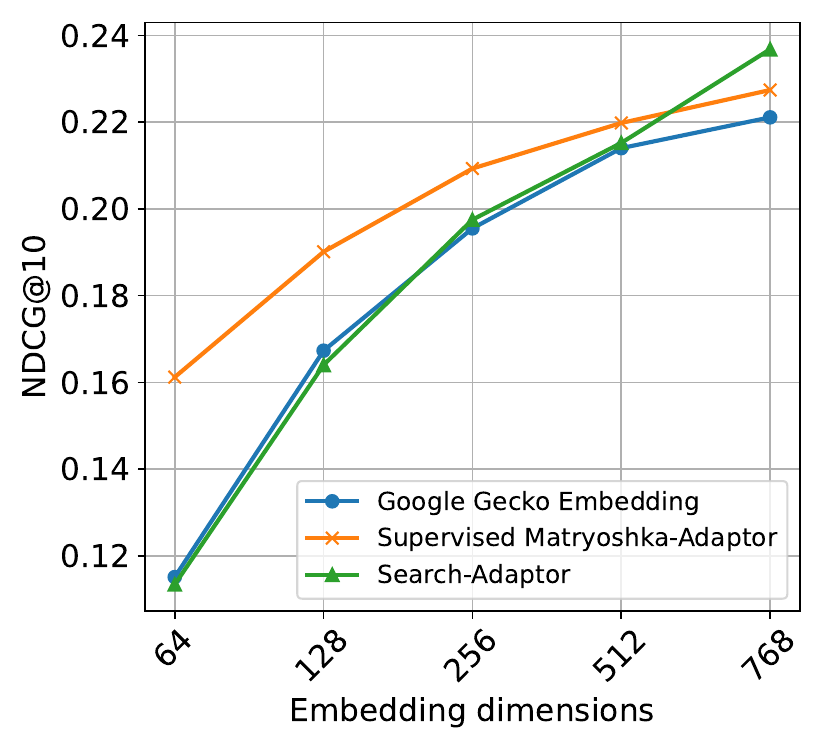}}\\
\subfloat[FiQA]{
\includegraphics[width=0.24\textwidth]{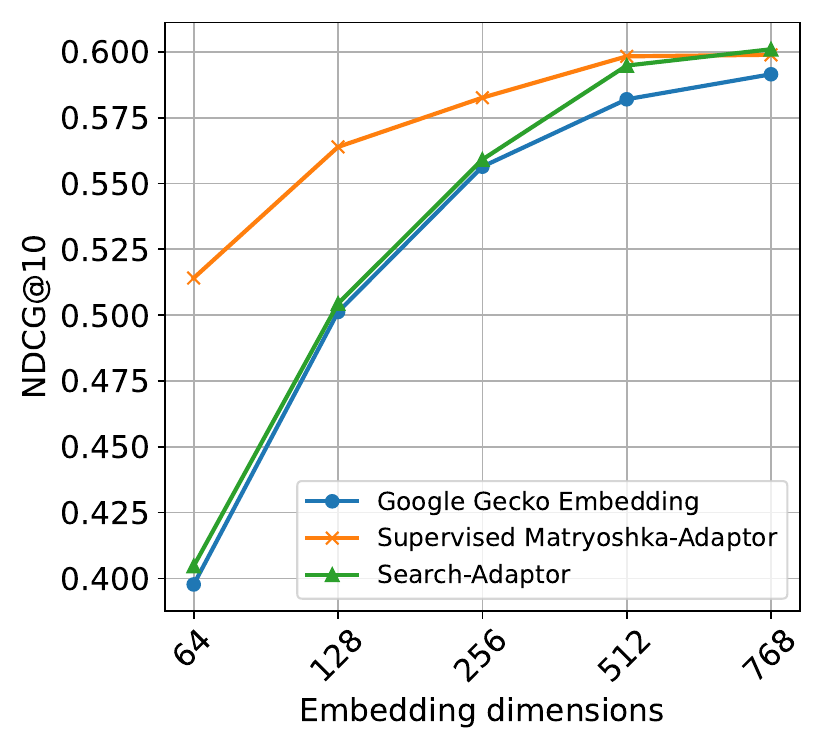}}
\subfloat[Trec-Covid]{
\includegraphics[width=0.24\textwidth]{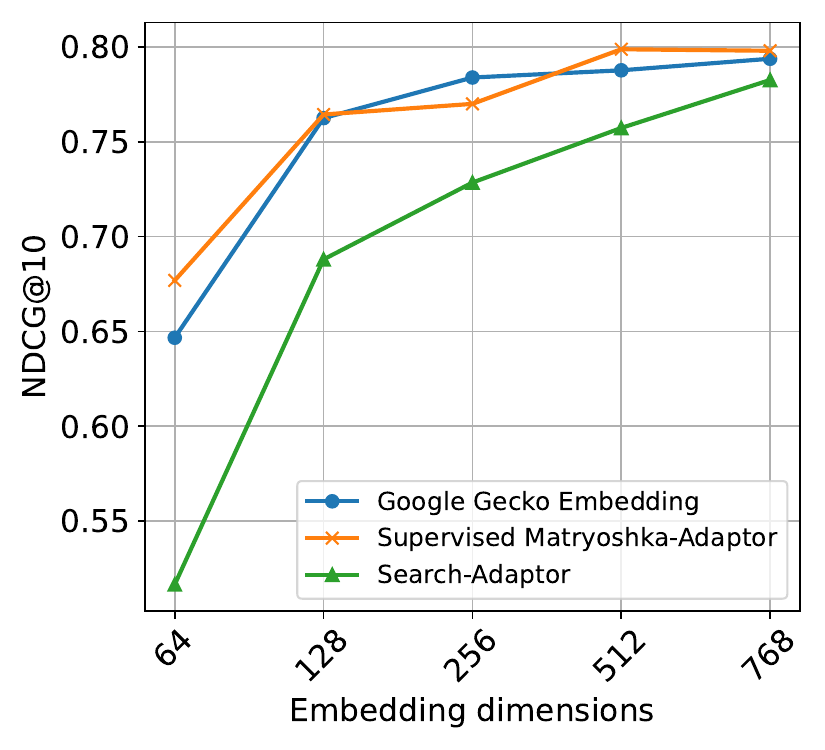}}
\subfloat[Touche]{
\includegraphics[width=0.24\textwidth]{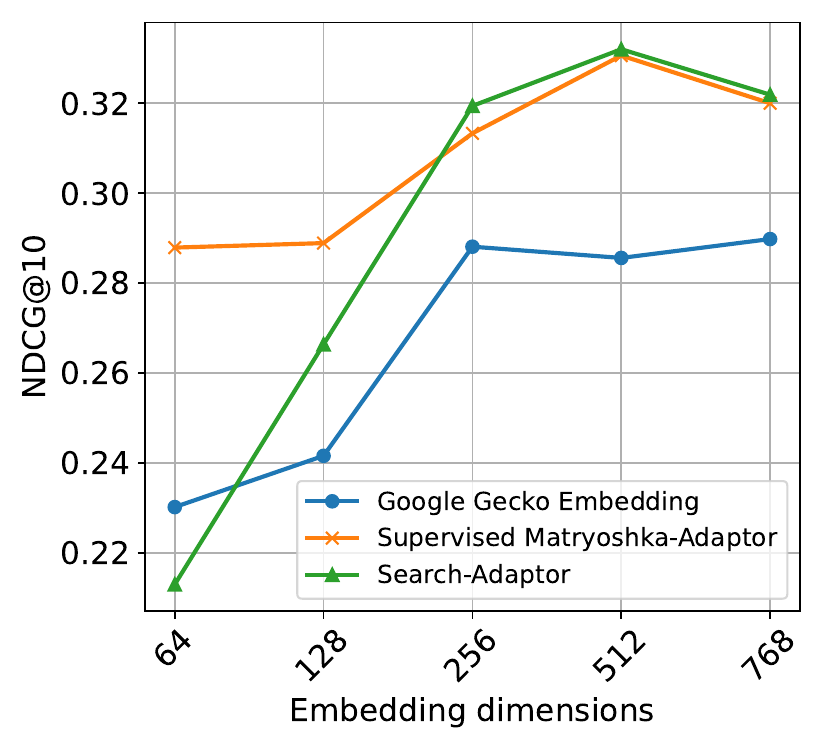}}
\subfloat[Quora]{
\includegraphics[width=0.24\textwidth]{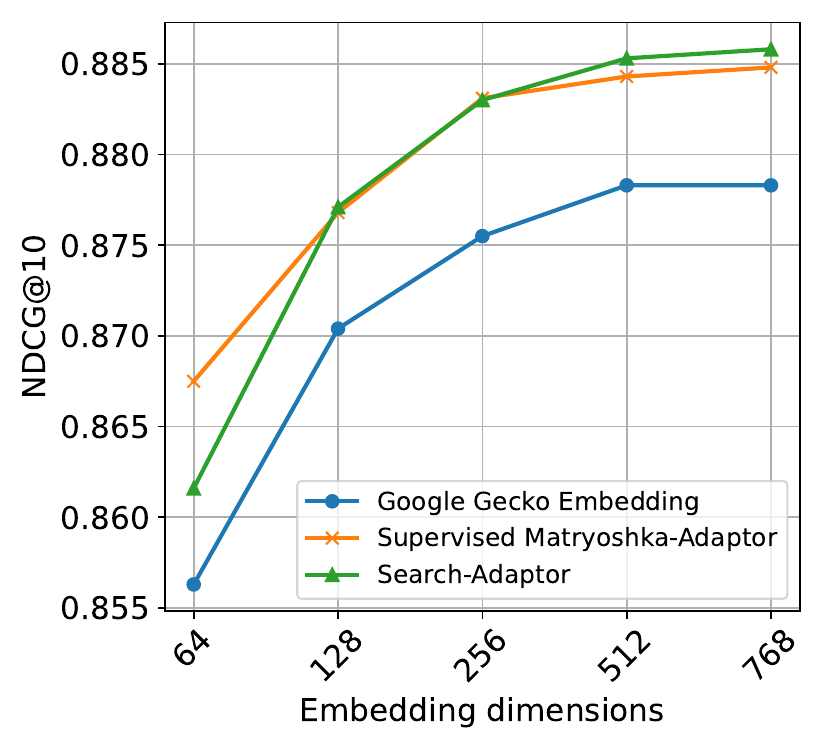}}\\
\subfloat[NQ]{
\includegraphics[width=0.24\textwidth]{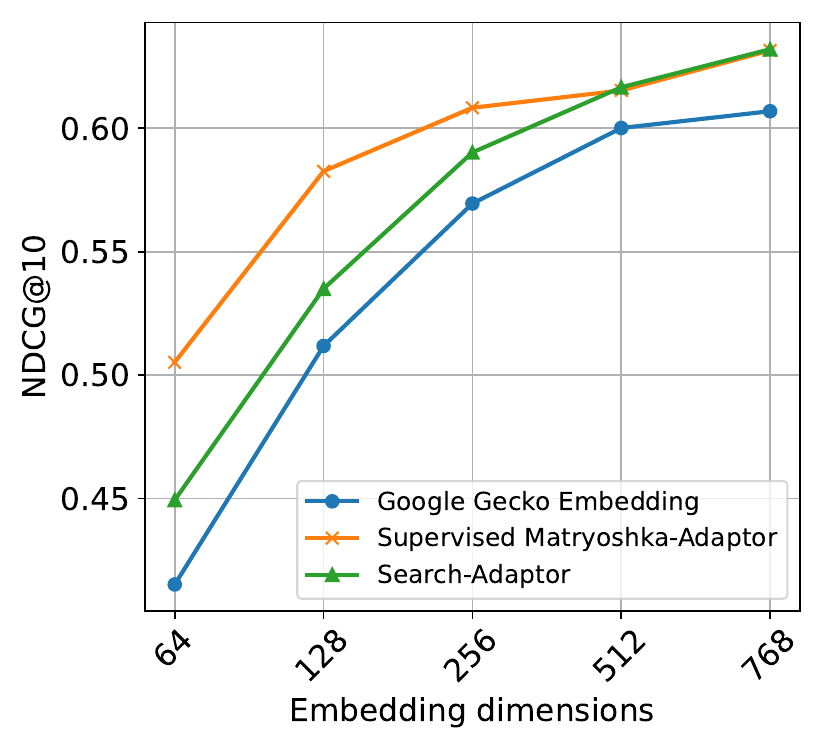}}
\subfloat[DBPedia]{
\includegraphics[width=0.24\textwidth]{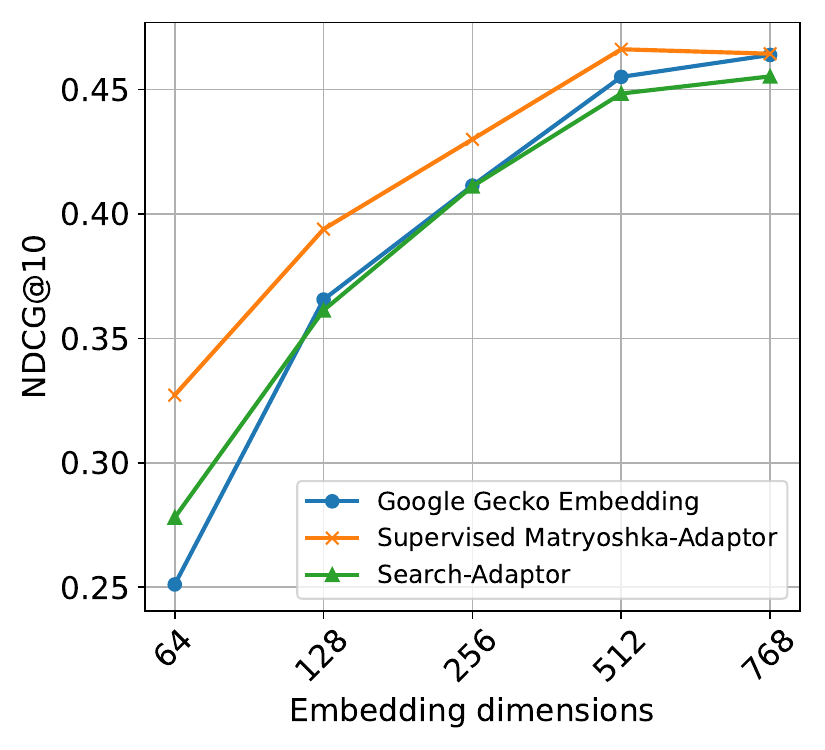}}
\subfloat[HotPotQA)]{
\includegraphics[width=0.24\textwidth]{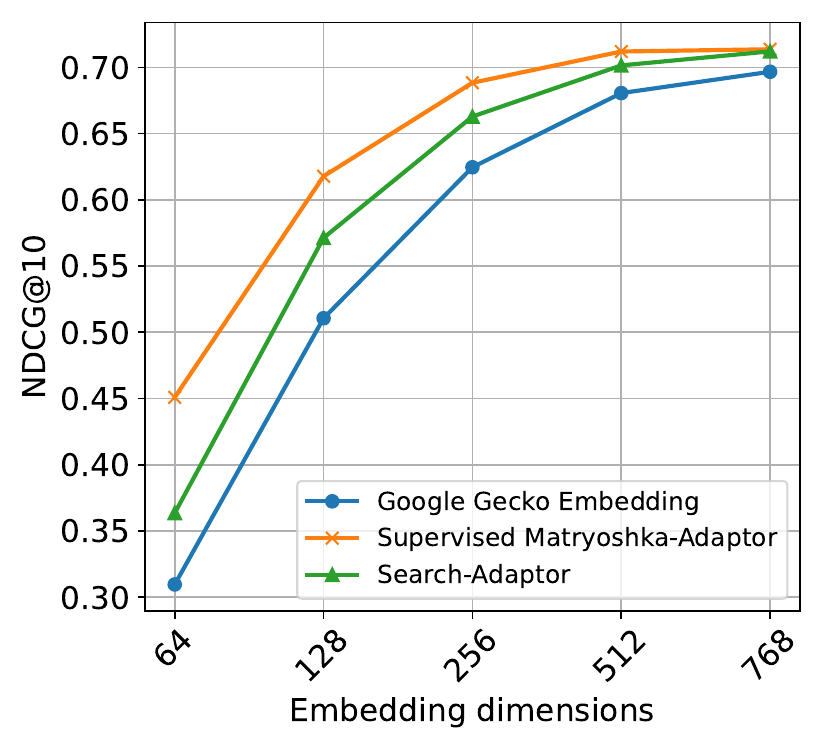}}
\subfloat[Fever]{
\includegraphics[width=0.24\textwidth]{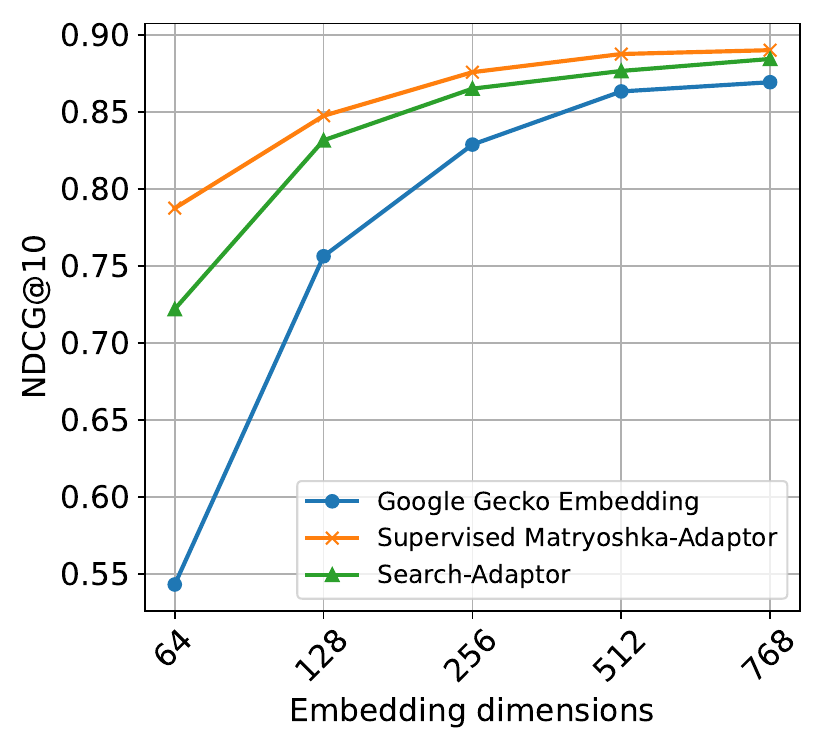}}\\
\subfloat[Climate-fever]{
\includegraphics[width=0.24\textwidth]{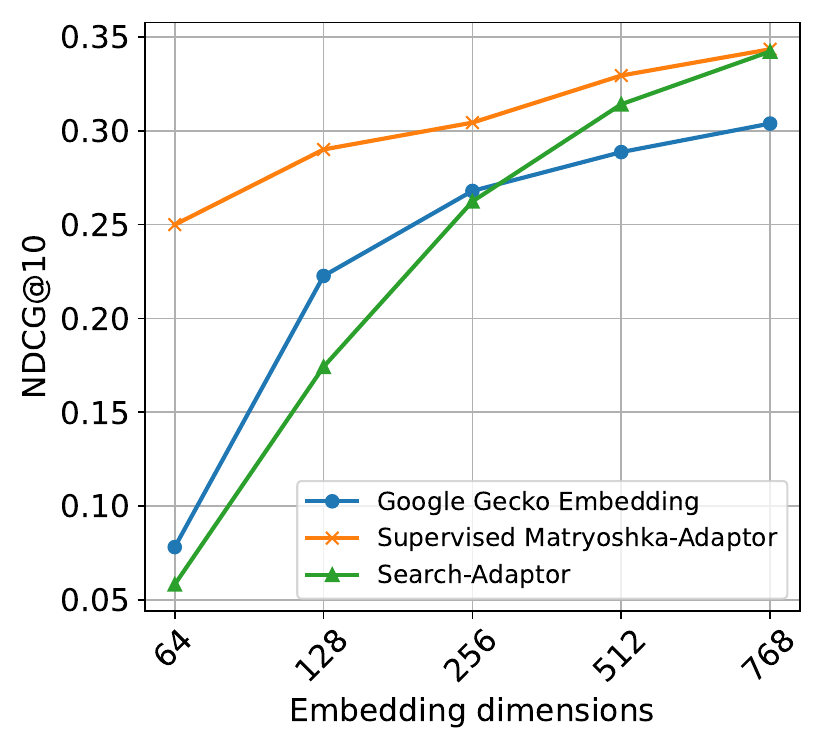}}
\caption{Experimental results of supervised Matryoshka-Adaptor with Google latest Gecko embedding models on 13 BEIR datasets.}
\label{fig:each_supervised_google_text_beir}
\end{figure*}

\newpage

\subsection{Supervised Matryoshka-Adaptor with OpenAI text-embedding-3-large models}

\begin{figure*}[h!]
\subfloat[NFCorpus]{
\includegraphics[width=0.24\textwidth]{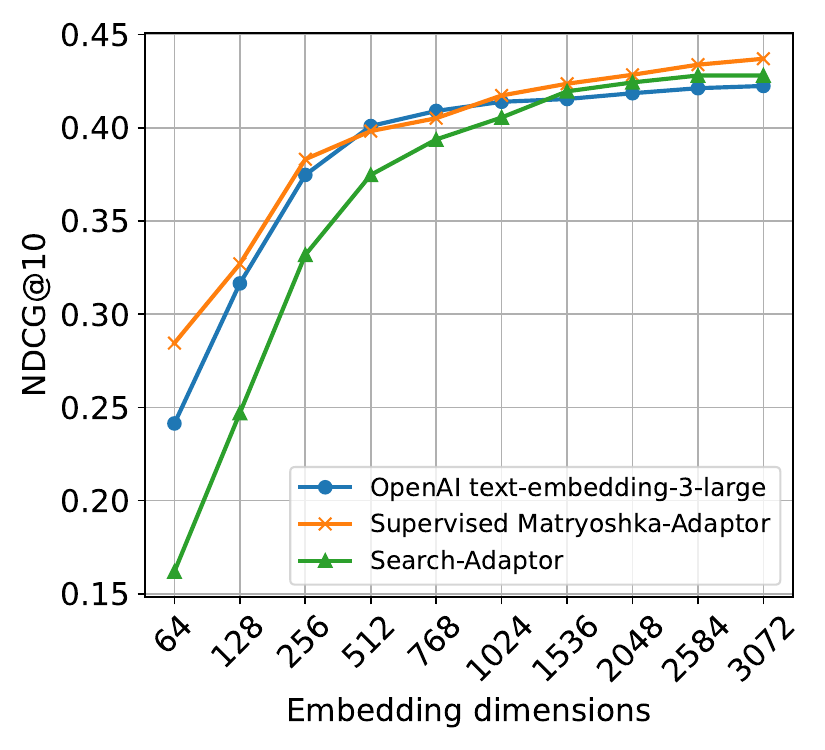}}
\subfloat[Scifact]{
\includegraphics[width=0.24\textwidth]{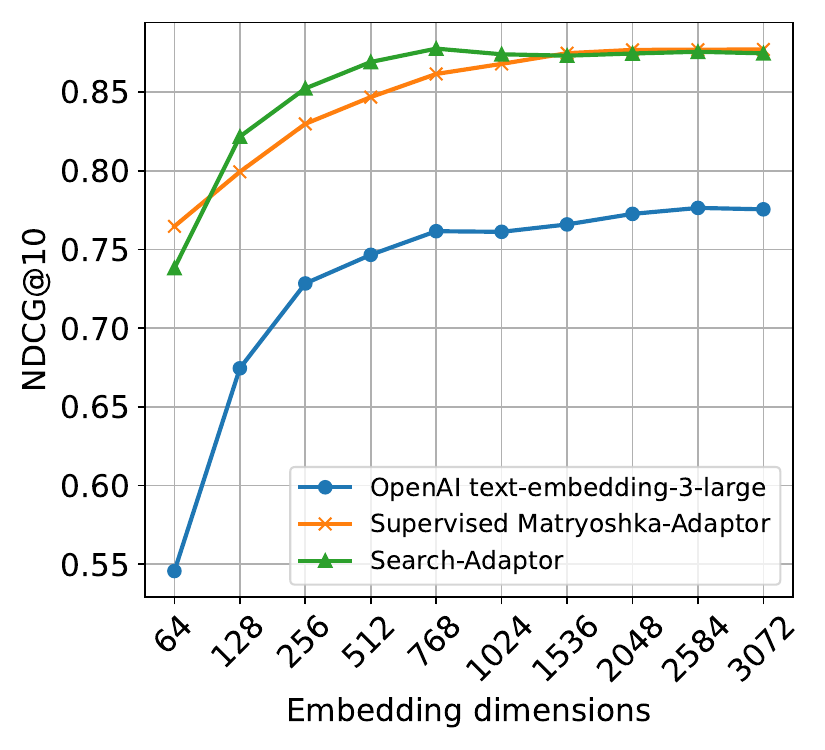}}
\subfloat[Arguana]{
\includegraphics[width=0.24\textwidth]{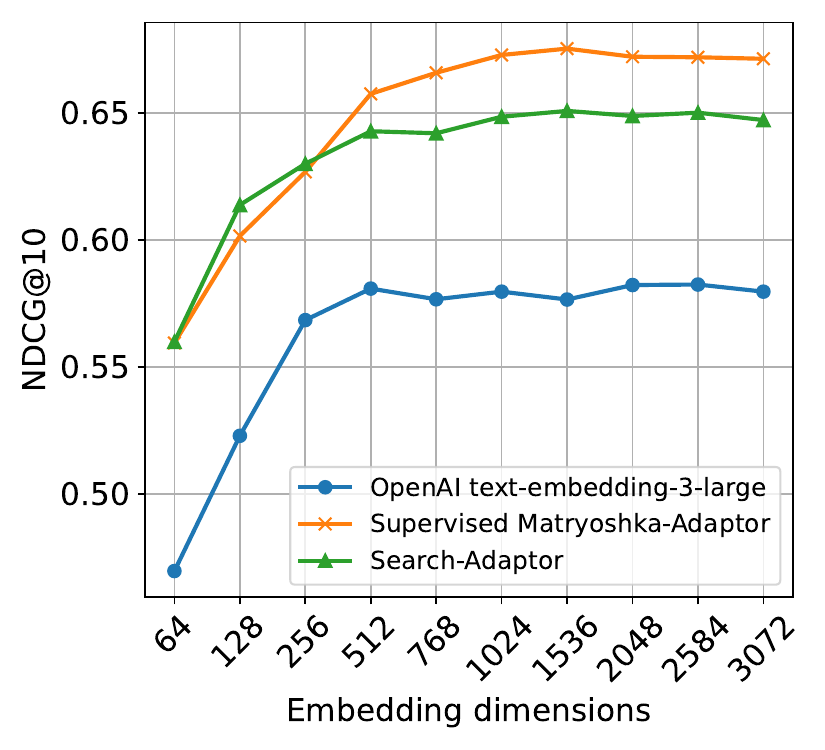}}
\subfloat[SciDocs]{
\includegraphics[width=0.24\textwidth]{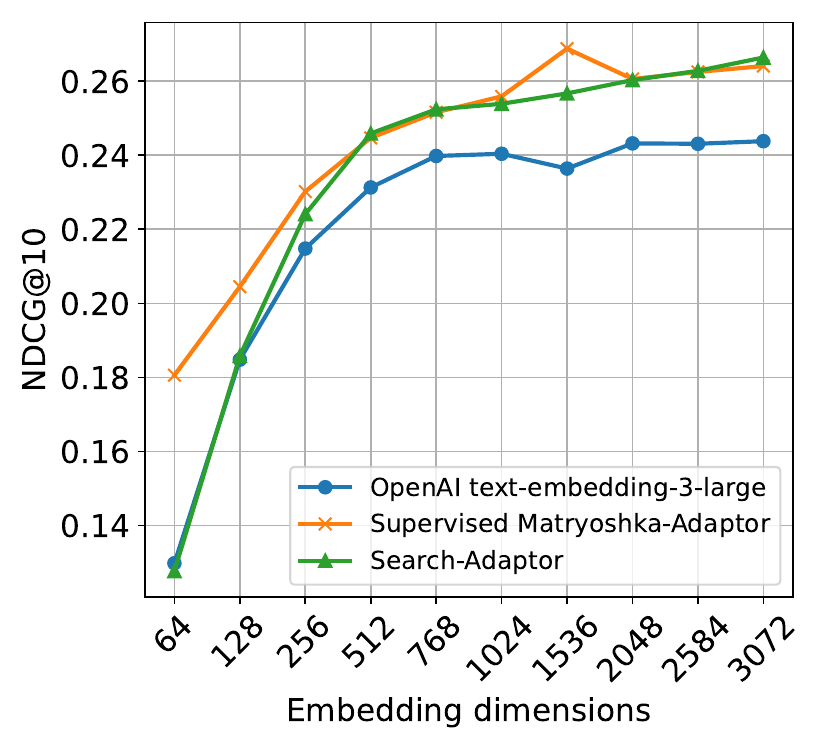}}\\
\subfloat[FiQA]{
\includegraphics[width=0.24\textwidth]{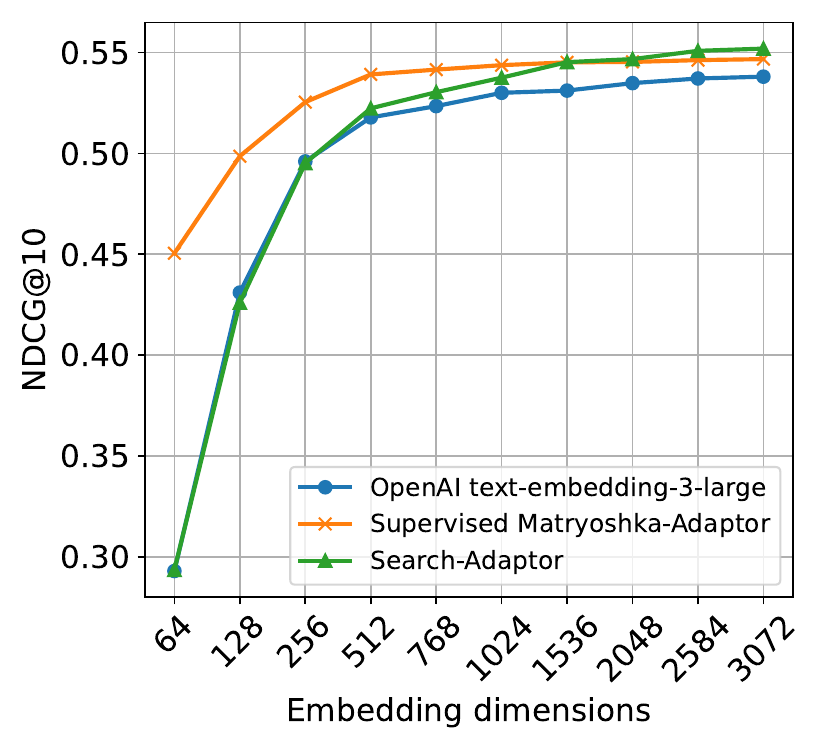}}
\subfloat[Trec-Covid]{
\includegraphics[width=0.24\textwidth]{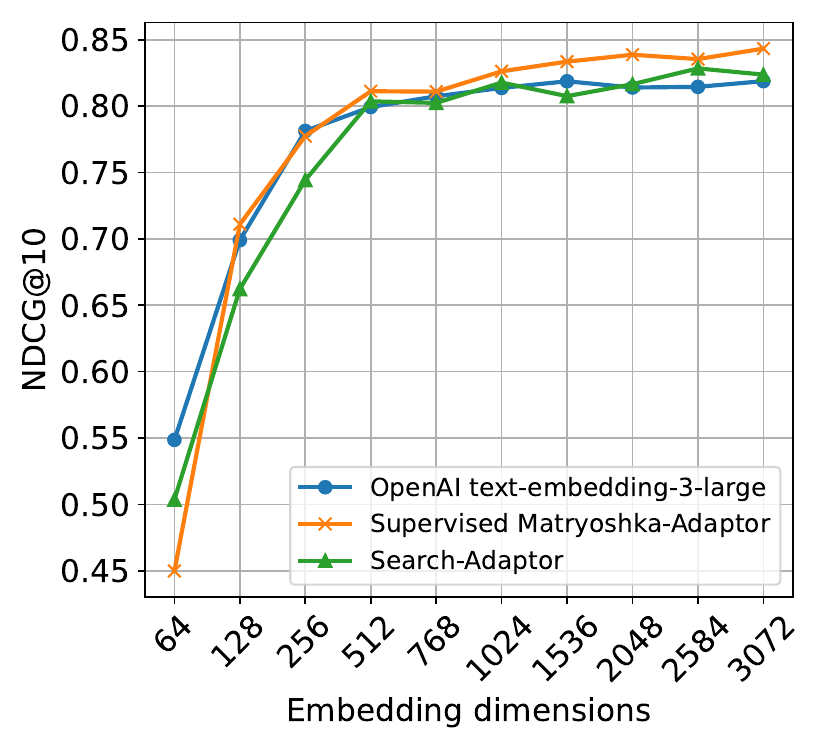}}
\subfloat[Touche]{
\includegraphics[width=0.24\textwidth]{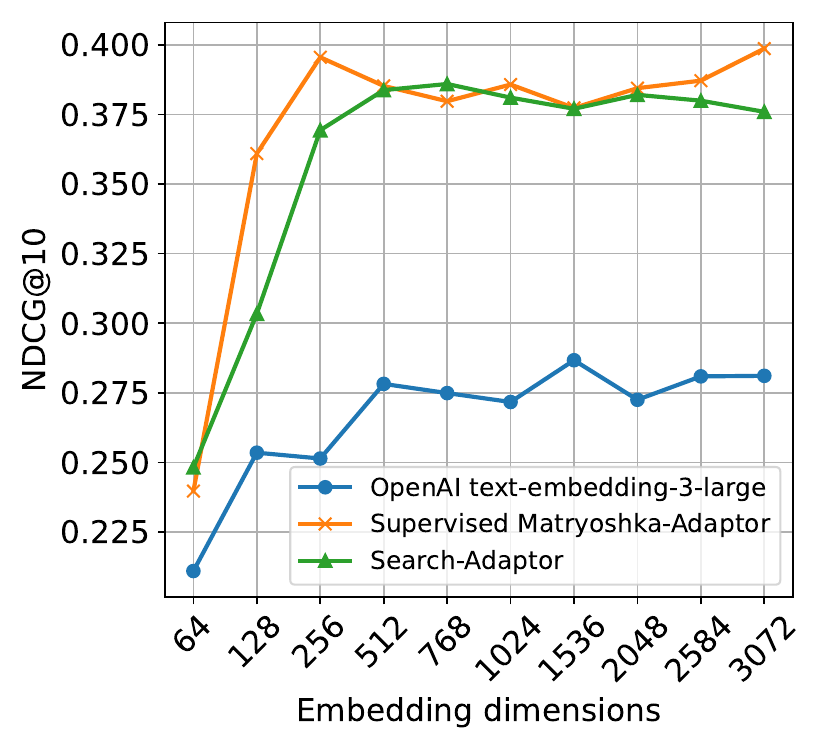}}
\subfloat[Quora]{
\includegraphics[width=0.24\textwidth]{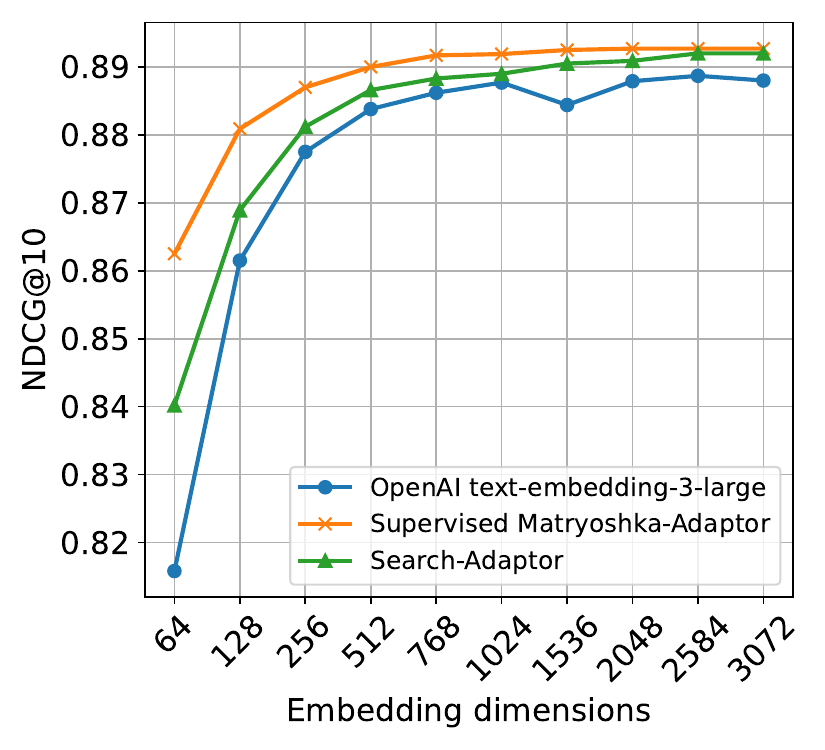}}\\
\caption{Experimental results of supervised Matryoshka-Adaptor with OpenAI text-embedding-3-large models on 8 BEIR datasets.}
\label{fig:each_supervised_openai_large_text_beir}
\end{figure*}

\subsection{Supervised Matryoshka-Adaptor with OpenAI text-embedding-3-small models}

\begin{figure*}[h!]
\subfloat[NFCorpus]{
\includegraphics[width=0.24\textwidth]{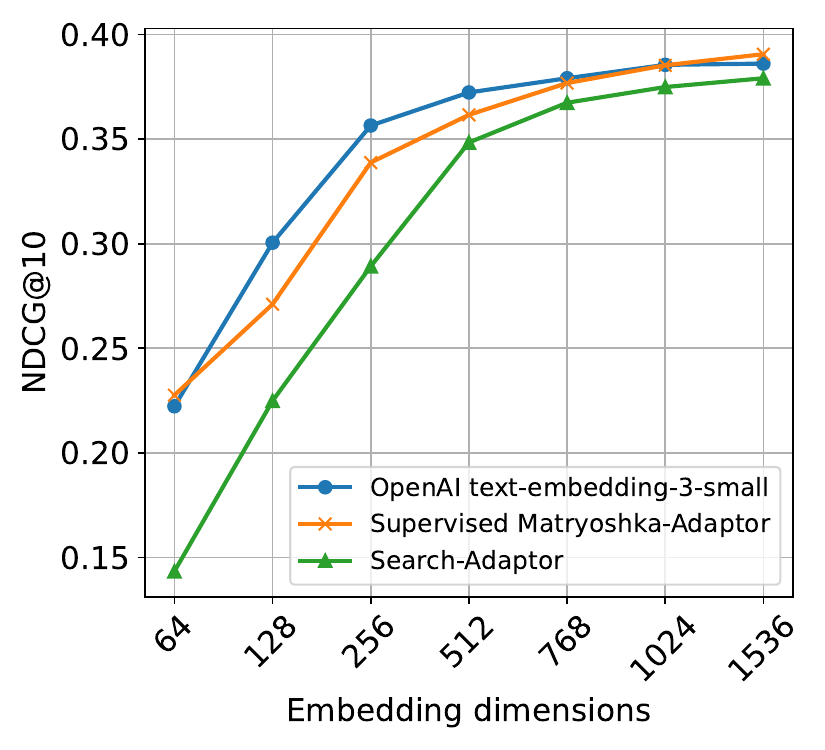}}
\subfloat[Scifact]{
\includegraphics[width=0.24\textwidth]{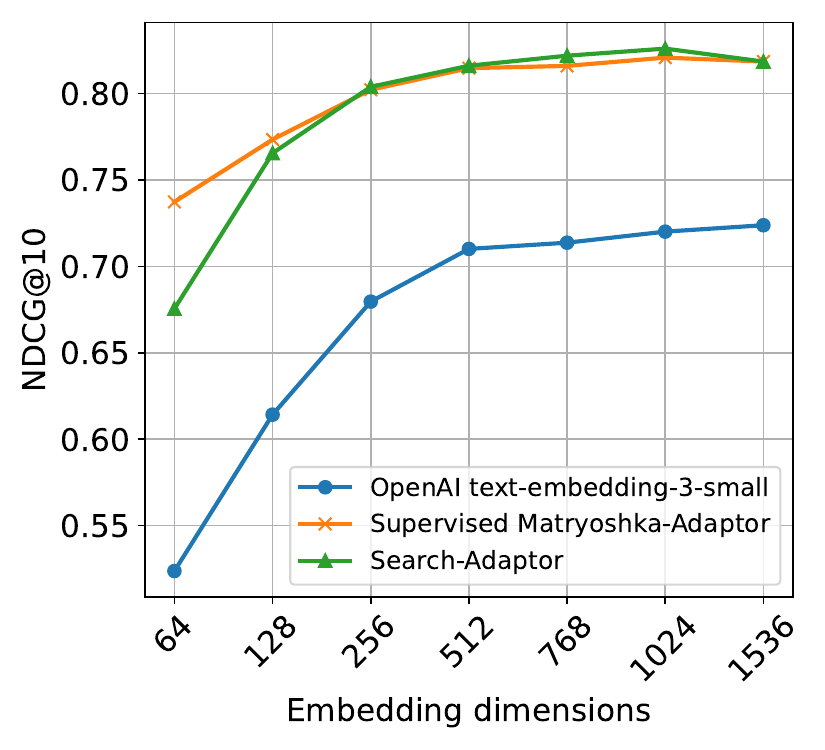}}
\subfloat[Arguana]{
\includegraphics[width=0.24\textwidth]{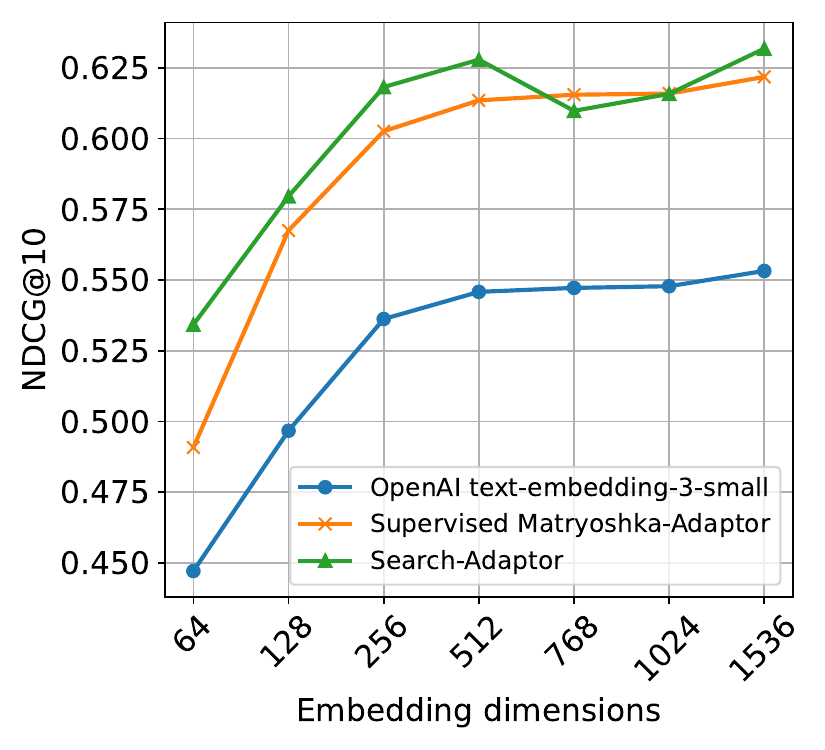}}
\subfloat[SciDocs]{
\includegraphics[width=0.24\textwidth]{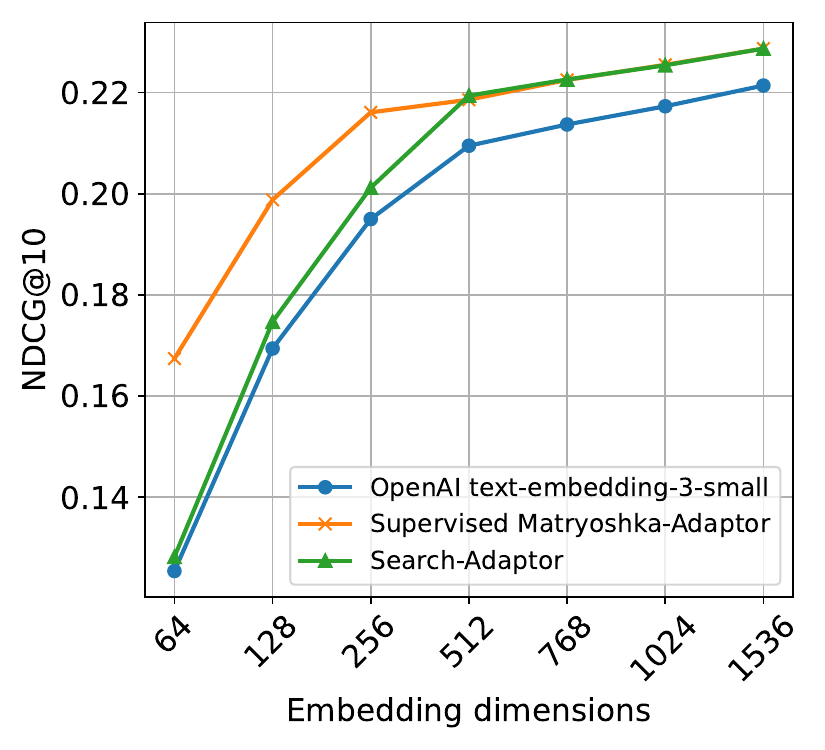}}\\
\subfloat[FiQA]{
\includegraphics[width=0.24\textwidth]{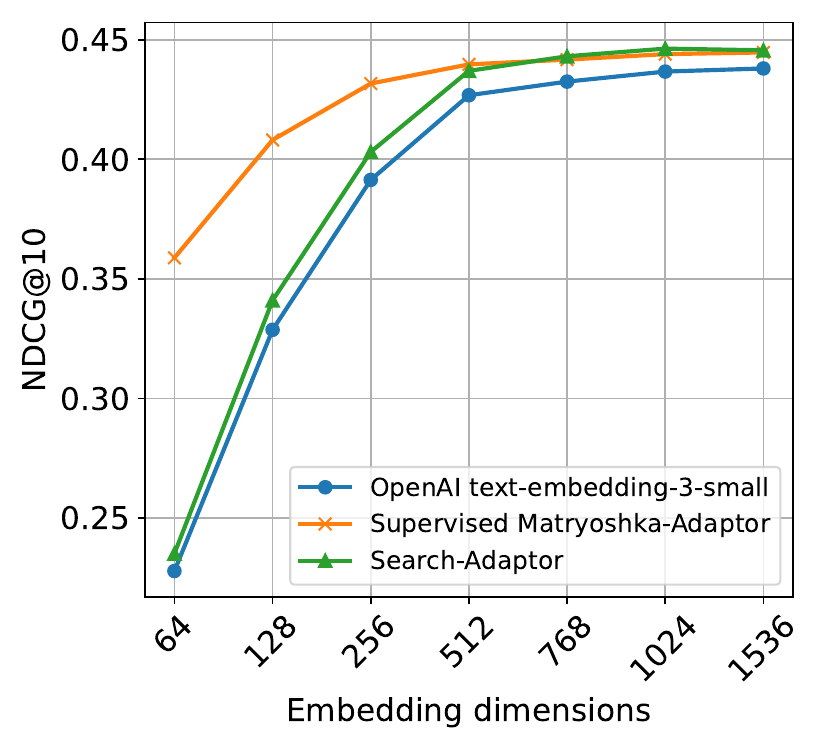}}
\subfloat[Trec-Covid]{
\includegraphics[width=0.24\textwidth]{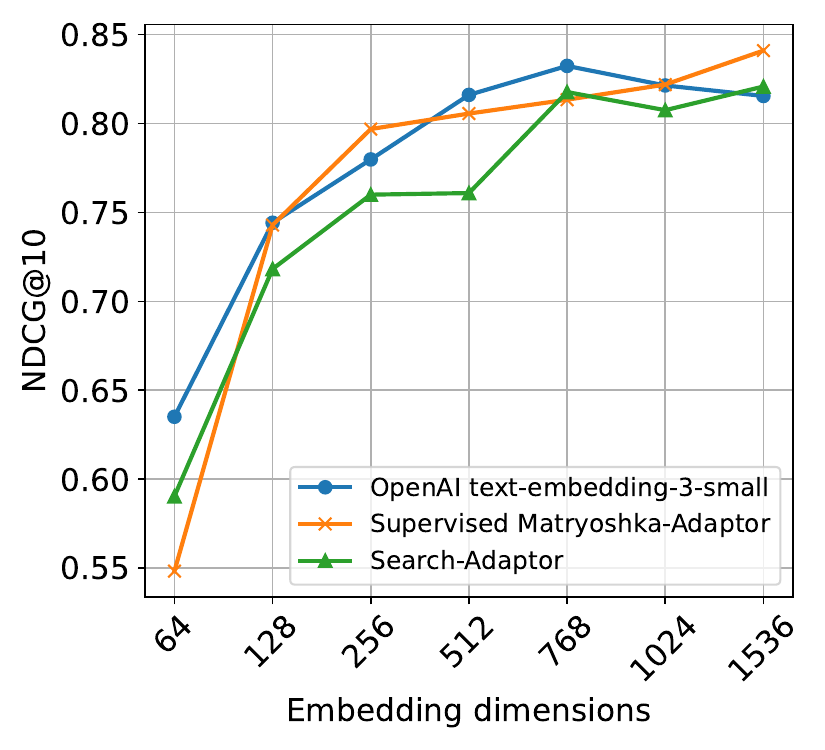}}
\subfloat[Touche]{
\includegraphics[width=0.24\textwidth]{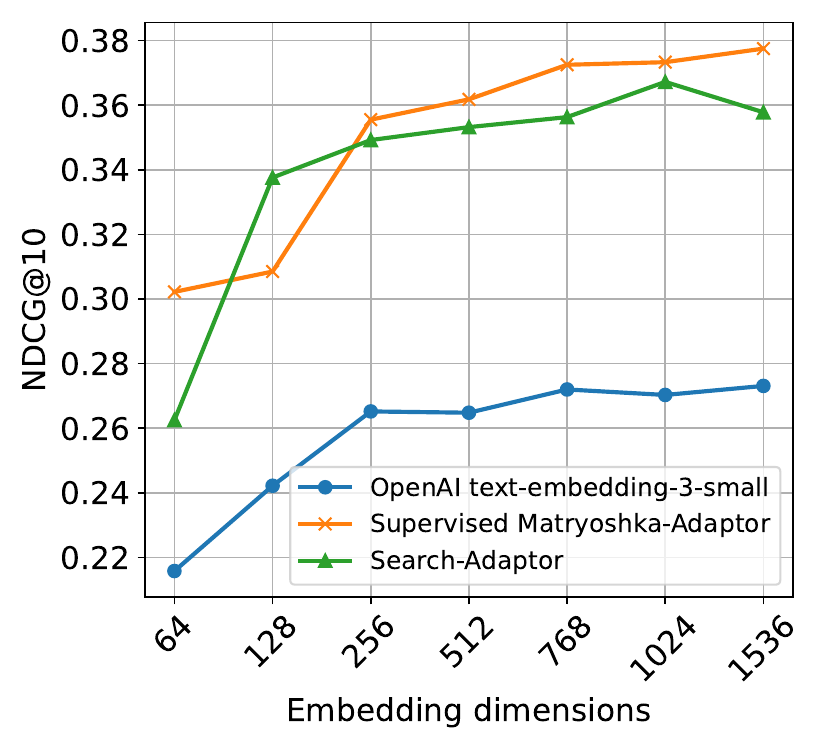}}
\subfloat[Quora]{
\includegraphics[width=0.24\textwidth]{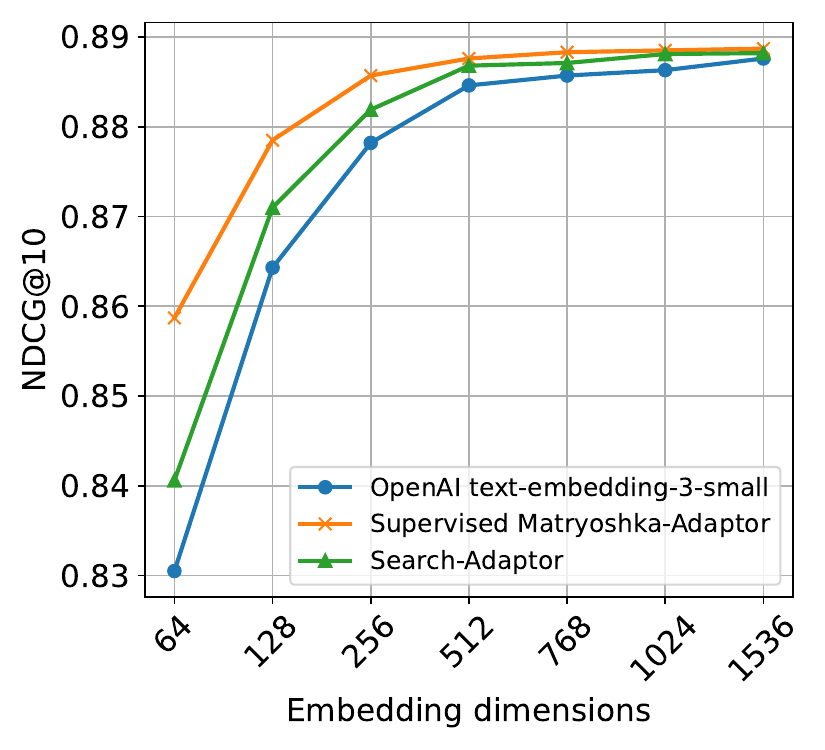}}\\
\caption{Experimental results of supervised Matryoshka-Adaptor with OpenAI text-embedding-3-small models on 8 BEIR datasets.}
\label{fig:each_supervised_openai_small_text_beir}
\end{figure*}

\newpage

\subsection{Supervised Matryoshka-Adaptor with Google Gecko multilingual embedding models}

\begin{figure*}[h!]
\subfloat[Yoruba (yo)]{
\includegraphics[width=0.24\textwidth]{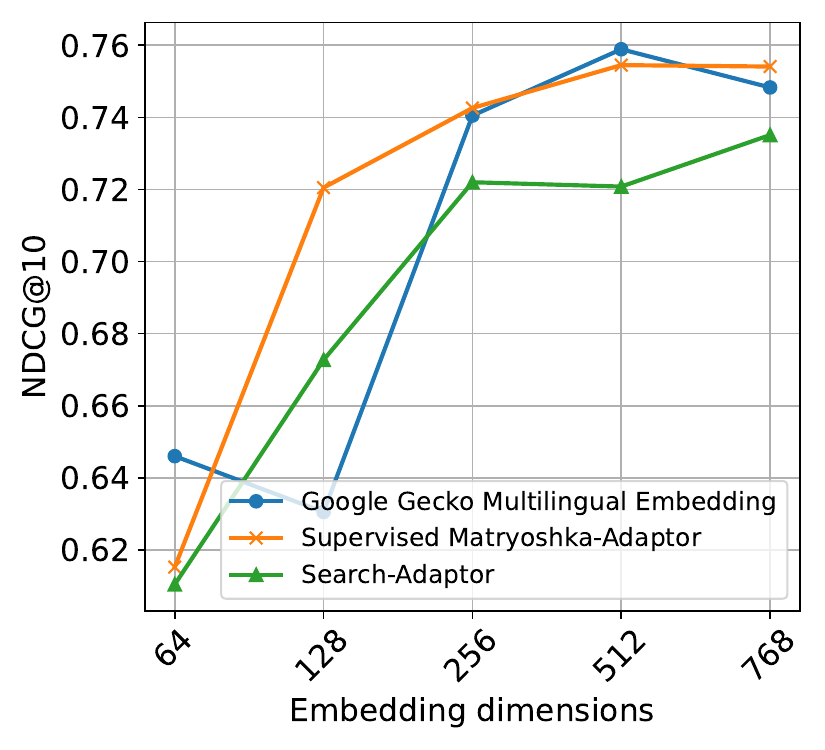}}
\subfloat[Swahilli (sw)]{
\includegraphics[width=0.24\textwidth]{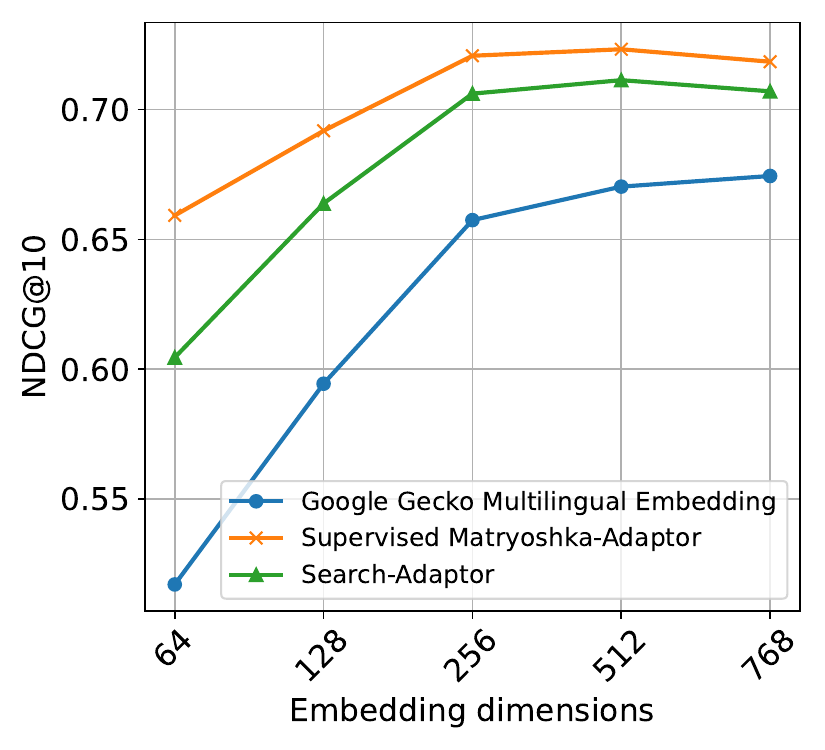}}
\subfloat[Bengali (bn)]{
\includegraphics[width=0.24\textwidth]{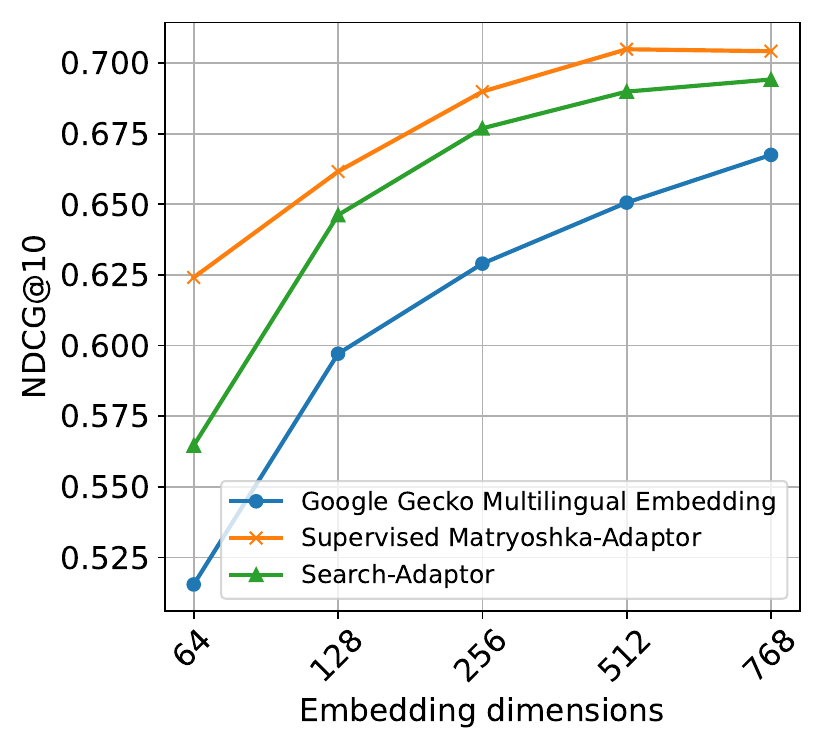}}
\subfloat[Hindi (hi)]{
\includegraphics[width=0.24\textwidth]{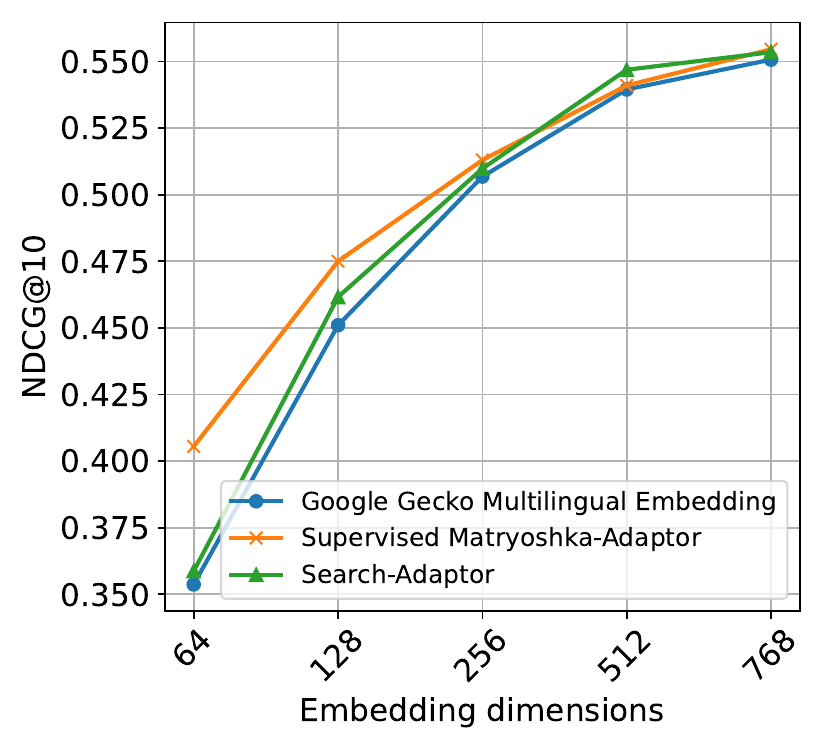}}\\
\subfloat[Telugu (te)]{
\includegraphics[width=0.24\textwidth]{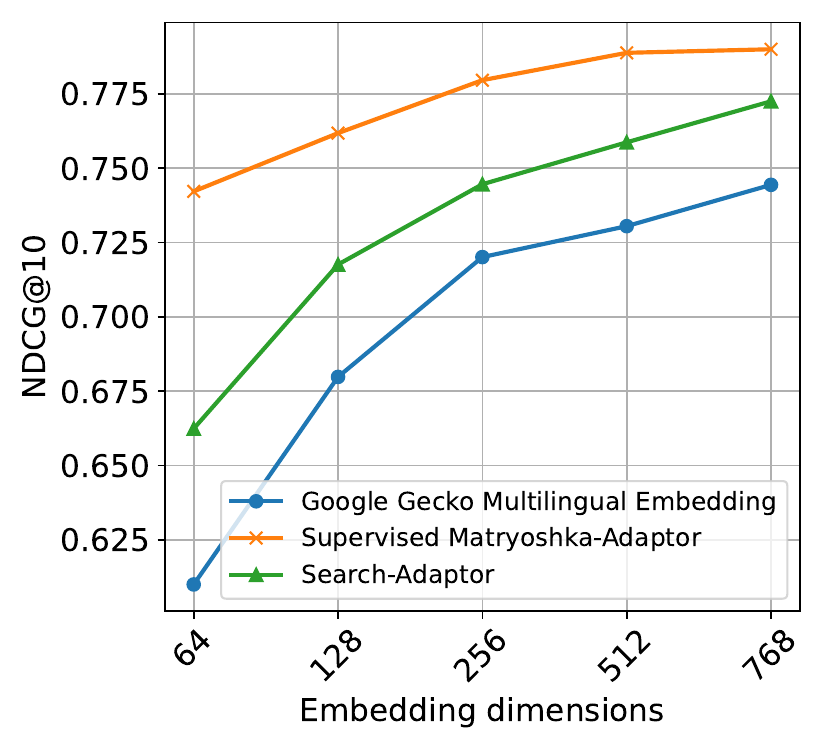}}
\subfloat[Thai (th)]{
\includegraphics[width=0.24\textwidth]{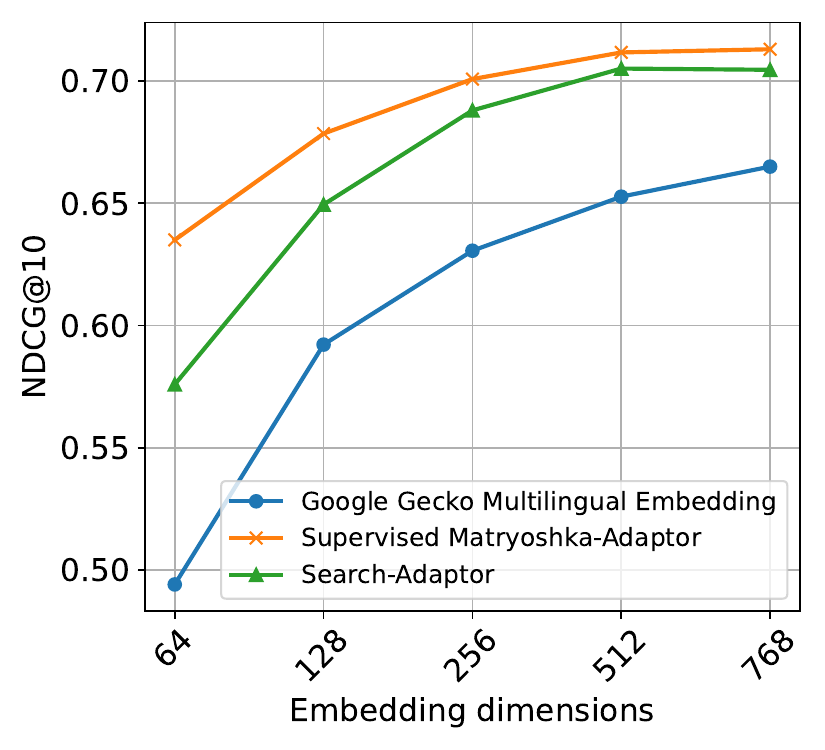}}
\subfloat[Indonesian (id)]{
\includegraphics[width=0.24\textwidth]{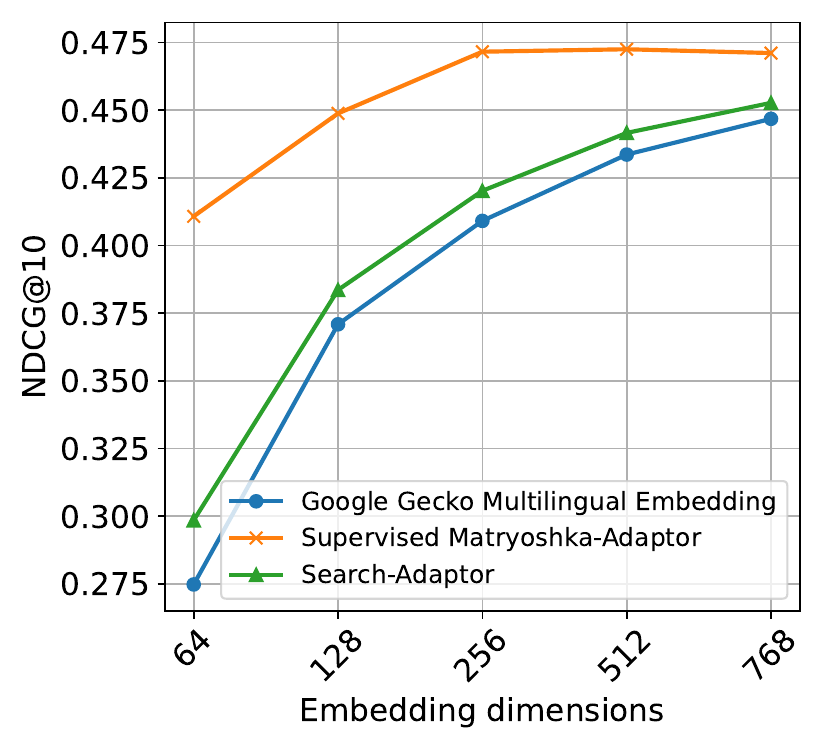}}
\subfloat[Korean (ko)]{
\includegraphics[width=0.24\textwidth]{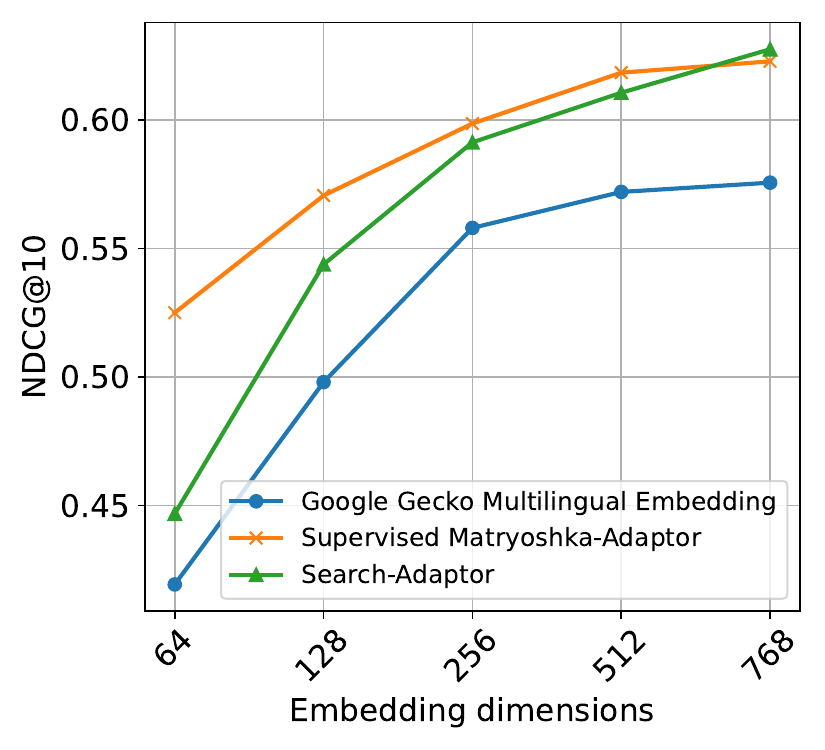}}\\
\subfloat[Finnish (fi)]{
\includegraphics[width=0.24\textwidth]{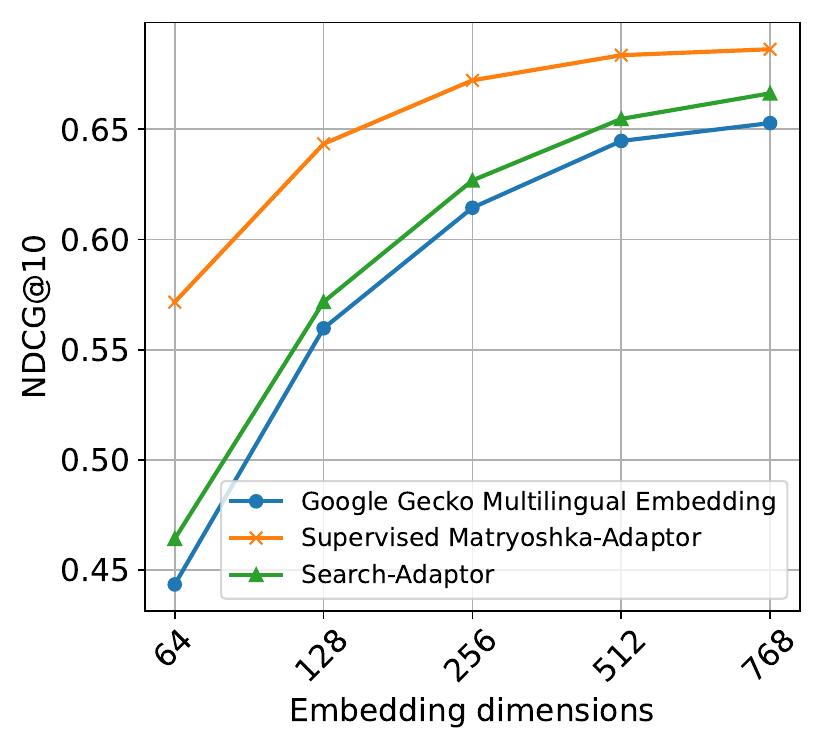}}
\subfloat[Arabic (ar)]{
\includegraphics[width=0.24\textwidth]{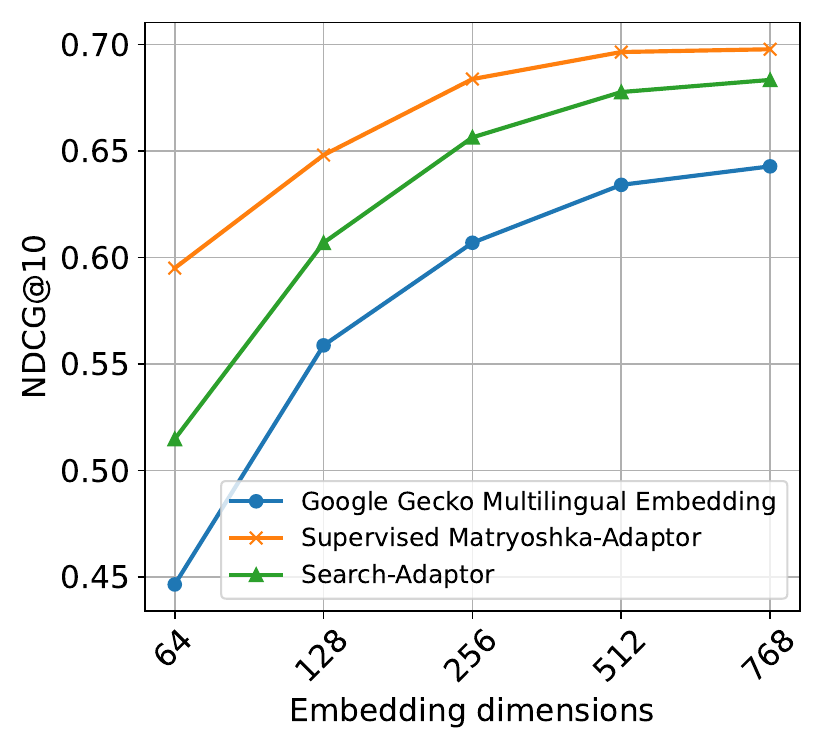}}
\subfloat[Persian (fa)]{
\includegraphics[width=0.24\textwidth]{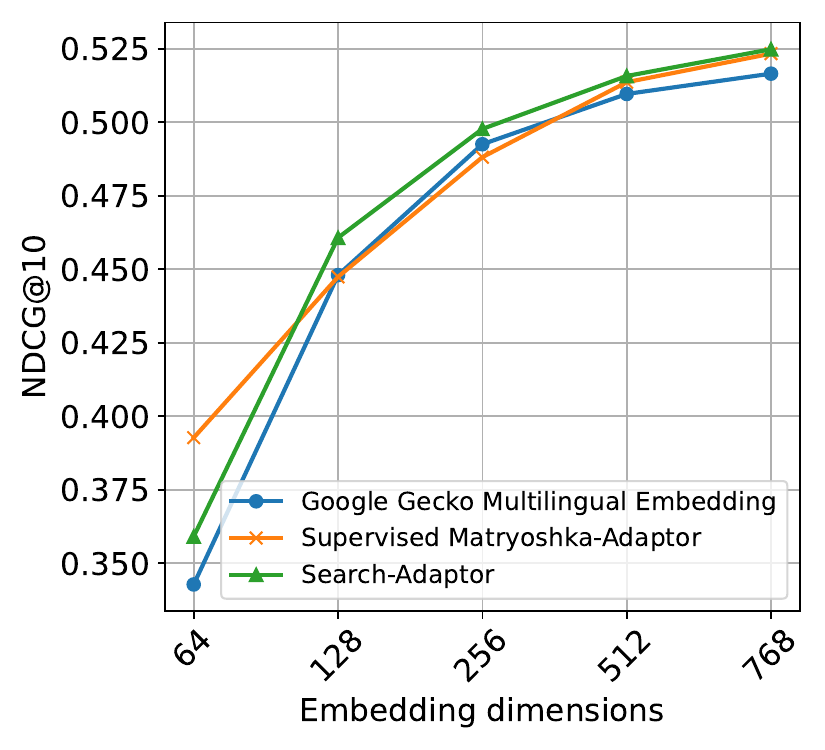}}
\subfloat[Chinese (zh)]{
\includegraphics[width=0.24\textwidth]{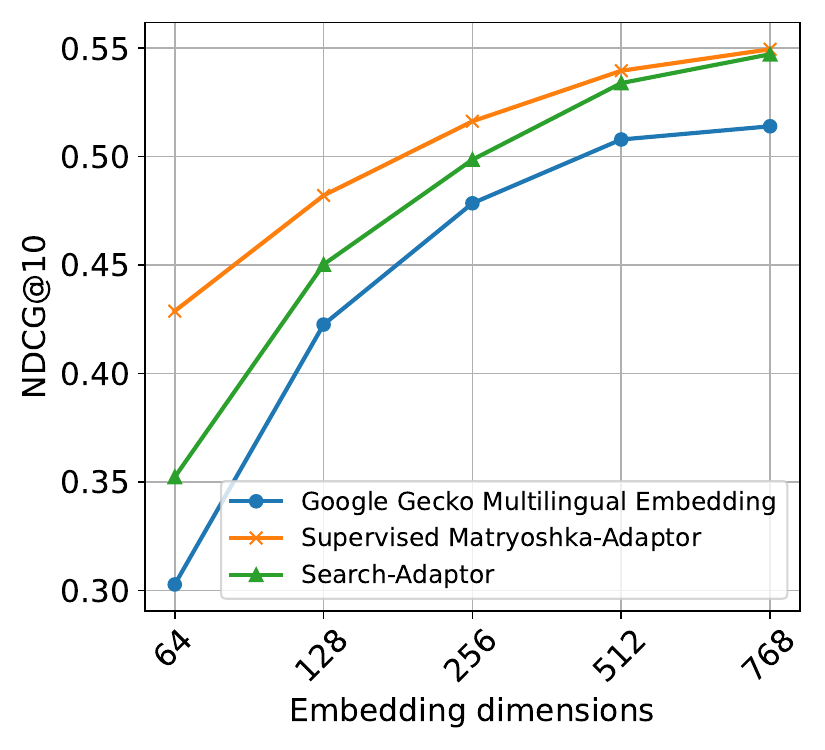}}\\
\subfloat[Japanese (ja)]{
\includegraphics[width=0.24\textwidth]{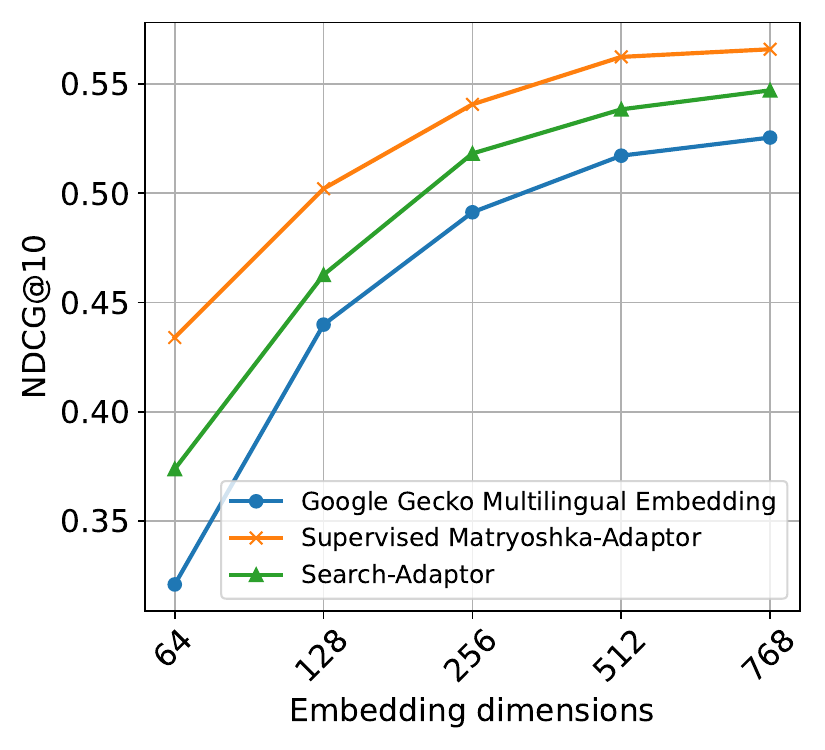}}
\subfloat[Russian (ru)]{
\includegraphics[width=0.24\textwidth]{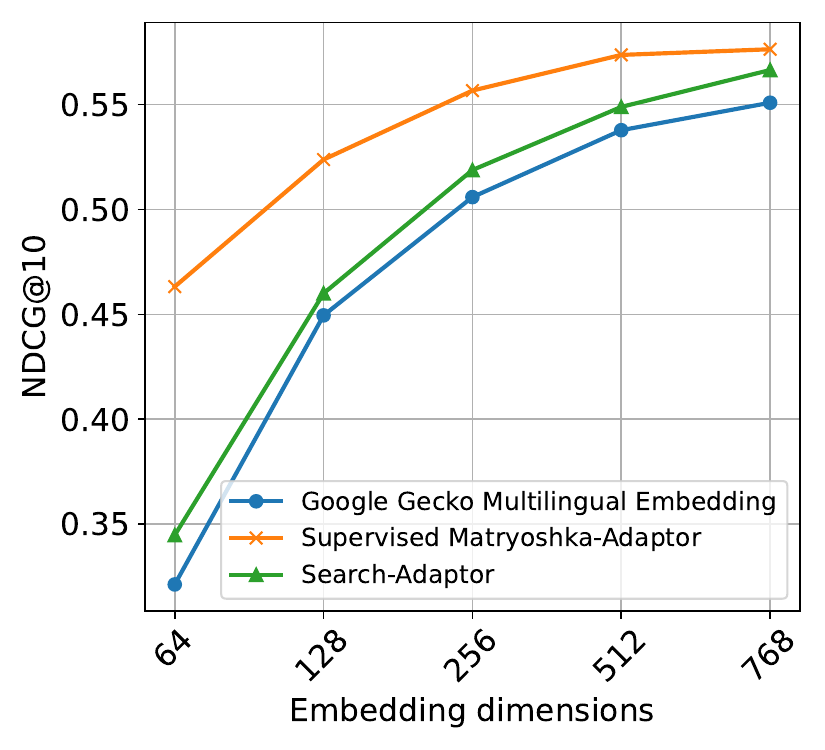}}
\subfloat[Spannish (es)]{
\includegraphics[width=0.24\textwidth]{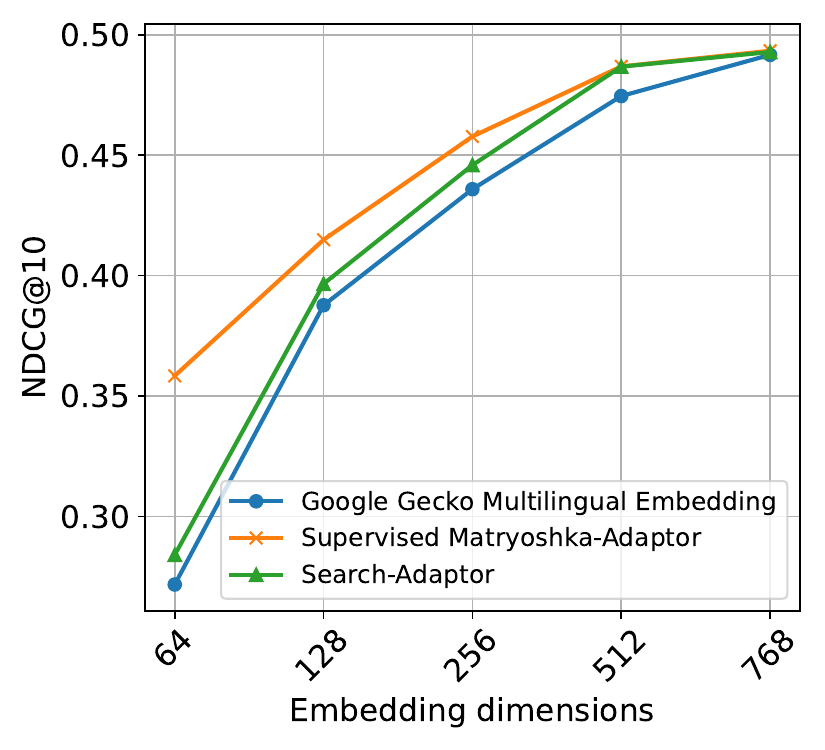}}
\subfloat[French (fr)]{
\includegraphics[width=0.24\textwidth]{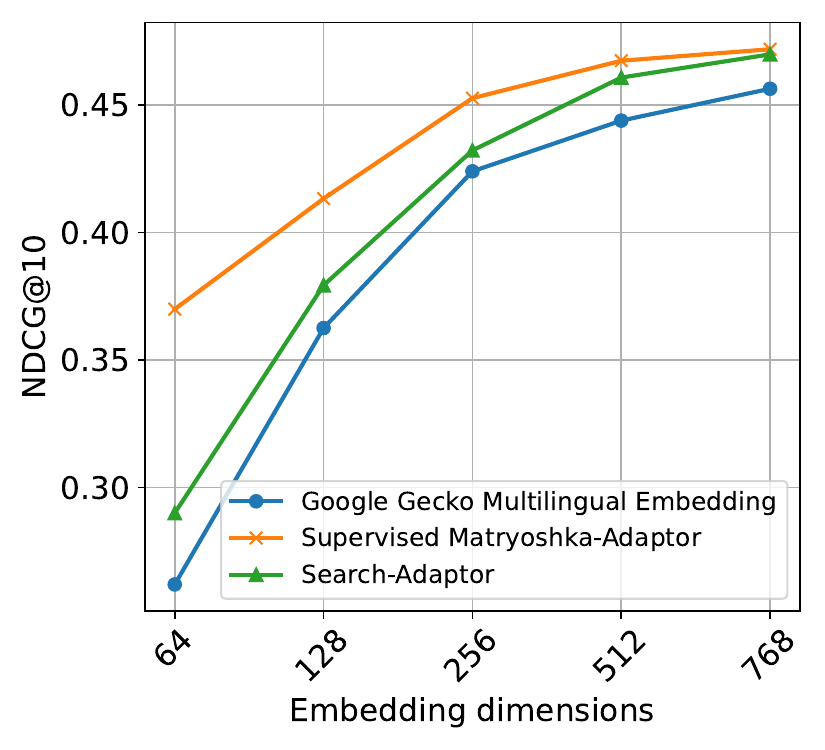}}\\
\subfloat[Germany (de)]{
\includegraphics[width=0.24\textwidth]{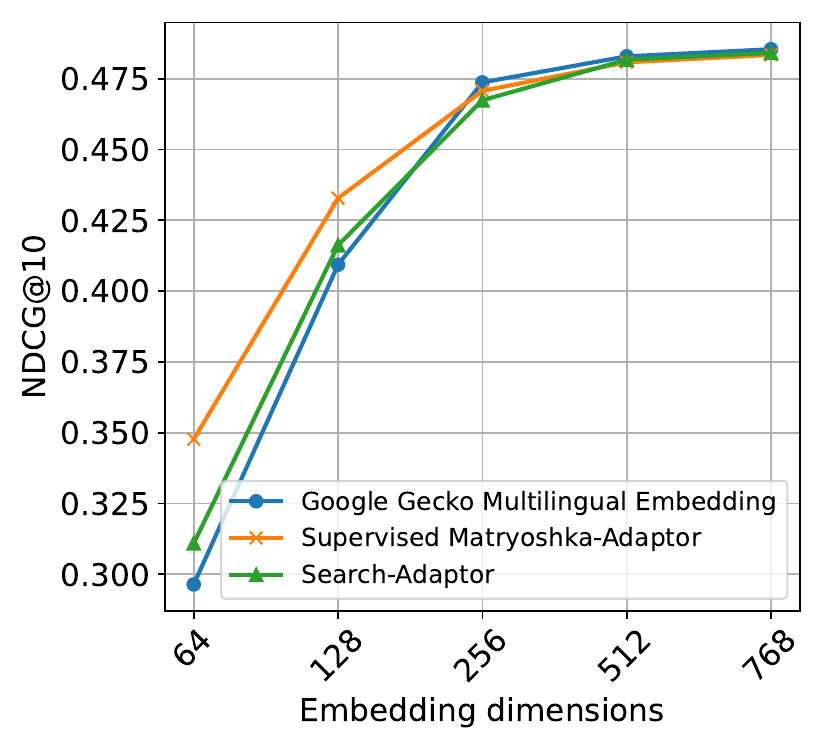}}\\
\caption{Experimental results of supervised Matryoshka-Adaptor with Google Gecko multilingual embedding models on 17 MIRACL datasets.}
\label{fig:each_supervised_google_multilingual_miracl}
\end{figure*}

\newpage

\subsection{Supervised Matryoshka-Adaptor with Google multimodal embedding models}

\begin{figure*}[h!]
\subfloat[Dresses]{
\includegraphics[width=0.24\textwidth]{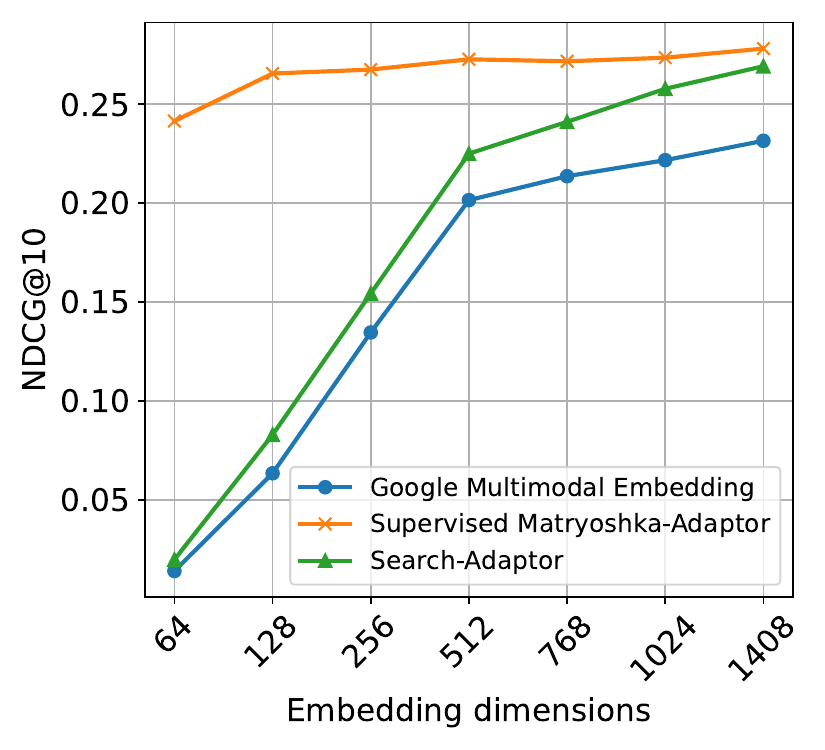}}
\subfloat[Jackets]{
\includegraphics[width=0.24\textwidth]{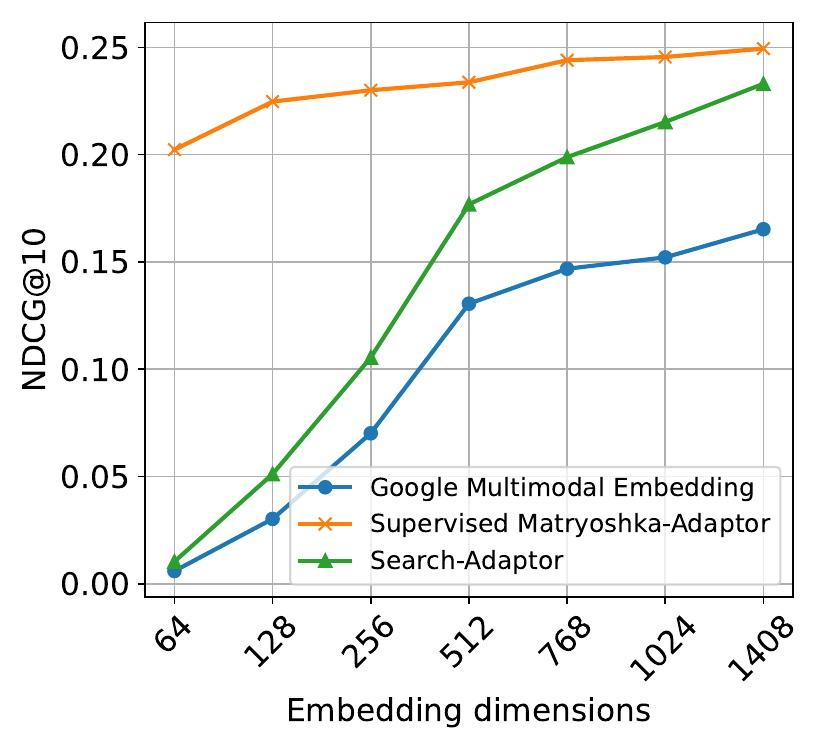}}
\subfloat[Pants]{
\includegraphics[width=0.24\textwidth]{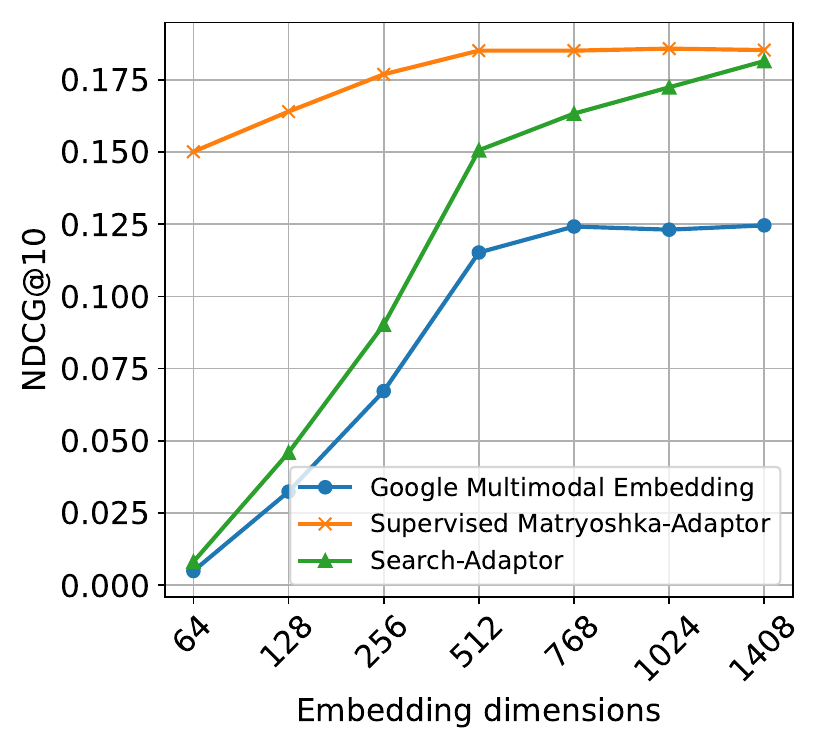}}
\subfloat[Skirts]{
\includegraphics[width=0.24\textwidth]{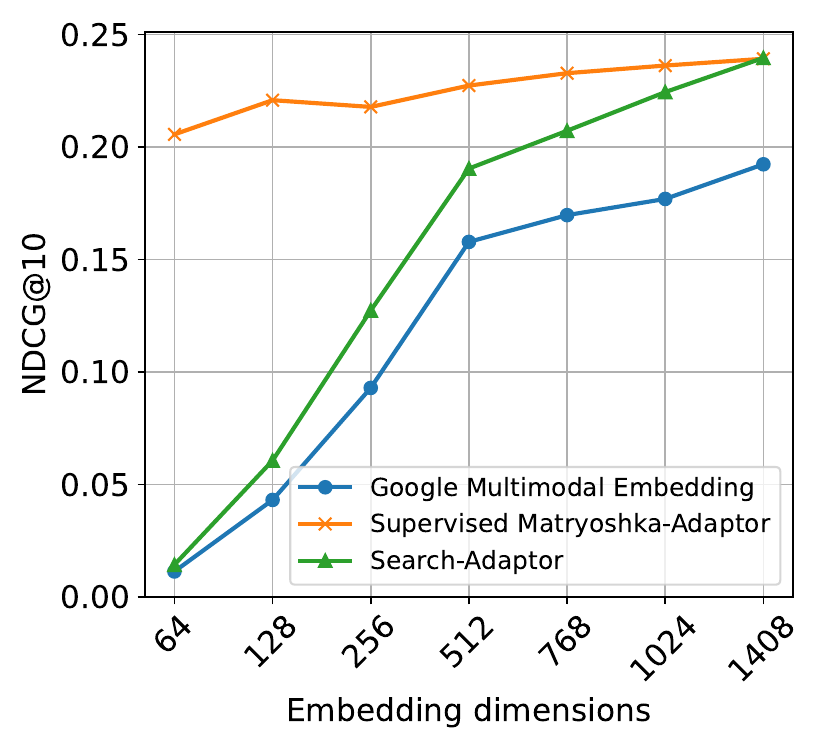}}\\
\subfloat[Tops]{
\includegraphics[width=0.24\textwidth]{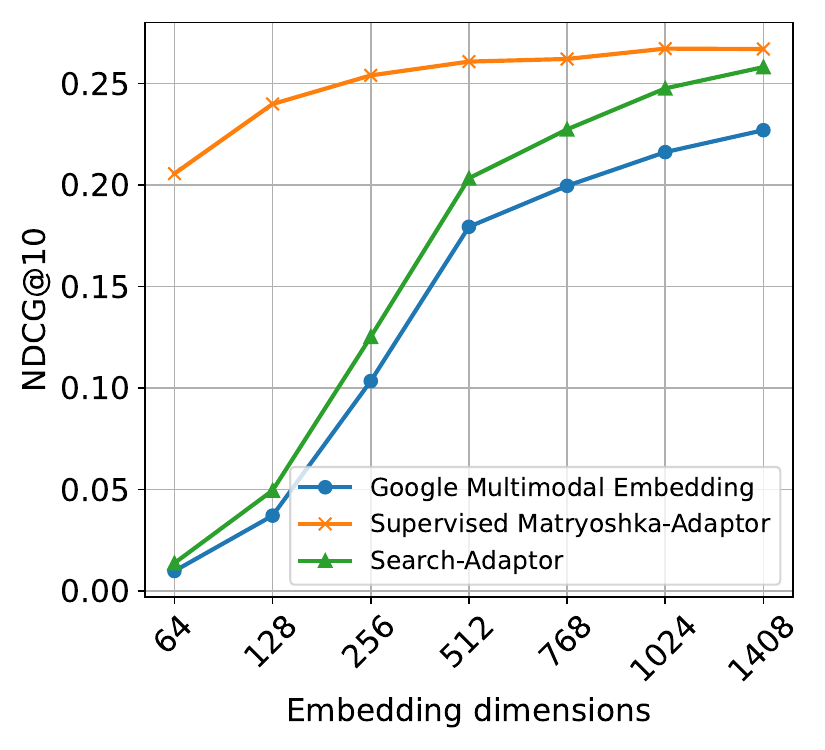}}
\caption{Experimental results of supervised Matryoshka-Adaptor with Google multimodal embedding models on 5 Fashion-200K datasets.}
\label{fig:each_supervised_google_multimodal_fashion}
\end{figure*}

\end{document}